%% file: preprint.tex
\documentclass{article}
\usepackage[preprint]{colm2025_conference}

\usepackage{microtype}
\usepackage[utf8]{inputenc} 
\usepackage[T1]{fontenc}    
\definecolor{green}{RGB}{0,150,10}
\definecolor{blue}{RGB}{0,148,181}
\definecolor{orange}{RGB}{194,153,107}
\usepackage[colorlinks,linkcolor=red,anchorcolor=green,citecolor=blue]{hyperref}
\usepackage{url}            

\usepackage{lineno}

\usepackage{booktabs}       
\usepackage{amsfonts}       
\usepackage{nicefrac}       
\usepackage{xcolor}         
\usepackage{amsmath}
\usepackage{marvosym}
\usepackage{graphicx}
\usepackage{colortbl}
\usepackage{multirow}
\usepackage{subcaption}
\usepackage{changes}
\usepackage{listings}
\usepackage{mdframed}
\usepackage{caption}
\usepackage{cleveref}
\usepackage{arydshln}
\usepackage{wrapfig}
\usepackage{titletoc}
\usepackage{tabularx}
\usepackage{geometry}
\usepackage{array}
\usepackage{tcolorbox}
\newcommand\DoToC{%
  \startcontents
  \printcontents{}{1}{\noindent \vskip3pt\vskip5pt}
  \vskip3pt\vskip5pt
}

\title{MME-Reasoning: A Comprehensive Benchmark for Logical\\ Reasoning in MLLMs}

\newcommand{\homepage}{\raisebox{-1.5pt}{\includegraphics[height=1em]{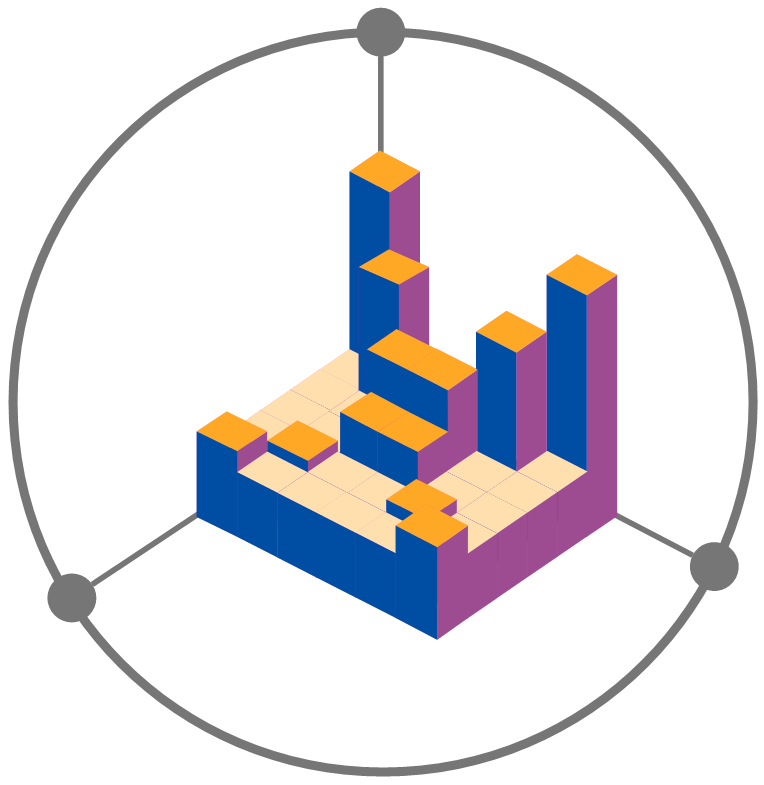}}}
\newcommand{\github}{\raisebox{-1.5pt}{\includegraphics[height=1em]{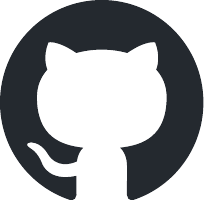}}}
\newcommand{\huggingface}{\raisebox{-1.5pt}{\includegraphics[height=1em]{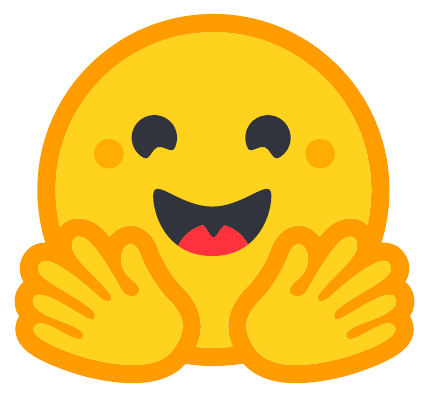}}}
\renewcommand{\thefootnote}{\fnsymbol{footnote}}
\author{%
  Jiakang Yuan$^{1,3,*}$, Tianshuo Peng$^{2,3,*}$, Yilei Jiang$^{2}$, Yiting Lu$^{4}$, Renrui Zhang$^{2}$,\\ \bf Kaituo Feng$^{2}$, Chaoyou Fu$^{5}$, Tao Chen$^{1,\text{\dag}}$, Lei Bai$^{3}$, Bo Zhang$^{3,\text{\dag}}$, Xiangyu Yue$^{2,3}$ \\ [1mm]
\textsuperscript{\rm 1} Fudan University~~
\textsuperscript{\rm 2} MMLab, The Chinese University of Hong Kong\\
\textsuperscript{\rm 3} Shanghai AI Laboratory~
\textsuperscript{\rm 4} University of Science and Technology of China~
\textsuperscript{\rm 5} Nanjing University\\ [1.5mm]
{\homepage\ \url{https://alpha-innovator.github.io/mmereasoning.github.io/}} \\
{\github\ \texttt{\url{https://github.com/Alpha-Innovator/MME-Reasoning}}} \\
  {\huggingface\ \texttt{\url{https://huggingface.co/datasets/U4R/MME-Reasoning}}}
}

\begin{document}

\maketitle

\renewcommand{\thefootnote}{}
\footnotetext{$^{*}$ Equal contribution, $^{\textrm{\dag}}$ Corresponding authors.}

\input{Sections/0_abs}

\begin{figure}[htbp]
\vspace{-10pt}
  \centering
  \includegraphics[width=0.95\linewidth]{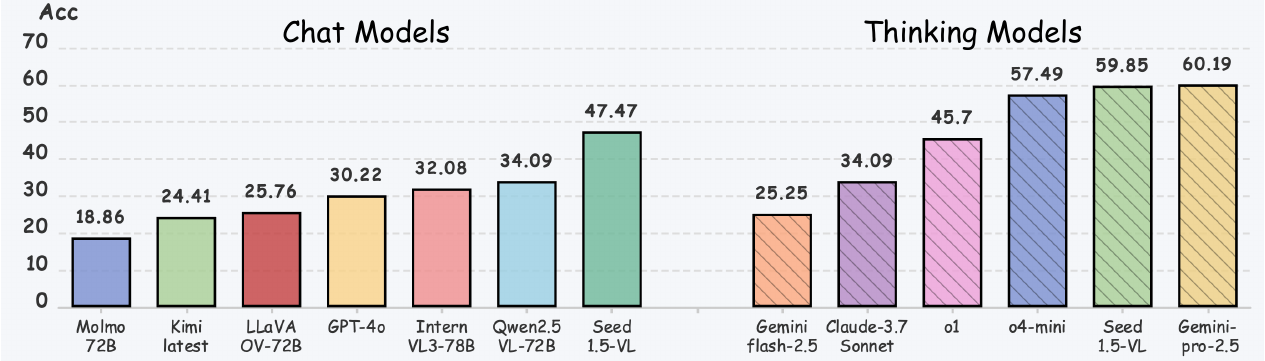}
  \caption{Performance comparison between thinking and chat models on MME-Reasoning.}
  \label{fig1}
\end{figure}

\input{Sections/1_intro}

\input{Sections/2_related}
\input{Sections/3_data}
\input{Sections/4_exp}

\input{Sections/5_conclusion}

\newpage

\bibliography{main}
\bibliographystyle{colm2025_conference}

\include{Sections/6_appendix}

\end{document}

%% file: Sections/0_abs.tex
\begin{abstract}
Logical reasoning is a fundamental aspect of human intelligence and an essential capability for multimodal large language models (MLLMs). Despite the significant advancement in multimodal reasoning, existing benchmarks fail to comprehensively evaluate their reasoning abilities due to the lack of explicit categorization for logical reasoning types and an unclear understanding of reasoning. To address these issues, we introduce \textbf{MME-Reasoning}, a comprehensive benchmark designed to evaluate the reasoning ability of MLLMs, which covers all three types of reasoning (\textit{i.e.}, inductive, deductive, and abductive) in its questions. We carefully curate the data to ensure that each question effectively evaluates reasoning ability rather than perceptual skills or knowledge breadth, and extend the evaluation protocols to cover the evaluation of diverse questions. Our evaluation reveals substantial limitations of state-of-the-art MLLMs when subjected to holistic assessments of logical reasoning capabilities. Even the most advanced MLLMs show limited performance in comprehensive logical reasoning, with notable performance imbalances across reasoning types. In addition, we conducted an in-depth analysis of approaches such as ``thinking mode'' and Rule-based RL, which are commonly believed to enhance reasoning abilities. These findings highlight the critical limitations and performance imbalances of current MLLMs in diverse logical reasoning scenarios, providing comprehensive and systematic insights into the understanding and evaluation of reasoning capabilities.
\end{abstract}


%% file: Sections/1_intro.tex
\section{Introduction}
\vspace{-8pt}

Logical reasoning~\citep{liu2025logicalsurvey}, a fundamental cognitive process of analyzing premises and evidence to reach valid conclusions, serves as the cornerstone of human intelligence. Multimodal reasoning~\citep{o1} enables humans to integrate information from different modalities, such as visual and text, which is essential for tackling complex tasks. 
Recently, with the emergence of reasoning large language models (LLMs)~\citep{dubey2024llama,yang2024qwen2} such as DeepSeek-R1~\citep{deepseekai2025deepseekr1incentivizingreasoningcapability}, injecting reasoning capability into multimodal large language models (MLLMs)~\citep{gpt4o,Qwen2.5-VL,li2024llavaov} has begun to be explored~\citep{peng2025lmm-r1,zhang2025r1-vl,huang2025vision-r1}. Despite the significant progress in reasoning MLLMs, a comprehensive evaluation of their capabilities still remains an open challenge. Therefore, it is particularly important to establish a fair and robust evaluation benchmark for assessing the reasoning capabilities of MLLMs and further accelerate the development of this field.


Currently, most benchmarks~\citep{fu2023mme,wang2024mathvision,lu2023mathvista,yue2023mmmu,yue2024mmmu,gong2024av,he2024olympiadbench} designed for multimodal reasoning primarily focus on knowledge-driven tasks. For example, MathVista~\citep{lu2023mathvista} and MathVerse~\citep{zhang2024mathverse} provide comprehensive evaluations of MLLMs' mathematical reasoning abilities. OlympiadBench~\citep{he2024olympiadbench} and EMMA~\citep{hao2025emma} expand the scope to include additional subjects, such as physics and chemistry. Apart from knowledge-driven tasks, some works~\citep{song2025visualpuzzles,chia2024puzzlevqa,zhang2025puzzlebench} have begun to decouple knowledge from logical reasoning, aiming to assess the reasoning abilities of MLLMs independent of specific domain knowledge. For instance, SciVerse~\citep{guo2025sciverse} and VisualPuzzles~\citep{song2025visualpuzzles} focus on reasoning-focused, knowledge-light tasks.


Despite recent advances, existing benchmarks still suffer from several problems as outlined below.


\textbf{\textit{First, lacking explicit categorization of reasoning and insufficient coverage of reasoning types.}} 
In logic, reasoning is typically classified into three types: \textbf{\textit{abduction, deduction, and induction}}~\citep{peirce2014illustrations}.
Most existing benchmarks primarily concentrate on evaluating MLLMs' inductive and deductive reasoning ability. For example, most of the questions in MathVerse~\citep{lu2023mathvista} belong to deductive reasoning, which uses rules and premises to derive conclusions. PuzzleVQA~\citep{chia2024puzzlevqa} only contains questions of inductive reasoning, which learns rules based on premises and conclusions. However, abductive reasoning ability (\textit{i.e.}, exploring premises to explain a conclusion based on the conclusion and rules) is rarely evaluated. 
\textbf{\textit{Second, the concept of reasoning is not clear enough,}} which is reflected in confusing perception with reasoning or equating reasoning with the complexity of the required knowledge. For example, MathVista~\citep{lu2023mathvista} contains many questions that can be answered through visual perception, while OlympiadBench~\citep{he2024olympiadbench} includes questions that require advanced domain knowledge, which the model may not have access to. This may lead to an inaccurate evaluation of MLLMs' reasoning ability.

\begin{wraptable}{r}{0.6\linewidth}
\vspace{-8pt}
\caption{Response token length on different datasets.}
\vspace{-4pt}
\centering
\small
\resizebox{1.0\linewidth}{!}{
\begin{tabular}{l c c c }
\toprule
\textbf{Model}  & \textbf{MathVista} & \textbf{MathVerse} & \textbf{MME-R.} \\
\cmidrule{1-4}
Qwen2.5-VL-7B & 209.5 & 207.6 & \textbf{442.8} \\
GPT-4o & 162.6 & 157.3 & \textbf{328.0} \\
Claude-3.7-Sonnet-T & 519.4 & 563.2 & \textbf{979.2} \\
\bottomrule[1pt]
\label{tab:length_compare}
\end{tabular}
}
\vspace{-20pt}
\end{wraptable}

\begin{figure}[t]
\vspace{-10pt}
  \centering
  \includegraphics[width=0.98\linewidth]{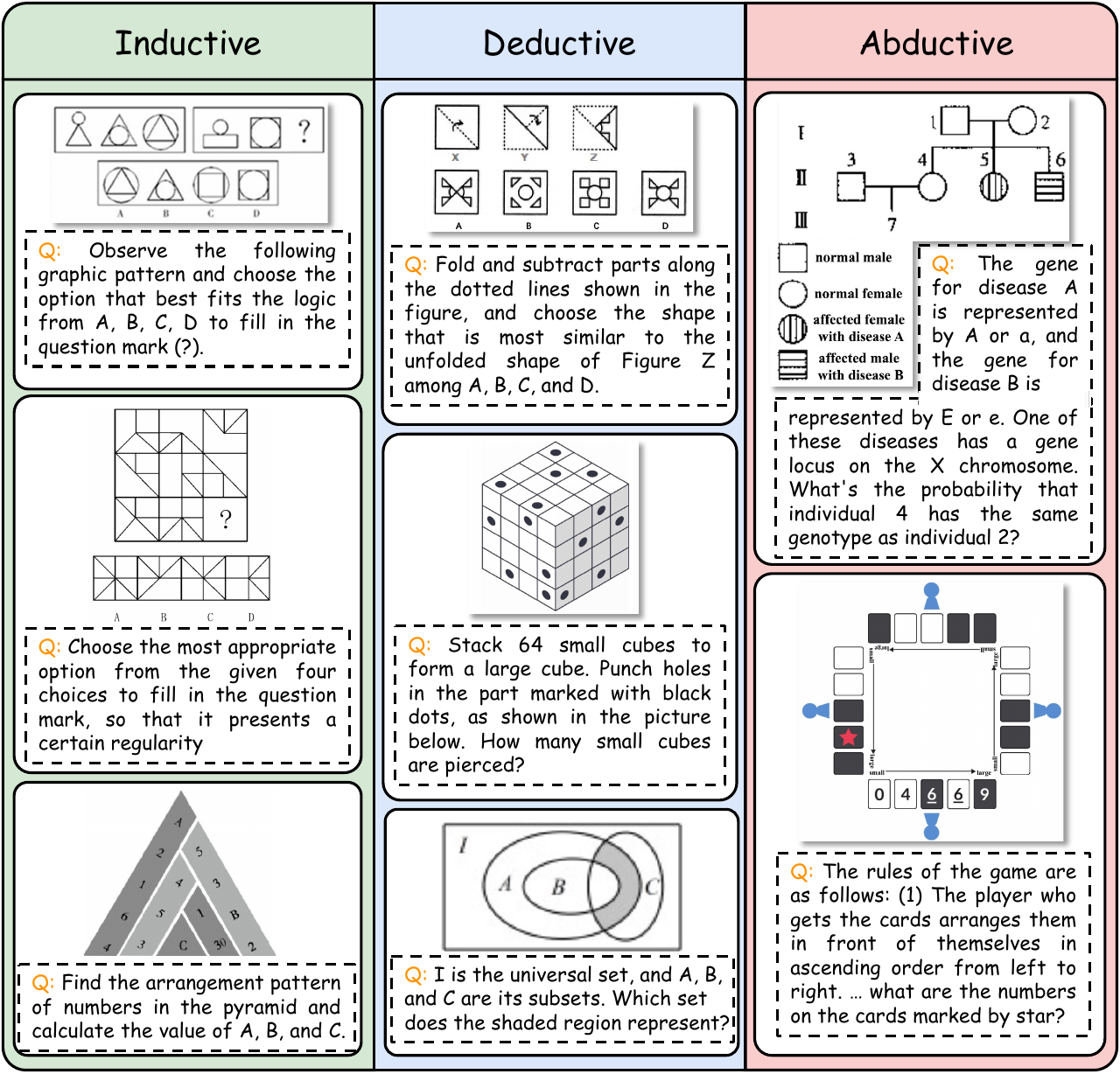}
  \caption{Example of questions in MME-Reasoning which covers comprehensive reasoning types.}
  \vspace{-14pt}
  \label{fig2:reasoning_case}
\end{figure}

To address these issues, we introduce MME-Reasoning, a comprehensive benchmark specifically designed to evaluate the reasoning capability of MLLMs. MME-Reasoning consists of 1,188 carefully curated questions that systematically cover types of logical reasoning (\textit{i.e.}, inductive, deductive, and abductive), while spanning a range of difficulty levels, as illustrated in Fig.~\ref{fig2:reasoning_case}. Besides, we identify 5 key abilities related to multimodal reasoning, including calculation, planning and exploring, spatial-temporal, pattern analysis, and casual chaining analysis, and annotate the type of ability assessed by each question. 
To ensure a true evaluation of reasoning ability, MME-Reasoning eliminates questions that can be answered purely through perception or require complex domain knowledge, thereby focusing on the core reasoning skills of the model. 
We report the average response lengths of three representative models across different datasets in Tab.~\ref{tab:length_compare}. Results show that responses on MME-Reasoning are significantly longer than those on previous reasoning benchmarks, indicating its challenging and rigorous demands on model reasoning.
Furthermore, MME-Reasoning incorporates a variety of evaluation methods, including multiple-choice, free-form, and rule-based (\textit{e.g.}, Sudoku Puzzles) questions. Employing multiple evaluation methods enables a wider variety of question types, thereby facilitating a more comprehensive evaluation of models' capabilities.

Experiments were conducted on state-of-the-art MLLMs, covering Chat and Thinking types of both open-source and closed-source, as presented in Fig.~\ref{fig1}.
Evaluations with MME-Reasoning reveal these key findings:

\begin{itemize}
\item \textbf{MLLMs exhibit significant limitations and pronounced imbalances in reasoning capabilities.}
Even the most advanced MLLMs achieve only limited results under holistic logical reasoning evaluation, with Gemini-Pro-2.5-Thinking scoring only 60.19\%, followed by Seed1.5-VL (59.85) and o4-mini (57.49\%). These results indicate that MME-Reasoning, through its comprehensive evaluation of all the logical reasoning types, establishes a systematic and challenging benchmark for multimodal reasoning.

\item \textbf{Abductive reasoning remains a major bottleneck for current MLLMs.}
While most models demonstrate competent deductive reasoning, their abductive reasoning lags significantly. Closed-source models exhibit an average gap of 5.38 points between deductive and abductive tasks, which further widens to 9.81 among open-source models, making abductive reasoning a key bottleneck. Since it underpins many real-world tasks, addressing this gap is crucial for improving overall reasoning.

\item \textbf{Reasoning length scales with task difficulty, benefiting performance but accompanied by marginal effects and decreasing token efficiency.}
Thinking Models exhibit longer reasoning chains, particularly on more difficult questions, demonstrating adaptive inference budgeting and enhanced depth of reasoning. 
A positive correlation between average token count (ATC) and accuracy supports the effectiveness of extended outputs, especially in complex tasks. However, this performance gain plateaus beyond a certain length, revealing diminishing returns.
\end{itemize}

%% file: Sections/2_related.tex
\vspace{-8pt}
\section{Related Works}

\label{sec:related_works}

\subsection{Multimodal Reasoning}

Chain-of-thought (CoT) reasoning~\citep{wei2022CoT} has emerged as a key paradigm for enhancing the reasoning capability of LLMs. By generating intermediate steps before the final answer, CoT enables more transparent and accurate decision-making, especially in complex tasks such as arithmetic, logical deduction, and commonsense reasoning. Inspired by its success in text-only settings, CoT has recently been extended to MLLMs, giving rise to multimodal chain-of-thought (MCoT) reasoning~\citep{jiang2024rapguardsafeguardingmultimodallarge,zhang2023multimodal,chen2023see, peng2024chimera, lu2025omnicaptioner,xia2024geox}. 
Early approaches such as Multimodal-CoT~\citep{zhang2023multimodal} and IPVR~\citep{chen2023see} demonstrate that generating intermediate reasoning steps significantly improves model performance in visual question answering. Other methods such as HoT~\citep{yao2023HoT}, BDoG~\citep{zheng2024picture}, and VisualSketchpad~\citep{hu2024visual} introduce graph structures, debating agents, and visual intermediate states to further enhance interpretability and reasoning depth. 

More recently, following the success of Deepseek-R1, the Generalized Reinforcement Preference Optimization (GRPO) algorithm has gained traction in the development of multimodal models. Methods such as MM-EUREKA~\citep{meng2025mmeureka}, Vt-R1~\citep{zhou2025VisualThinker-R1-Zero}, LMM-R1~\citep{peng2025lmmr1}, and R1-V~\citep{chen2025r1v} adapt GRPO to solve mathematical geometry tasks, demonstrating promising reflective reasoning capabilities. 
Other works, including VLM-R1~\citep{shen2025vlmr1}, Visual-RFT~\citep{liu2025visual}, and Seg-Zero~\citep{liu2025segzero}, apply GRPO to enhance visual competencies such as grounding, object detection, and classification. The algorithm has also been extended to video and audio modalities through models such as Video-R1~\citep{feng2025video}, and R1-Omni~\citep{zhao2025r1omni}. 

\subsection{Multimodal Reasoning Benchmarks}

Recent benchmarks have advanced the evaluation of multimodal reasoning, particularly in visual-language settings. Early works such as CLEVR~\citep{johnson2016clevrdiagnosticdatasetcompositional} and GQA~\citep{DBLP:conf/cvpr/HudsonM19} assess compositional and spatial reasoning, while more recent benchmarks such as MathVista~\citep{DBLP:conf/iclr/LuBX0LH0CG024}, PuzzleBench~\citep{zhang2025puzzlebench}, ChartX~\citep{xia2024chartx} and PuzzleVQA~\citep{chia2024puzzlevqa} emphasize symbolic logic or pattern discovery. However, these benchmarks typically focus on narrow subtypes of reasoning—especially inductive logic—and fail to offer a holistic evaluation across deductive, inductive, and abductive paradigms.

Furthermore, many existing datasets conflate perception with reasoning. Tasks solvable via recognition or superficial pattern matching are often labeled as reasoning challenges, while high-difficulty benchmarks such as GPQA~\citep{rein2023gpqagraduatelevelgoogleproofqa}, OlympiaBench~\citep{he2024olympiadbench} and MME-CoT~\citep{jiang2025mmecotbenchmarkingchainofthoughtlarge} overly depend on domain-specific knowledge rather than logical inference. Evaluation protocols are also limited—most rely on multiple-choice formats and lack support for open-ended or rule-based assessment. In contrast, our benchmark provides a fine-grained evaluation of visual reasoning, explicitly covering the three classical reasoning types.


%% file: Sections/3_data.tex
\section{The MME-Reasoning Benchmark}

\begin{figure}[t]
  \centering
  \includegraphics[width=1.0\linewidth]{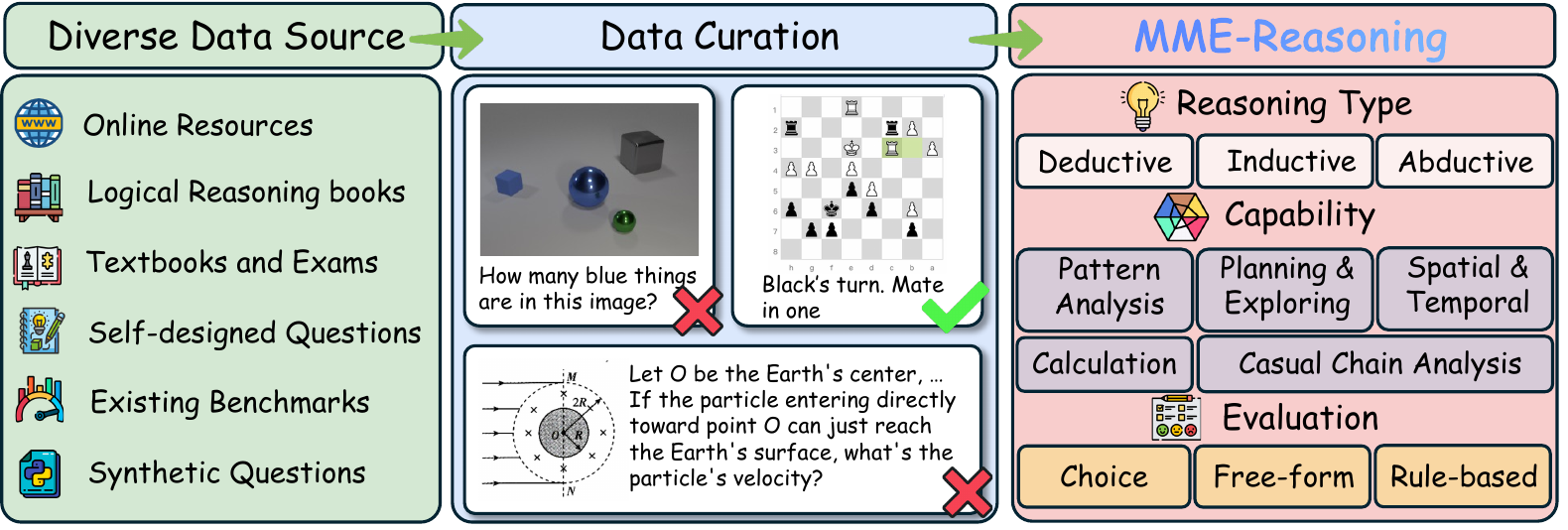}
  \caption{The overall construction process of MME-Reasoning.}
  \label{fig:pipeline}
\end{figure}

We introduce MME-Reasoning, a comprehensive benchmark designed to evaluate the reasoning ability of MLLMs. MME-Reasoning consists of 1,188 questions, including 1,008 newly collected items. MME-Reasoning comprehensively covers three types of reasoning (\textit{i.e.}, inductive, deductive, and abductive) and includes three question types (\textit{i.e.}, multiple-choice, free-form, and rule-based). We further divided MME-Reasoning into three difficulty levels (\textit{i.e.}, easy, medium, and hard). The key statistics and construction pipeline of MME-Reasoning are shown in Tab.~\ref{tab:statistics} and Fig.~\ref{fig:pipeline}.


\subsection{Design Principles of MME-Reasoning}
\label{sec:design_principle}

To ensure a comprehensive evaluation of multimodal reasoning and address issues present in previous benchmarks, such as incomplete coverage of reasoning types, unclear definitions of reasoning, and insufficient evaluation methods, MME-Reasoning is guided by the following principles: 
\textbf{\textit{1) Comprehensiveness.}} According to Charles Sanders Peirce's classification of reasoning, deduction, induction, and abduction can be distinguished based on different arrangements of rule, case, and result. Therefore, a comprehensive evaluation of reasoning ability should include all three types of reasoning tasks. 
\textbf{\textit{2) Beyond Perception.}} Each question should be carefully designed to ensure that the answer is obtained through a reasoning process instead of simple visual recognition. 
\textbf{\textit{3) Minimizing Knowledge Reliance.}} It is essential to ensure that the questions do not require complex domain knowledge, thereby preventing models from being penalized for the absence of specialized information. In MME-Reasoning, the domain expertise is limited to K12 or below. 
\textbf{\textit{4) Diverse evaluation formats.}} The benchmark should consist of diverse question types, avoiding incomplete evaluation caused by a narrow range of task types.

\begin{table}[t]
    \centering
	\begin{minipage}{0.43\linewidth}
		\centering
        \caption{Statistics of MME-Reasoning.} 
        \label{tab:statistics}
        \vspace{-2pt}
        \resizebox{1.0\linewidth}{!}{
        \begin{tabular}{l c }

        \toprule
         \textbf{Statistics} & \textbf{Number}  \\
         \cmidrule{1-2}
         \textbf{\textit{Total}} & 1188 (100\%) \\
         ~- Newly-add questions & 84.85\% \\
         ~- Sampled questions & 15.15\% \\
         \cmidrule{1-2}
         \textbf{\textit{Question Type}} & \\
         ~- Multi-choice questions  & 58.50\% \\
         ~- Free-form questions & 31.57\% \\
         ~- Rule-based questions  & 9.93\% \\
         \cmidrule{1-2}
         \textbf{\textit{Image Type}} & \\
         ~- Single-image questions & 58.50\% \\
         ~- Multi-image questions & 31.57\% \\
         \cmidrule{1-2}
         \textbf{\textit{Disciplinary}} \\
         ~- Disciplinary questions & 31.48\% \\
         ~- Non-discipl. questions & 68.52\% \\
        \bottomrule[1pt]
        \end{tabular}

        }
	\end{minipage}
	\hfill
	\begin{minipage}{0.55\linewidth}
	\centering
		\setlength{\abovecaptionskip}{0.1cm}
		\includegraphics[width=\linewidth]{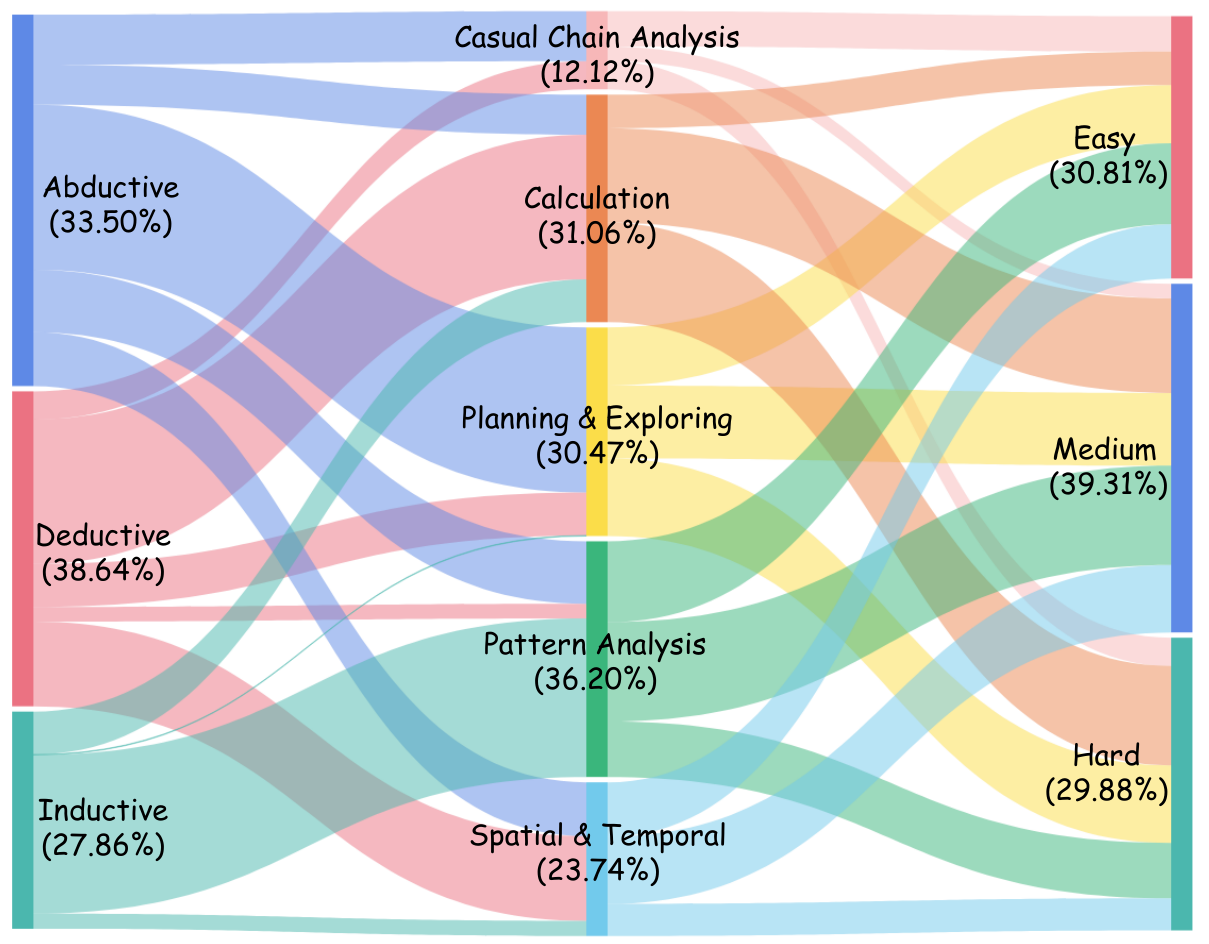}
		\captionof{figure}{Overview of MME-Reasoning.}
		\label{fig:pie_chart}
	\end{minipage}
    \vspace{-5pt}
\end{table}

\subsection{Data Collection and Curation}

\textbf{Data Collection.} We initiate by collecting questions related to multimodal reasoning from a variety of sources, including
\textbf{\textit{1) Textbooks}} can provide subject exam questions (\textit{e.g.}, mathematics, physics, chemistry, and biology). To evaluate reasoning ability, the chemistry and biology questions mainly focus on reaction process inference, and genetic lineage inference. 
\textbf{\textit{2) Online resources, books on logical practice, and Chinese Civil Service Examination (Logic Test)}} primarily includes IQ test questions, logic games (\textit{e.g.}, Mate-in-one), and other tasks highly related to logical reasoning. 
\textbf{\textit{3) Synthetically generated questions.}} Some visual reasoning problems, such as Number Bridge, Sudoku, and mazes, can be generated based on specific rules. We develop code to produce a wide variety of such logic puzzles, covering different types and a range of difficulty levels. 
\textbf{\textit{4) Questions from existing benchmarks.}} We sample 80 questions from PuzzleVQA~\citep{chia2024puzzlevqa} and 100 questions from MMIQ~\citep{cai2025mmiq}, excluding questions based on shape size identification, as such questions may not effectively assess the model’s reasoning ability. 
\textbf{\textit{5) Self-designed questions.}} We mainly construct questions related to spatial and temporal reasoning. The spatial reasoning questions involve tasks such as determining relative spatial relationships and navigation, with the question design methodology inspired by VSIBench~\citep{yang2024vsibench}. For temporal reasoning, the questions mainly focus on sequence judgment. We sample frames from videos in YouCook2~\citep{zhou2018youcook2} and VideoMME~\citep{fu2024video} as the sources of images. 
Note that for questions with well-defined rules such as Number Bridge Puzzles, we include the corresponding rules as part of each question. The composition of MME-Reasoning is shown in Fig.~\ref{fig:pie_chart} and please refer to the Appendix for more details about the question source and type.

\textbf{Data Curation.} We initially collect around 4k questions from various sources mentioned above. Following the design principles of MME-Reasoning, we conduct a careful manual curation process to ensure the quality of the benchmark. Specifically, we exclude questions that depend solely on visual recognition, require complex domain-specific knowledge, too easy to evaluate the reasoning ability. This curation process ensures that the remaining questions are well-aligned with our goal of evaluating visual reasoning ability, rather than perceptual skills or the breadth of specialized knowledge. For questions with multiple possible answers, we first try to convert them into rule-based (will be introduced in Sec.~\ref{sec:evaluation_protocals}) or multiple-choice questions; otherwise, discard them. Additionally, we remove questions that place excessive demands on instruction-following ability. Finally, to comprehensively evaluate the multimodal reasoning ability, we balance the distribution of questions across the three reasoning types. This approach prevents the benchmark from being overly biased towards evaluating the ability of any single reasoning type. Through this data curation process, we filter 1,008 questions from the initially collected questions.

\textbf{Metadata Annotation.} Further, we annotate questions in MME-Reasoning with information including question type (\textit{i.e.}, multiple-choice, free-form, and rule-based), difficulty (\textit{i.e.}, easy, medium, hard), capability (\textit{i.e.}, pattern analysis, planning and exploring, spatial and temporal, calculation, casual chain analysis), and reasoning type (\textit{i.e.}, deductive, inductive, and abductive). For specific rules for annotating metadata, please refer to our appendix.


\subsection{Evaluation Protocols}
\label{sec:evaluation_protocals}

\begin{figure}[t]
    \centering
    \includegraphics[width=\linewidth]{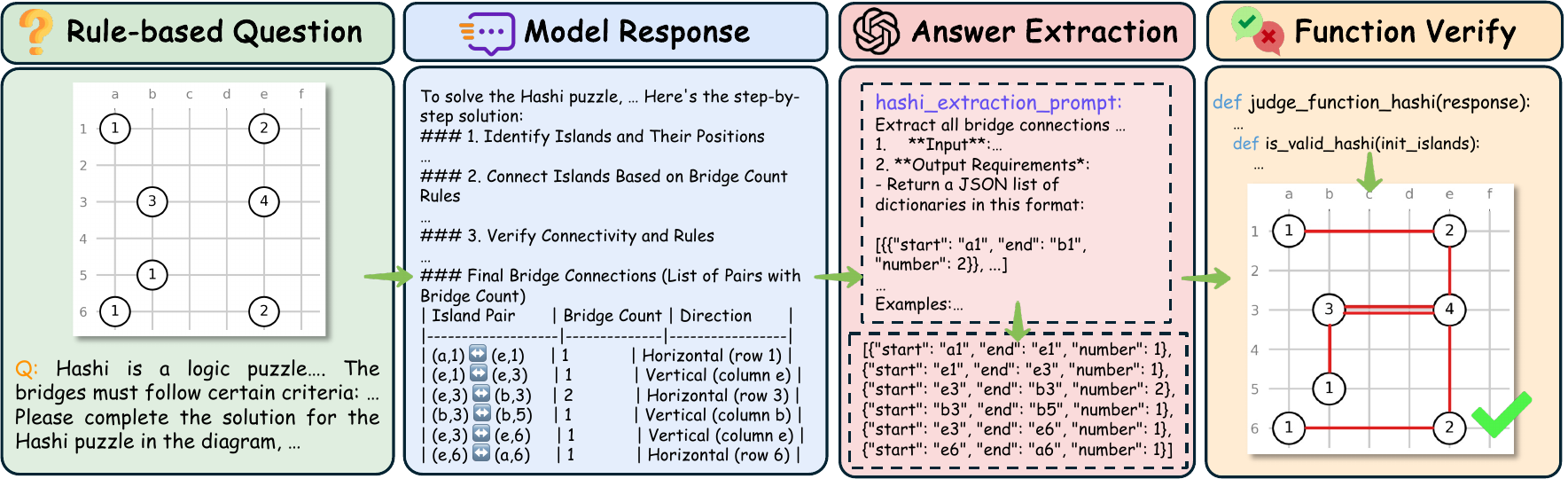}
    \caption{Evaluation of rule-based questions.}
  \vspace{-16pt}
    \label{fig:rule_based_example}
\end{figure}

Following MathVista~\citep{lu2023mathvista}, the evaluation consists of two steps: extracting answers and judging answers. For different types of questions (\textit{i.e.}, multiple-choice, free-form, and rule-based), we designed specific prompts for GPT to extract answers. These prompts are composed of extraction rules and examples that are similar to MathVista~\citep{lu2023mathvista}. For multiple-choice questions, we match the extracted answers with the reference answers. For free-form questions, we use GPT to judge the consistency between the extracted answers and the reference answers following MathVerse~\citep{zhang2024mathverse}. For rule-based questions, we first use GPT to extract answers and convert them into an intermediate format, which is then judged using specific scripts. For example, in a Number Bridge problem, we first use GPT to extract the start and end points of each bridge, then convert the answers into a specific matrix format, and finally determine correctness based on predefined rules, as illustrated in Fig.~\ref{fig:rule_based_example}.

%% file: Sections/4_exp.tex
\section{Experiments}

\subsection{Experimental Settings}
We conduct extensive evaluations on state-of-the-art MLLMs include:

\textbf{Thinking Models.} We first evaluate several thinking MLLMs that focus on improving the models' multimodal reasoning which can be divided into \textbf{\textit{Close-source models}} including (1) GPT-o1~\citep{o1}, and o4-mini~\citep{o4-mini}; (2) Gemini-2.5-Flash-Thinking and Gemini-2.5-Pro-Thinking~\citep{deepmind_gemini_report}; (3) Claude-3.7-Sonnet-Thinking, Claude-4-Sonnet-Thinking~\citep{Claude}; (4) Seed1.5-VL-Thinking~\citep{guo2025seed1}; and \textbf{\textit{Open-source models}} including (1) QvQ-72B-Preview~\citep{qvq-72b-preview}; (2) Kimi-VL-A3B-Thinking~\citep{team2025kimi-vl}; (3)LlamaV-o1~\citep{thawakar2025llamav}; (4) Virgo-72B~\citep{du2025virgo}.

\textbf{Chat Models.} Further, we also evaluate SoTA chat models as follows. \textbf{\textit{Close-source models}}: (1) GPT-4o~\citep{gpt4o}; (2) Claude-3.7-Sonnet~\citep{Claude} (3) Kimi-latest~\citep{team2025kimi}; (4) Seed1.5-VL~\citep{guo2025seed1}. \textbf{\textit{Open-source models}}: (1) Qwen-2.5-VL (7B, 32B, 72B)~\citep{Qwen2.5-VL}; (2) InternVL-3 (8B, 38B, 78B)~\citep{zhu2025internvl3}; (3) LLaVA-Onevision-72B~\citep{li2024llavaov}; (4) Molmo (7B-O, 7B-D, 72B)~\citep{deitke2024molmo}; (5) Kimi-VL-A3B-Instruct~\citep{team2025kimi-vl}. 

\textbf{Rule based RL Models.} Rule-based Reinforcement Learning (RL) has been shown to be a highly promising strategy for eliciting reasoning paradigms in models. Therefore, we further evaluated MLLMs trained using Rule-based RL, including: (1) R1-VL~\citep{zhang2025r1-vl}, (2) R1-Onevision~\citep{yang2025r1-onevision}, 
(3) Vision-R1~\citep{huang2025vision-r1}, (4) MM-Eureka (7B, 32B)~\citep{meng2025mmeureka}, (5) VL-Rethinker (7B, 72B)~\citep{wang2025vl-rethinker}.

We use GPT-4o-mini to extract answers from model responses. Due to rate limits, we sample 302 questions to construct mini-set with the same distribution for o1's evaluation, all other models are evaluated on the entire benchmark.

\subsection{Main Results}
Tab.~\ref{tab:main_result} shows the performance comparison of different MLLMs and prompting strategies.

\input{Tables/main_results}

\textbf{MME-Reasoning poses significant challenges for vision-language reasoning.}
The best-performing model, Gemini-2.5-Pro-Thinking, achieved an average score of 60.2\%.
The latest MLLM, Seed1.5-VL, achieved a comprehensive score of 59.9.
Representative reasoning models o4-mini and o1 obtained scores of 57.5 and 45.7, respectively.
Qwen2.5-VL and Claude-3.7-Sonnet achieved scores of 35.9 and 57.2 on OlympiadBench, yet only reached 34.1 on MME-Reasoning.
These results indicate that the benchmark sets stringent standards for evaluating models' logical reasoning capabilities by comprehensively assessing three distinct reasoning types. 

\textbf{Prominent bias in logical reasoning performance within MLLMs.} In almost all cases, models exhibit dominant deductive reasoning performance, while abductive reasoning is considerably weaker.
Closed-source models demonstrate an average deductive advantage of 5.38 over abductive reasoning, which widens to 9.81 among open-source models, making abductive reasoning a significant bottleneck in comprehensive logical reasoning performance.
Deductive reasoning maintains a high proportion in the training corpus due to its widespread distribution.
Abductive reasoning processes usually involve larger exploration spaces and richer assumptions, hypotheses, and reflections, making its data challenging to scale.
However, non-deductive reasoning plays a central role in general reasoning scenarios and many scientific discoveries. These findings highlight the necessity for researchers to develop a more comprehensive understanding of models' logical reasoning abilities to facilitate their application in real-world scenarios.
Moreover, the models' scores under different reasoning types are typically score below 40, indicating that MME-Reasoning provides a promising metric for evaluating reasoning capabilities from multiple perspectives.

\begin{figure}[t]
\vspace{-8pt}
  \centering
  \includegraphics[width=0.95\linewidth]{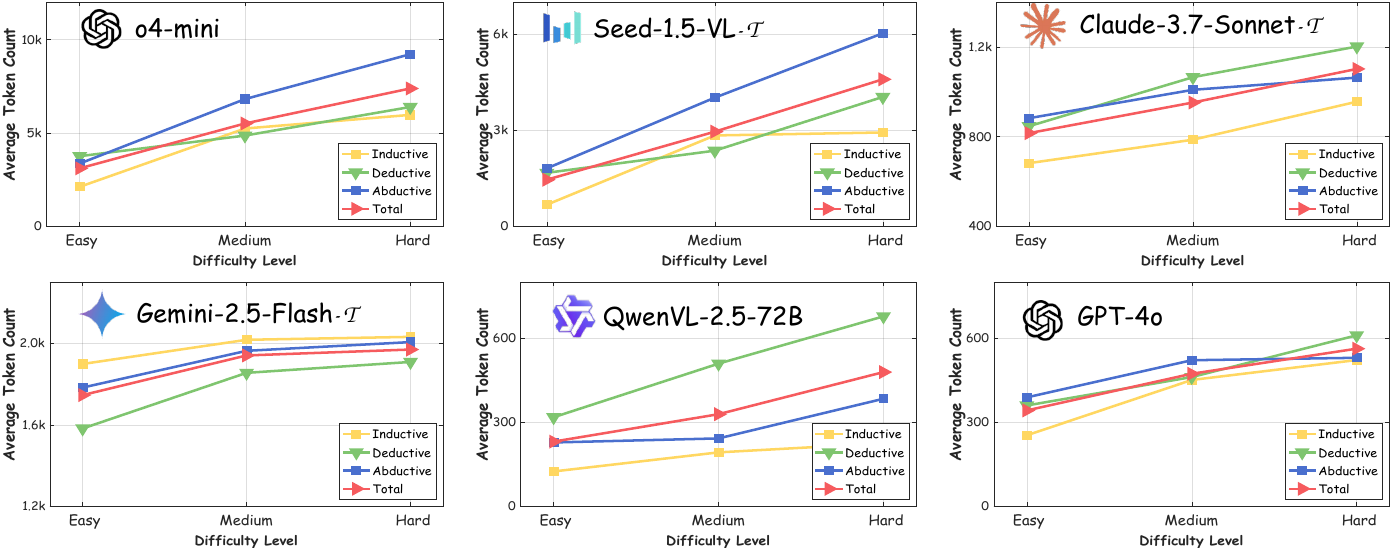}
  \vspace{-2pt}
  \caption{Comparison of Difficulty Level and Average Token Count on MME-Reasoning.}
  \vspace{-12pt}
  \label{fig:difficulty_tokens}
\end{figure}

\begin{figure}[t]
\vspace{-2pt}
    \centering
	\begin{minipage}{0.54\linewidth}
		\centering
        \vspace{5pt}
        \setlength{\abovecaptionskip}{0.10cm}
		\includegraphics[width=0.90\linewidth]{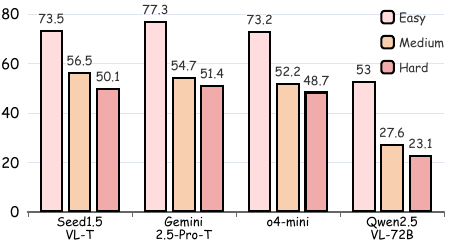}
		\caption{Results within different difficulty levels.}
        
		\label{fig:diff_acc}
	\end{minipage}
	\hfill
	\begin{minipage}{0.45\linewidth}
	\centering
		\setlength{\abovecaptionskip}{0.1cm}
		\includegraphics[width=0.90\linewidth]{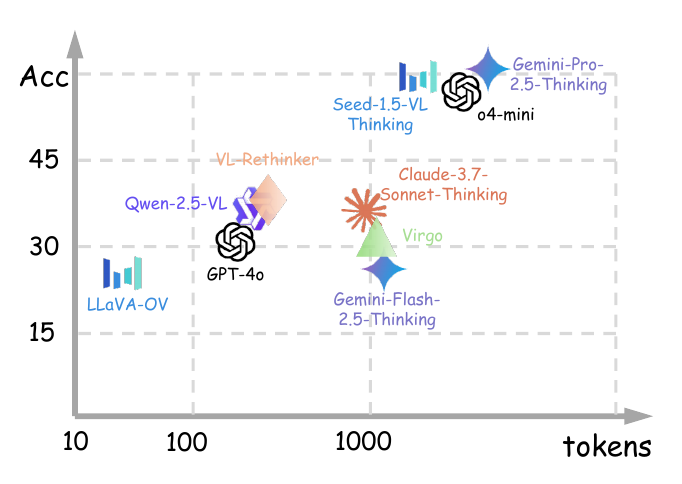}
        \vspace{-6pt}
		\caption{Response tokens vs. Performance.}
		\label{fig:token_acc}
	\end{minipage}
\end{figure}

\textbf{Limited performance in open-ended reasoning scenarios.} Models generally demonstrate relative advantages in Casual Chain Analysis but perform poorly on tasks involving Plan \& Exploration.
This may benefit from the autoregressive paradigm continuously aiding models in learning causal dependencies within input sequences. However, it also highlights a critical shortcoming: current state-of-the-art models struggle with planning and exploration in open-ended problem-solving spaces.
To advance models in solving difficult practical problems, it is critical to innovate learning paradigms and strategies generation mechanisms suitable for open scenarios.

\textbf{Thinking capability directly contributes to enhanced logical reasoning.} Models employing "thinking mode" typically generalize test-time scaling to reasoning scenarios through generating longer chains-of-thought (CoT), reflections, and self-corrections. 
In most cases, "thinking models" significantly outperform their base version. QvQ improved by 1.1 compared to Qwen2.5-VL, and VL-Rethinker improved by 1.7 compared to Qwen2.5-VL. 
This effect is more pronounced among closed-source models: Seed1.5-VL-T outperformed Seed1.5-VL by 12.4, and o1 exceeded GPT-4o by 15.5. 
Further experiments concerning thinking models will be elaborated in subsequent sections.

\textbf{Rule-based RL does not always work.} Rule-based RL has shown significant potential in activating the "thinking mode" of foundational models, encouraging longer output and reflection to tackle hard problems. 
However, we observed that methods adopting rule-based RL do not consistently outperform their base models. 
Most models at the 7B scale experienced performance degradation.
This suggests that the potential of rule-based RL remains inadequately realized, failing to effectively extend advantages demonstrated in LLMs into multimodal domains and possibly reducing generalization.
Thus, innovation in training paradigms, rather than merely replicating R1, is urgently needed.

\subsection{Fine Grained Analysis of Reasoning Behavior}

\textbf{Does increasing the length of the reasoning process help?} To investigate whether increased output length consistently leads to improved accuracy, we selected 10 representative models, including Chat Models (\textit{e.g.}, GPT-4o) and Thinking Models (\textit{e.g.}, o4-mini). In Fig.~\ref{fig:token_acc}, we present the semi-log plot of average token count (ATC) versus accuracy.
The overall trend reveals that models with longer outputs tend to achieve higher scores, indicating the effectiveness of extending the reasoning process to enhance logical reasoning performance.
As the token number increases, model performance exhibits a exponential growth pattern, suggesting diminishing returns from simply increasing output length.
Compared to Thinking Models, Chat Models demonstrate higher token efficiency.
These findings highlight the computational cost associated with scaling up inference for improved performance. 
Balancing reasoning efficiency and model effectiveness remains a challenge for future research.

\begin{figure}[t]
\vspace{-6pt}
  \centering
  \includegraphics[width=0.98\linewidth]{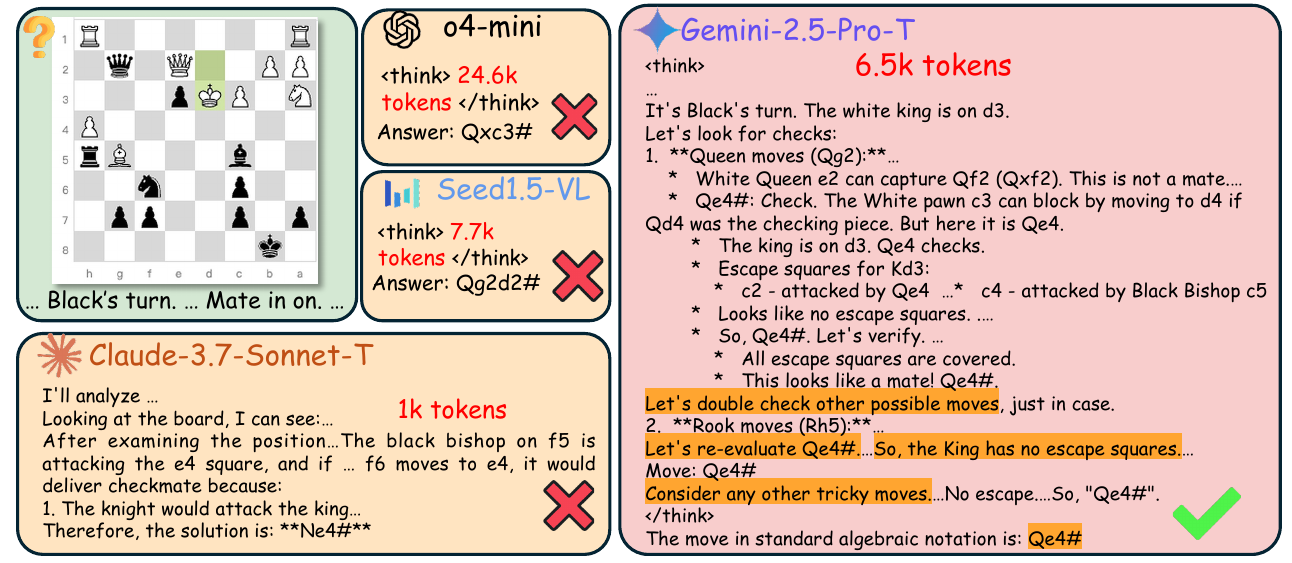}
  \vspace{-4pt}
  \caption{Case study of a Mate-in-one problem.}
  \vspace{-8pt}
  \label{fig:case_study}
\end{figure}

\textbf{Is the length of the reasoning process strongly correlated with task difficulty?} To examine whether models spontaneously allocate more inference budget to more challenging questions, we conducted research on using representative Thinking Models such as o4-mini and Chat Models such as GPT-4o. We first analyzed the accuracy of different models across varying levels of difficulty, as shown in Fig.~\ref{fig:diff_acc}. With increasing difficulty, model performance declines significantly, confirming the validity of MME-Reasoning's difficulty stratification and providing a foundation for subsequent analyses. Besides, Fig.~\ref{fig:difficulty_tokens} illustrates the trend of ATC across different reasoning types and difficulty levels. 
It reveals a consistent pattern: overall, output length increases steadily with rising difficulty. This trend holds across varying output lengths, model categories, and reasoning types.
Compared to Chat Models, Thinking Models exhibit a more pronounced increase in ATC as difficulty rises. For instance, the ATC of Seed1.5-VL increases by up to 3k tokens, and o4-mini by up to 5k tokens. In contrast, the ATC increase for Qwen2.5-VL and GPT-4o remains within 300 tokens.


\subsection{Case Study}
In Fig.~\ref{fig:case_study}, we present an example of abductive reasoning which demands planning and exploration. From this case, several key observations can be identified:
\textit{\textbf{(1)Long reasoning process}}: The selected models generated over 1k tokens in response, with o4-mini producing up to 24.6k tokens. This demonstrates that MME-Reasoning constitutes a highly challenging benchmark for multimodal reasoning.
\textit{\textbf{(2)Planning in the problem-solving process}}: The response includes multiple iterations of \textit{``hypothesis generation (possible movement) – feasibility verification (check escape squares) – check ”}, indicating that the model spontaneously engages in structured planning and reflection to explore solutions within an open-ended problem-solving spaces.
\textit{\textbf{(3)Repetitive reflection}}: We observed that the model tends to revisit and reflect on the same reasoning paths multiple times—up to 7 instances in some cases. This behavior may result in significant computational overhead and informational redundancy. Balancing reasoning efficiency with performance remains a critical issue to be addressed.

%% file: Tables/main_results.tex
\begin{table}[t!]
\vspace{-16pt}
\caption{Performance comparison of state-of-the-art MLLMs on MME-Reasoning. The top three are highlighted in \colorbox{blue!30}{blue}. $\dagger$ indicates the model was evaluated on the mini-set. ``T" represents ``Thinking".}
\vspace{1pt}
\label{tab:main_result}
\centering
\small
\resizebox{1.0\linewidth}{!}{
\begin{tabular}{l c c c c c c c c c}
\toprule
\multirow{2}{*}{\textbf{Model}}  & \multicolumn{5}{c}{\textbf{Model Capability}} & \multicolumn{3}{c}{\textbf{Reasoning Type}} & \multirow{2}{*}{\textbf{AVG.}}  \\
\cmidrule(lr){2-6} \cmidrule(lr){7-9}
& CAL. & P\& E. & PA. & S\&T. & CCA. & DED. & IND. & ABD.  & \\
\cmidrule{1-10}
\multicolumn{10}{c}{\textbf{\textit{Close-source \& Thinking}}} \\
\noalign{\vspace{2pt}} 
\hdashline
\noalign{\vspace{2pt}}
\rowcolor{blue!30}
Gemini-2.5-Pro-T & \textbf{68.0} & \textbf{64.4} & 53.7 & \textbf{52.1} & \textbf{90.3} & 64.0 & 51.7 & \textbf{62.8} & \textbf{60.2} \\
\rowcolor{blue!20}
Seed1.5-VL-T & 67.2 & 62.7 & 56.0 & 47.2 & 82.6 & \textbf{64.5} & \textbf{52.3} & 60.8 & 59.9 \\
\rowcolor{blue!10}
o4-mini & 63.1 & 58.3 & \textbf{57.2} & \textbf{50.4} & 59.0 & 60.6 & 51.4 & 59.0 & 57.5 \\
o1$^\dagger$ & 50.0 & 38.5 & 41.5 & 43.7 & 52.4 & 50.8 & 42.3 & 42.3 & 45.7 \\
Claude-4-Sonnet-T & 33.3 & 35.9 & 33.0 & 36.2 & 47.9 & 39.4 & 32.0 & 35.7 & 36.1 \\
Claude-3.7-Sonnet-T & 30.4 & 27.6 & 32.3 & 38.3 & 46.5 & 34.6 & 36.2 & 31.7 & 34.1 \\
Gemini-2.5-Flash-T & 19.8 & 21.3 & 20.9 & 33.0 & 38.9 & 28.1 & 22.1 & 24.6 & 25.2 \\
\cmidrule{1-10}
\multicolumn{10}{c}{\textbf{\textit{Close-source \& Chat}}} \\
\noalign{\vspace{2pt}} 
\hdashline
\noalign{\vspace{2pt}}
Seed1.5-VL & 52.0 & 42.0 & 38.4 & 44.0 & 72.9 & 54.9 & 45.0 & 41.0 & 47.5 \\
GPT-4o & 21.4 & 22.1 & 30.5 & 38.6 & 36.8 & 29.0 & 34.7 & 27.9 & 30.2 \\
Claude-3.7-Sonnet & 29.0 & 24.6 & 32.8 & 35.5 & 46.5  & 35.7 & 38.7 & 26.1 & 33.3 \\
Kimi-Latest & 21.4 & 17.4 & 19.8 & 29.1 & 41.0 & 27.7 & 25.4 & 19.9 & 24.4 \\

\cmidrule{1-10}
\multicolumn{10}{c}{\textbf{\textit{Open-source \& Thinking}}} \\
\noalign{\vspace{2pt}} 
\hdashline
\noalign{\vspace{2pt}}
QVQ-72B-Preview & 37.4 & 27.1 & 28.8 & 35.8 & 57.6 & 41.6 & 33.5 & 29.1 & 35.2 \\
Virgo-72B & 30.4 & 22.9 & 26.1 & 36.2 & 47.2 & 37.7 & 32.6 & 24.4 & 31.8 \\
VL-Rethinker-72B & 33.6 & 28.4 & 31.4 & 37.2 & 59.7 & 39.0 & 36.0 & 31.9 & 35.8 \\
VL-Rethinker-7B & 24.7 & 17.7 & 23.5 & 39.4 & 42.4 & 34.4 & 29.9 & 22.9 & 29.3 \\
MM-Eureka-Qwen-32B & 23.0 & 25.7 & 25.6 & 36.2 & 50.7 & 32.9 & 30.5 & 28.1 & 30.6 \\
MM-Eureka-Qwen-7B & 27.1 & 19.3 & 22.3 & 31.9 & 50.0 & 32.7 & 28.7 & 22.6 & 28.2 \\
R1-VL-7B & 16.3 & 11.6 & 17.7 & 30.9 & 26.4 & 25.3 & 21.8 & 15.8 & 21.1 \\
Vision-R1-7B & 18.2 & 18.0 & 17.9 & 34.4 & 36.1 & 27.4 & 26.3 & 18.1 & 24.0 \\
R1-Onevision-7B-RL & 19.5 & 12.2 & 20.0 & 31.6 & 27.1 & 27.7 & 24.8 & 14.6 & 22.5 \\
Kimi-VL-A3B-T & 28.7 & 16.0 & 19.5 & 32.3 & 35.4 & 33.3 & 25.1 & 18.1 & 25.9 \\
\cmidrule{1-10}
\multicolumn{10}{c}{\textbf{\textit{Open-source \& Chat}}} \\
\noalign{\vspace{2pt}} 
\hdashline
\noalign{\vspace{2pt}}
Qwen2.5-VL-72B & 31.7 & 25.1 & 27.2 & 37.9 & 53.5 & 39.0 & 32.3 & 29.9 & 34.1 \\
Qwen2.5-VL-32B & 32.2 & 26.8 & 24.4 & 39.0 & 52.1 & 40.5 & 27.5 & 29.6 & 33.2 \\
Qwen2.5-VL-7B & 22.2 & 18.2 & 21.9 & 35.1 & 36.1 & 31.4 & 27.5 & 20.9 & 26.8 \\
InternVL3-78B & 26.0 & 24.0 & 26.5 & 41.8 & 50.0 & 35.1 & 33.8 & 27.1 & 32.1 \\
InternVL3-38B & 23.0 & 18.5 & 23.0 & 38.3 & 41.7 & 33.5 & 29.0 & 22.1 & 28.4 \\
InternVL3-8B & 19.5 & 19.6 & 22.6 & 31.6 & 41.0 & 28.1 & 29.9 & 21.4 & 26.4 \\
Molmo-72B & 12.5 & 11.9 & 14.7 & 28.7 & 28.5 & 23.1 & 18.4 & 14.3 & 18.9 \\
Molmo-7B-D & 11.7 & 8.6 & 8.1 & 27.3 & 23.6 & 20.7 & 10.9 & 11.1 & 14.7 \\
LLaVA-OV-72B & 17.1 & 18.0 & 23.9 & 32.3 & 38.9 & 27.4 & 30.5 & 19.9 & 25.8 \\
Kimi-VL-A3B & 18.7 & 11.9 & 21.4 & 34.0 & 27.8 & 25.9 & 26.3 & 17.1 & 23.1 \\
\bottomrule[1pt]
\end{tabular}
}
\vspace{-10pt}
\end{table}

%% file: Sections/5_conclusion.tex
\section{Conclusion}

We introduce MME-Reasoning, a comprehensive benchmark designed to evaluate MLLMs' logical reasoning abilities across inductive, deductive, and abductive reasoning types. Through careful data curation and an expanded evaluation protocol, our benchmark provides a holistic assessment of reasoning capabilities, beyond simple perception or high-level knowledge. Our experiments reveal that existing MLLMs still face significant challenges and exhibit notable performance imbalances across different reasoning types. These findings underscore the need for further research and development to enhance the reasoning abilities of MLLMs, paving the way for more generalizable AI systems.

%% file: Sections/6_appendix.tex
\appendix

\section*{Technical Appendices and Supplementary Material for MME-Reasoning}

\definecolor{c1}{HTML}{ad2933}

{
    \hypersetup{linkcolor=c1}
    \DoToC
}

\newpage
\section{More Experimental Results}
\label{sec:exp_res}

\subsection{Full Results on MME-Reasoning}
\label{subsec:res_full}
\input{Tables/main_results_full}
We present the performance of more baselines on MME-Reasoning in Tab~\ref{tab:main_result_full}, including OpenVLThinker~\citep{deng2025openvlthinker}, LMM-R1-MGT-PerceReason~\citep{peng2025lmm}, Mulberry~\citep{yao2024mulberry}, LlamaV-o1~\citep{thawakar2025llamav} and Qwen2-VL series~\citep{wang2024qwen2}.

\subsection{Full Results on Mini-set of MME-Reasoning}
\label{subsec:res_mini}

\input{Tables/mini_full}

\input{Tables/main_results_mini}

We randomly sampled 25\% of the questions and conducted manual review to ensure that the diversity of image types was maintained. These sampled questions were then used to to construct the Mini-set. 
We also analyzed the question distributions of both the Mini-set and the Full-set to ensure the sampled questions retained the same distribution. The statistical results are presented in Tab~\ref{tab:mini_full_stat}.

We provide the performance of all baseline models on the Mini-set in Tab.~\ref{tab:main_result_full_mini}. All baseline models achieved similar performance on both the Full-set and the Mini-set, further demonstrating the consistency of Mini-set and the comparability of model performance across different splits.

\subsection{Human Performance}
\label{subsec:res_human}
To evaluate expert-level performance on MME-Reasoning, we further report human performance on the mini-set of MME-Reasoning. 
As shown in Tab.~\ref{tab:main_result_full_mini}, the human expert achieved an overall score of 83.4—significantly outperforming the best-performing thinking model, Seed1.5-VL-T, which scored 62.6.
Looking deeper into the reasoning types, the human expert scored 85.8, 76.9, and 85.6 on deductive, inductive, and abductive reasoning respectively, all of which are notably higher than the scores of the best-performing model. Moreover, the human expert demonstrated a particularly strong ability in abductive reasoning, with performance comparable to that in deductive reasoning—which is the key focus in current multi-modal reasoning research.
This strength aligns with a few top-performing models, but stands in contrast to most baseline models, which show clear weaknesses in abductive reasoning. These results highlight the significant gap that still exists between current thinking \& chat models and human-level performance in comprehensive multimodal reasoning evaluation.
Expanding complex reasoning tasks beyond domain-specific knowledge questions to include a broader range of reasoning types and more diverse tasks will be a crucial step toward addressing these current limitations.

\subsection{Results on Different Question Types}
\label{subsec:res_question_type}
\input{Tables/main_results_question_type}
We also evaluated the model's performance across different question types and present the results in Tab.~\ref{tab:main_result_question_type}.

\subsection{Results with Test-Time Compute Scaling}
\input{Tables/results_mcts}
\label{subsec:res_tts}
To evaluate whether the use of Test-Time Compute Scaling (TTS) methods can improve model performance on MME-Reasoning, we take Qwen2.5-VL-7B as an example and use Qwen2.5-VL-32B as the Reward Model. The evaluation is conducted using the Monte Carlo Tree Search (MCTS) algorithm, with the settings: \textit{branch = 3} and \textit{max-iteration = 18}.
The results are shown in Table~\ref{tab:mcts_results}. 

Under the MCTS-based setting, the model's performance dropped noticeably across all reasoning types. We attribute this decline to two main factors:
(1) Questions in MME-Reasoning often involve complex parallel reasoning, hypothesis generation, and reflection, rather than simple linear logical progression. These characteristics may not be effectively captured by the Reward Model.
(2) The limited capabilities of the Reward Model result in guidance that lacks practical utility.

We leave further exploration of TTS methods for reasoning to future work and hope that MME-Reasoning can serve as a representative benchmark for developing more general and comprehensive TTS algorithms in reasoning tasks.

\subsection{Results with CoT Prompt}
\label{subsec:res_cot}
\input{Tables/results_cot}

\begin{table}[h]
\centering
\renewcommand{\arraystretch}{1.3}
\begin{tabularx}{\textwidth}{@{}>{\bfseries}l X@{}}
\toprule
Model & CoT Prompt \\
\midrule
Qwen2.5-VL & Let's think step by step. \\
InternVL3 &
Answer the preceding question. The last line of your response should follow this format: 'Answer: \$FINAL\_ANSWER' (without quotes), where 'FINAL\_ANSWER' is your conclusion based on the reasoning provided. If you are uncertain or the problem is too complex, make a reasoned guess based on the information provided. Avoid repeating steps indefinitely---provide your best guess even if unsure. Think step by step logically, considering all relevant information before answering.
 \\
\bottomrule
\end{tabularx}
\caption{Chain-of-Thought Prompts for Different Models}
\label{tab:cot_prompt}
\end{table}

Chain-of-Thought (CoT) prompting increases output length by encouraging explicit output of the thought process, thereby enhancing reasoning performance. 
To investigate the impact of CoT on performance in MME-Reasoning, we evaluated the Qwen2.5-VL and InternVL3 series using CoT prompts shown in Tab.~\ref{tab:cot_prompt}. The results are presented in Tab.~\ref{tab:cot_results}.

We observed that the Qwen2.5-VL models naturally tend to generate their reasoning process, so adding a CoT prompt did not significantly increase output length. In contrast, InternVL3 models, under default settings, tend to directly output the final answer, and the CoT prompt substantially increased output length.

In terms of performance, adding the CoT prompt consistently led to performance degradation for the Qwen2.5-VL series. For InternVL3, performance dropped for the 7B model but improved for the larger 38B and 78B models. One possible hypothesis is that for models already inclined to produce long outputs, explicit CoT instructions might introduce noise into the reasoning process. Conversely, for models that tend to answer questions directly, smaller models struggle to produce helpful and correct CoT outputs, but as model size increases, they begin to benefit noticeably from relatively accurate reasoning processes.

\subsection{Token Usage of Thinking Models}

\begin{figure}[t]
  \centering
  \includegraphics[width=0.98\linewidth]{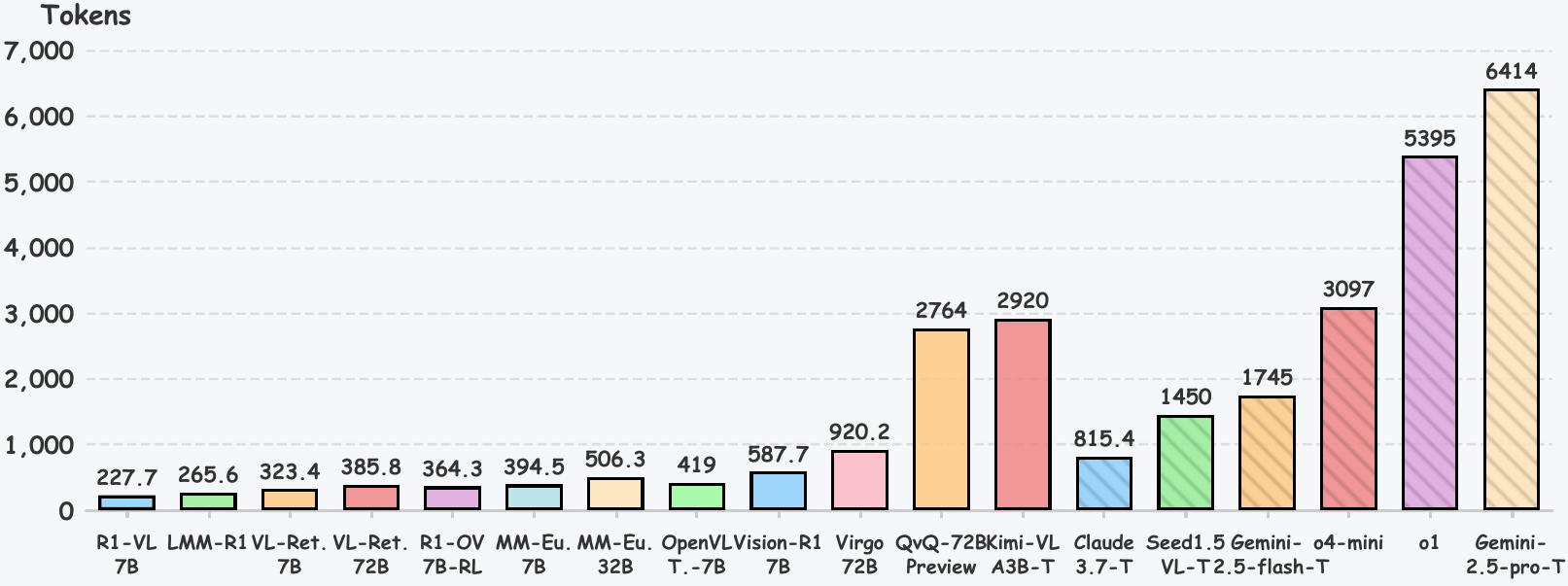}
  \caption{Average token usage of open \& closed-source thinking models on MME-Reasoning.}
  \label{fig:output_length}
\end{figure}

In Fig.~\ref{fig:output_length}, we present the average token length of different thinking models on MME-Reasoning. Overall, there is a clear trend indicating that better model performance is often associated with longer reasoning paths. However, we also observe diminishing returns between output length and performance in both open-source and closed-source models.

Additionally, although current rule-based reinforcement learning (RL) models show a promising trend of increased output length during training, no significant length gains were observed on MME-Reasoning. This limitation may stem from the limited types and inappropriate complexity of the reasoning tasks. Therefore, exploring how different types of reasoning tasks can better stimulate the effectiveness of RL in reasoning may be a valuable direction for future research.

\subsection{Results of Captioner \& LLMs}
\label{subsec:res_cap}

\input{Tables/results_captioner_LLM}

We used GPT-4o as the captioner to generate visual descriptions for each question as a substitute for the images. Then we evaluated existing LLMs with "thinking mode," and the results are presented in Tab.~\ref{tab:caption_results}.
As shown in the results, even when only indirectly perceiving image content through textual descriptions,  QwQ~\citep{qwq32b} and R1~\citep{deepseekai2025deepseekr1incentivizingreasoningcapability} achieved impressive scores of 41.9 and 46.9 respectively—surpassing even Claude-3.7-Sonnet-Thinking.
These findings indicate that there is still substantial room for improvement in extending long-term reasoning capabilities from LLMs to the multimodal domain. This gap may be due, in part, to degradation in the foundational model’s capabilities during the vision-language alignment process. Additionally, the diversity of reasoning tasks specific to multimodal settings has yet to be thoroughly explored.

\section{Details of Annotation}
\label{sec:details_anno}

\subsection{Difficult Annotation}
For each question, we assign a difficulty label: \textbf{\textit{Easy}}, \textbf{\textit{Medium}}, or \textbf{\textit{Hard}}, based on the cognitive load required to solve it. The labeling criteria are as follows:
\begin{itemize}
    \item \textbf{\textit{Easy:}} The question typically has a straightforward and quick solution that can be correctly answered by a human expert within 2 minutes.
    \item \textbf{\textit{Medium:}} The question generally requires some reasoning steps and one to two rounds of trial and reflection, and can be correctly answered by a human expert within 2 to 5 minutes.
    \item \textbf{\textit{Hard:}} The question usually requires more than two attempts and reflections, or involves the use of tools such as auxiliary lines or drafts to support the thought process. It may or may not be solved by a human expert within 10 minutes.
\end{itemize}

\subsection{Reasoning Type Annotation}
For each question, we assign a reasoning type label: \textbf{\textit{Deductive}}, \textbf{\textit{Inductive}}, or \textbf{\textit{Abductive}}, based on the dominant reasoning method required in its solution. The labeling criteria are as follows:
\begin{itemize}
\item \textbf{\textit{Deductive:}} Involves deriving a necessary conclusion from given premises and general rules through step-by-step inference. Examples include math problems, physics problems, and certain puzzles.
\item \textbf{\textit{Inductive:}} Involves observing specific phenomena, summarizing general patterns or rules, and extrapolating based on those patterns. Examples include figure series and analogy questions.
\item \textbf{\textit{Abductive:}} Involves forming hypotheses or explanations based on known phenomena and then verifying them. These problems typically have a large solution space. Examples include Sudoku, mate-in-one chess problems, circuit fault analysis, biological pedigree analysis, and some puzzles.
\end{itemize}

It should be noted that although the solutions to some puzzles, such as Sudoku, can theoretically be derived through deductive reasoning, in the actual process of human reasoning, we often resort to assuming a certain move and then verifying its validity. This hypothesis–verification–backtracking mechanism leads us to consider these a form of abductive reasoning.

\subsection{Capability Annotation}
For each question, we also assign one or more capability labels based on the primary abilities being tested. The available labels are: \textbf{\textit{Pattern Analysis}}, \textbf{\textit{Planning and Exploring}}, \textbf{\textit{Spatial and Temporal}}, \textbf{\textit{Calculation}}, and \textbf{\textit{Causal Chain Analysis}}. A question may have multiple capability labels. The labeling criteria are as follows:
\begin{itemize}
\item \textbf{\textit{Pattern analysis:}} Requires identifying patterns in shape, color, size, or other visual features within the image.
\item \textbf{\textit{Planning and exploring:}} Requires explicit planning of the answering process, involving exploration within solution space and iterative verification or reflection.
\item \textbf{\textit{Spatial and Temporal:}} Requires understanding spatial relationships or temporal sequences represented in the visual input.
\item \textbf{\textit{Calculation:}} Involves performing numerical calculations based on given quantitative conditions to arrive at a correct result.
\item \textbf{\textit{Causal Chain Analysis:}} Requires reasoning about causal relationships across multiple nodes based on limited information, or understanding dynamic processes in the problem and identifying key events.
\end{itemize}

\section{Details of Implementation}

Some of the data in MME-Reasoning are sourced from ScanNet~\citep{dai2017scannet}, Arkitscenes~\citep{baruch2021arkitscenes}, VideoMME~\citep{fu2024video}, MM-IQ~\citep{cai2025mmiq}, PuzzleVQA~\citep{chia2024puzzlevqa}. We further filter most of the data and reformulate the questions. We use gpt-4o-mini to extract the answer of all responses and judge the answer of free-form questions. The cost fluctuates with the length of the MLLM's response. As an example, extracting and judging the response of Qwen2.5-VL-72B costs around \$0.1. We use VLMEvalKit\footnote{\url{https://github.com/open-compass/VLMEvalKit}} to evaluate all the models. For models larger than 30B, we use vllm\footnote{\url{https://github.com/vllm-project/vllm}} to reduce the inference time. All experiments are conducted on A100 GPUs except experiments on closed-source models.

\section{Details of Evaluation}

\subsection{Prompts for Answer Extraction}
\label{subsec:detail_ans_ext}

\begin{figure}[htbp]
  \centering
  \includegraphics[width=0.9\linewidth]{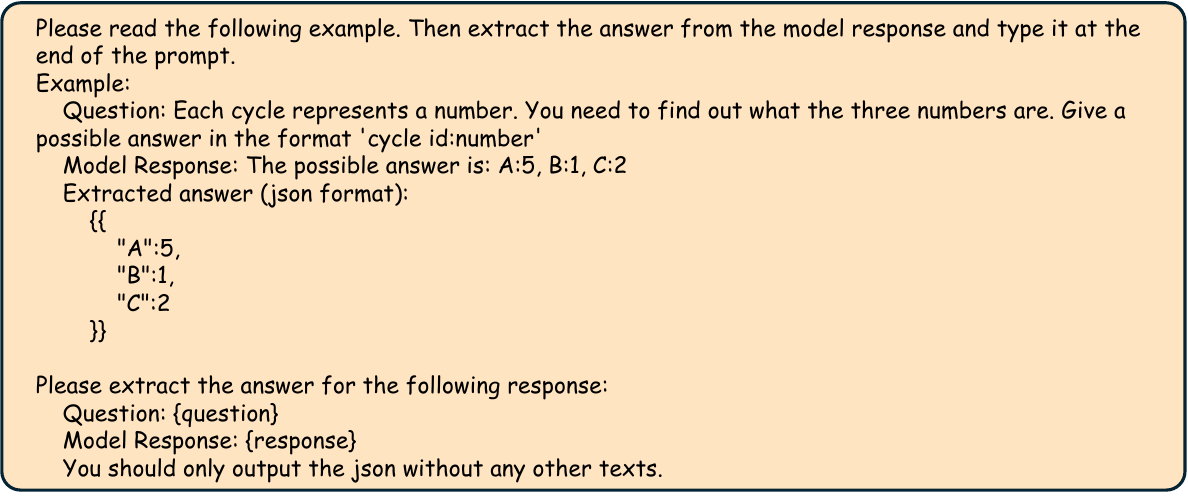}
  \vspace{-5pt}
  \caption{Prompt for tasks answering in `id : answer' format.}
  \label{prompt:id_answer_pair}
  \vspace{-8pt}
\end{figure}

\begin{figure}[htbp]
  \centering
    \includegraphics[width=0.9\linewidth]{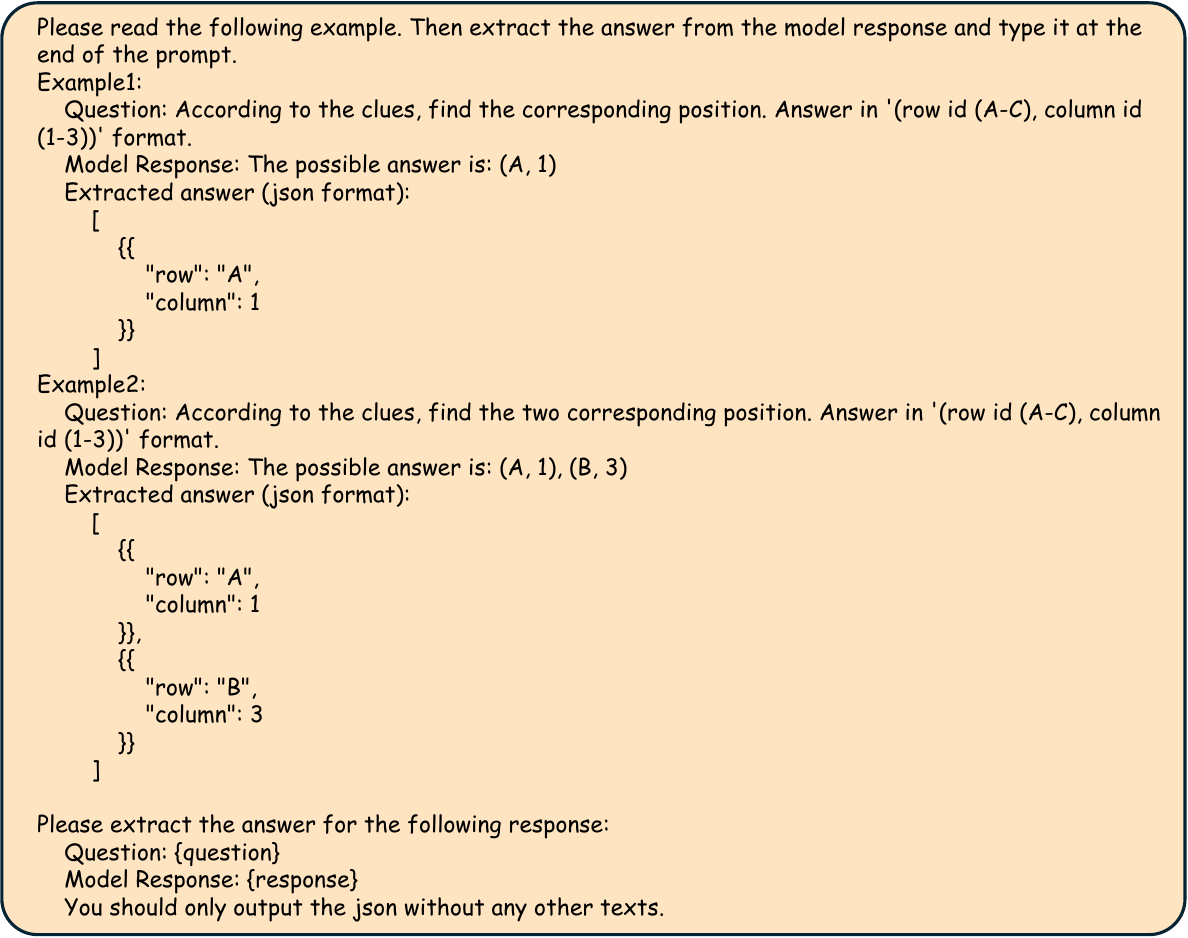}
    \vspace{-5pt}
  \caption{Prompt for tasks answering in `coordinates' format. }
  \label{prompt:coordinate}
\end{figure}

\begin{figure}[htbp]
  \centering
    \includegraphics[width=0.9\linewidth]{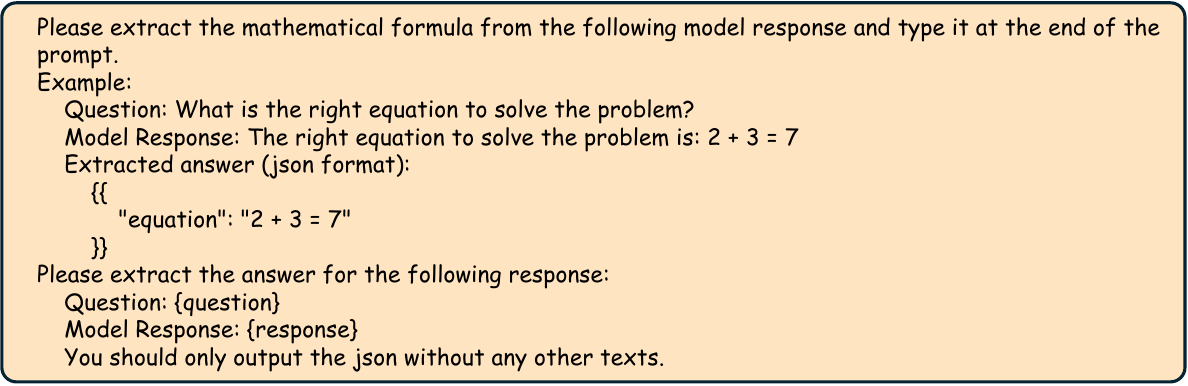}
    \vspace{-5pt}
  \caption{Prompt for tasks answering in `formula' format.}
  \label{prompt:formula}
  \vspace{-5pt}
\end{figure}

\begin{figure}[htbp]
  \centering
    \includegraphics[width=0.9\linewidth]{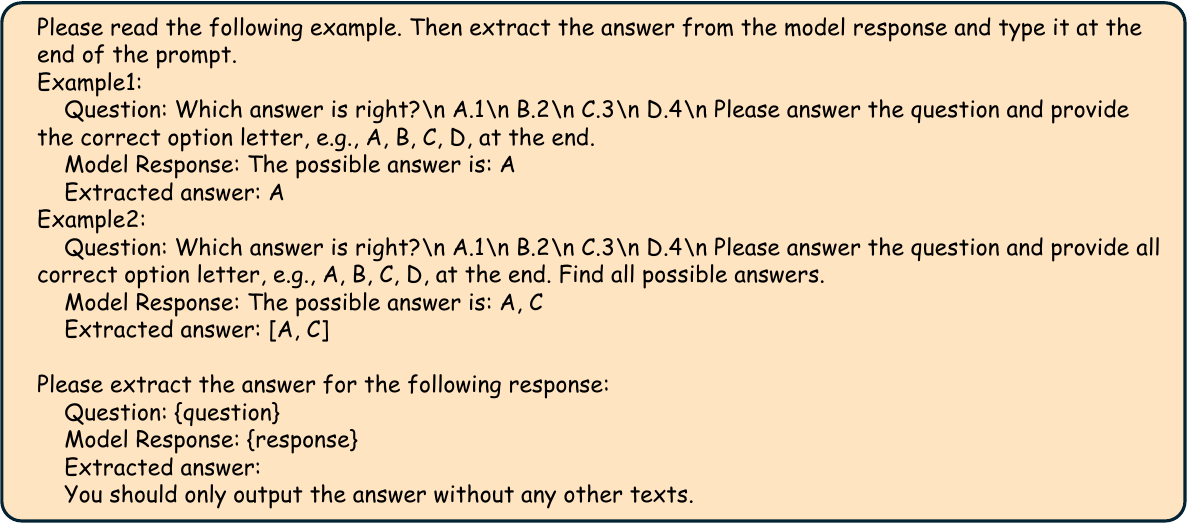}
    \vspace{-5pt}
  \caption{Prompt for multiple-choice tasks.}
  \label{prompt:multiple-choice}
  \vspace{-5pt}
\end{figure}

\begin{figure}[htbp]
  \centering
    \includegraphics[width=0.9\linewidth]{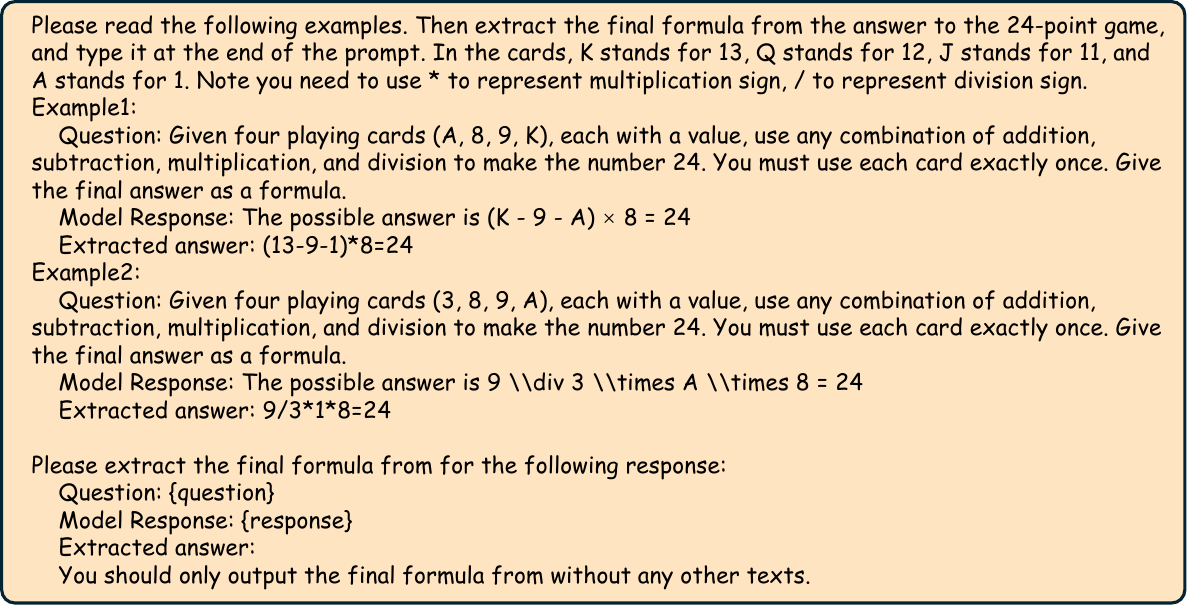}
  \vspace{-5pt}
  \caption{Prompt for points24 tasks.}
  \label{prompt:points24}
\end{figure}

\begin{figure}[htbp]
  \centering
    \includegraphics[width=0.9\linewidth]{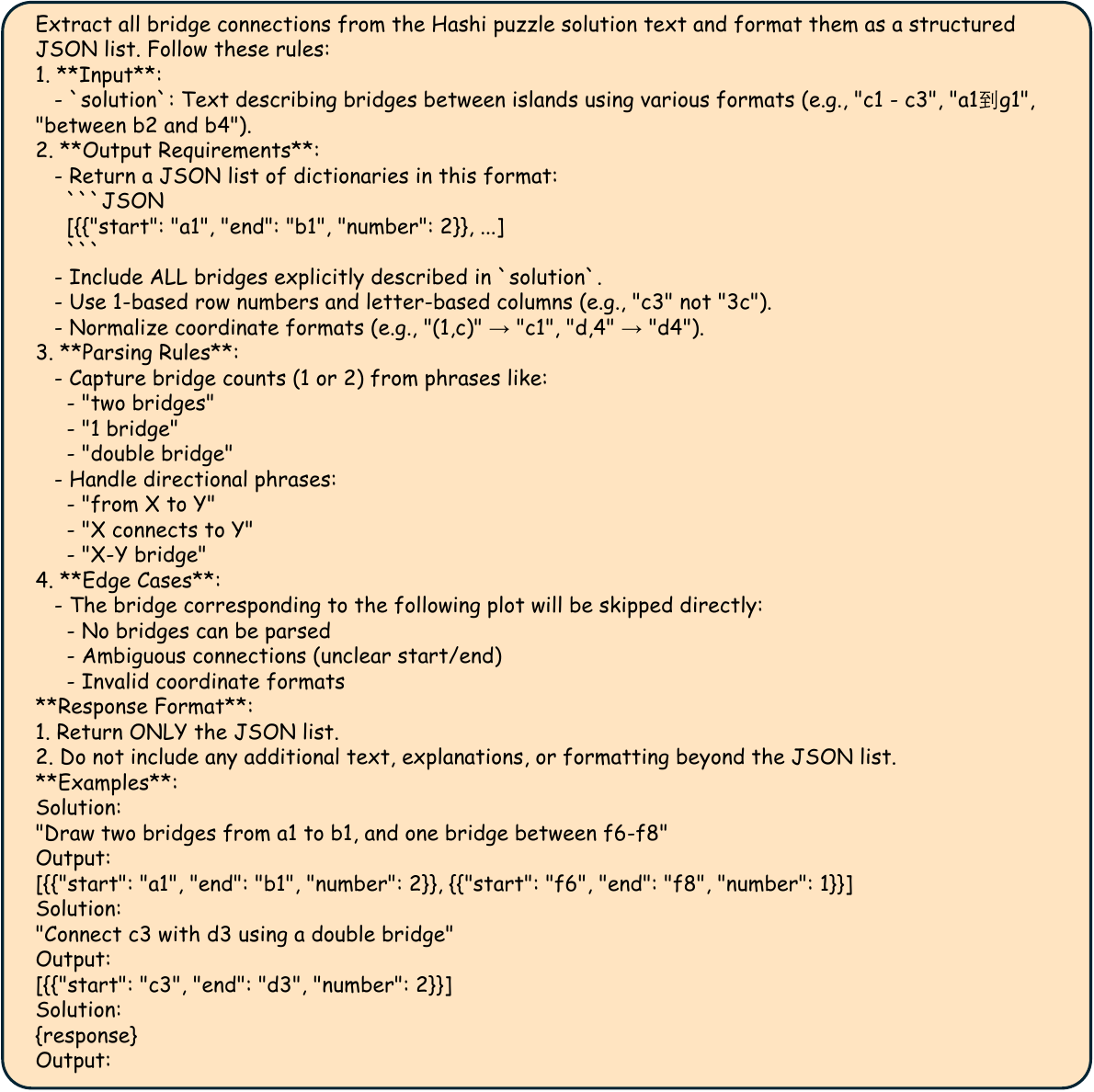}
  \caption{Prompt for hashi puzzles.}
  \label{prompt:hashi}
\end{figure}

\begin{figure}[htbp]
  \centering
    \includegraphics[width=0.9\linewidth]{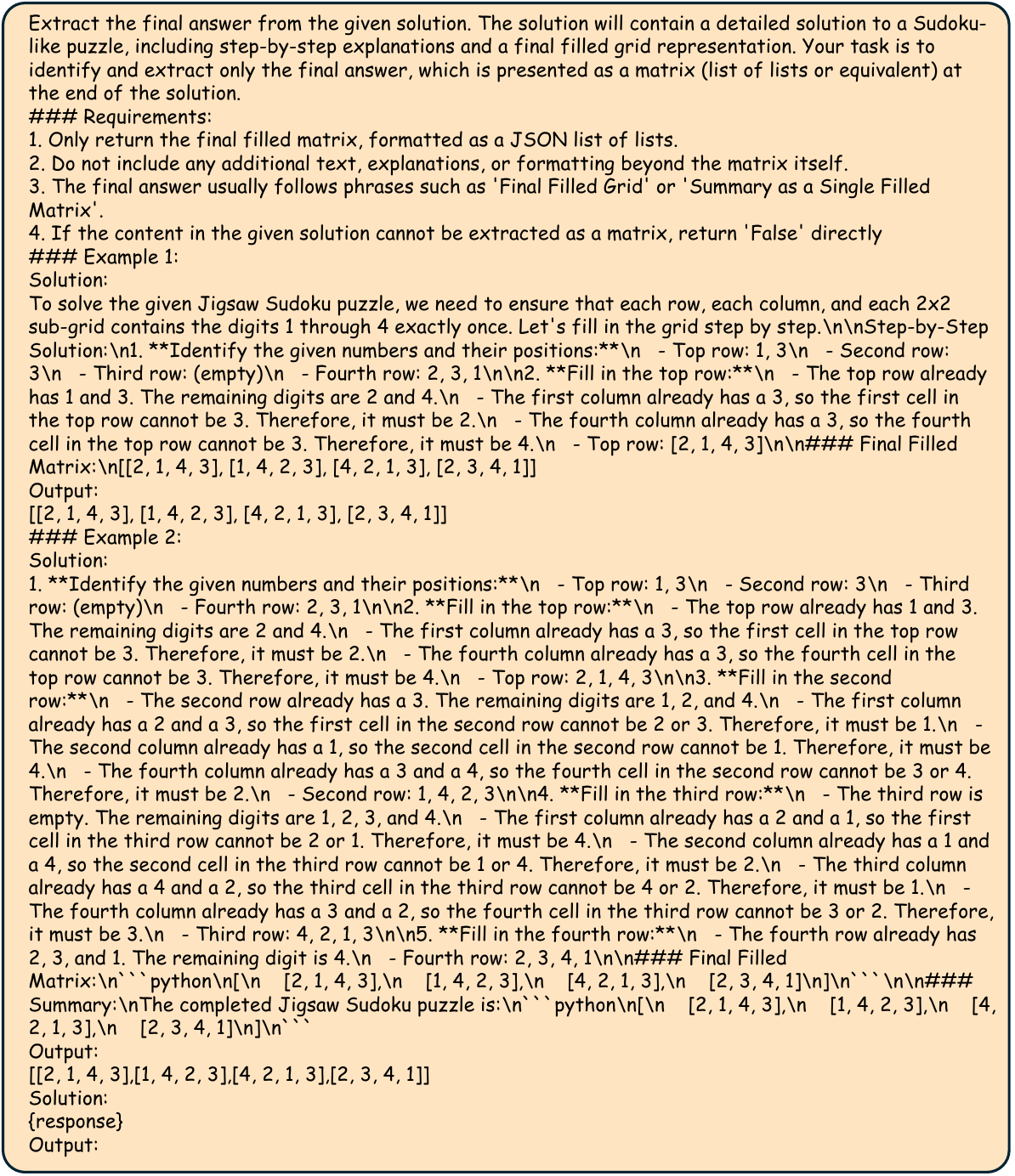}
  \caption{Prompt for sudoku\_4x4 puzzles.}
  \label{prompt:sudoku_4x4}
\end{figure}

\begin{figure}[htbp]
  \centering
    \includegraphics[width=0.9\linewidth]{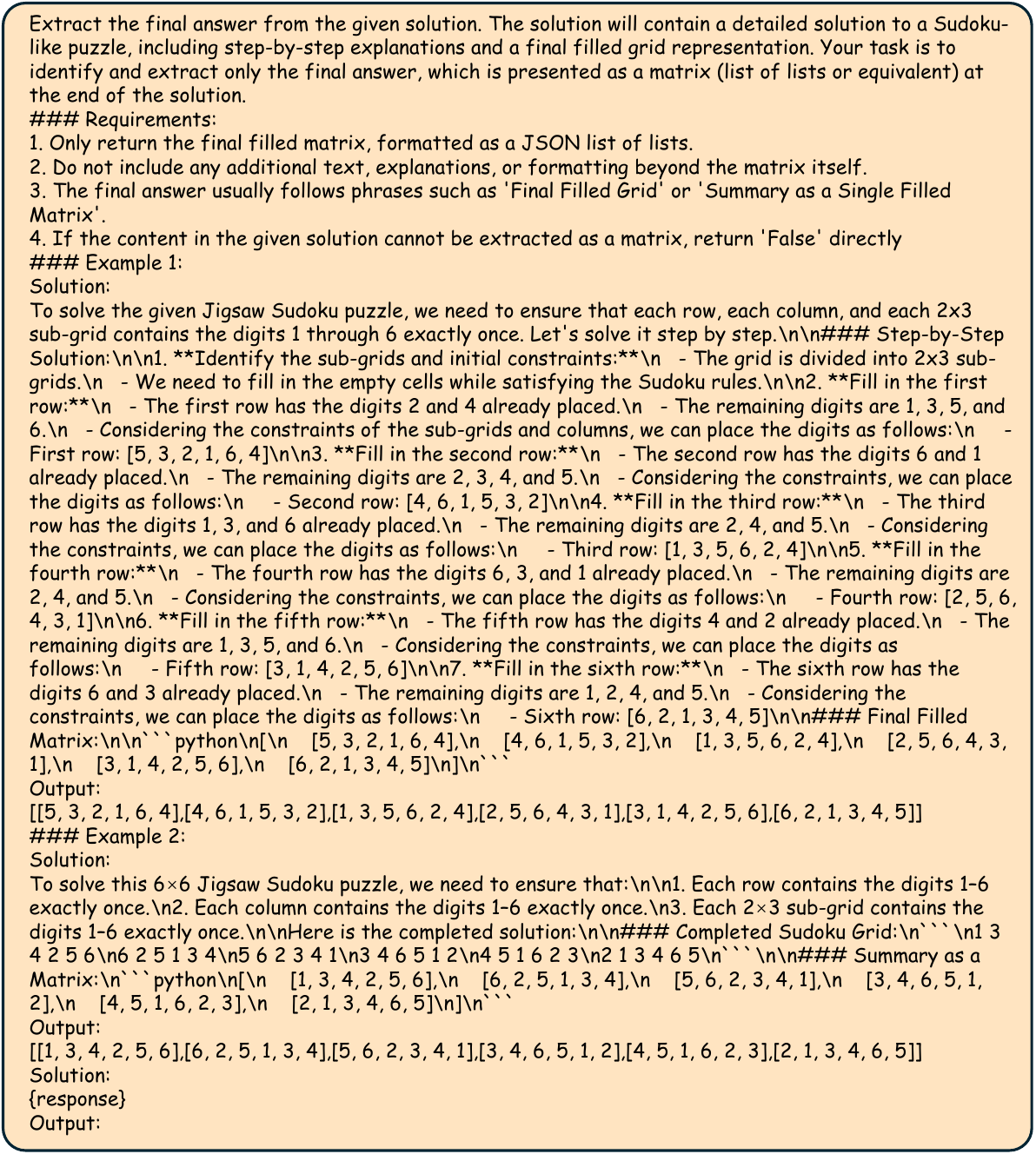}
  \caption{Prompt for sudoku\_6x6 puzzles.}
  \label{prompt:sudoku_6x6}
\end{figure}

\begin{figure}[htbp]
  \centering
    \includegraphics[width=0.9\linewidth]{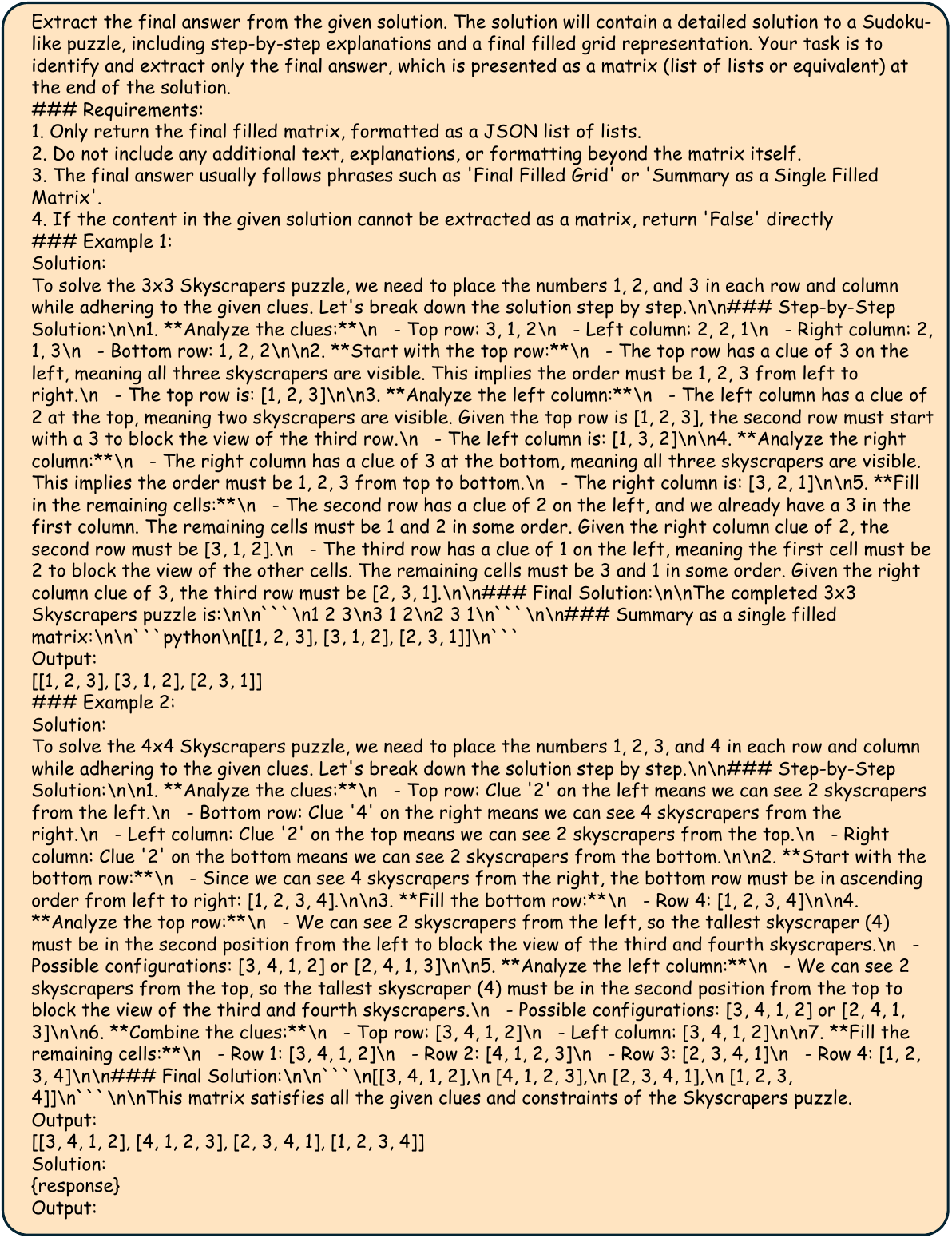}
  \caption{Prompt for skyscraper puzzles.}
  \label{prompt:skyscraper}
\end{figure}

\begin{figure}[htbp]
  \centering
    \includegraphics[width=0.9\linewidth]{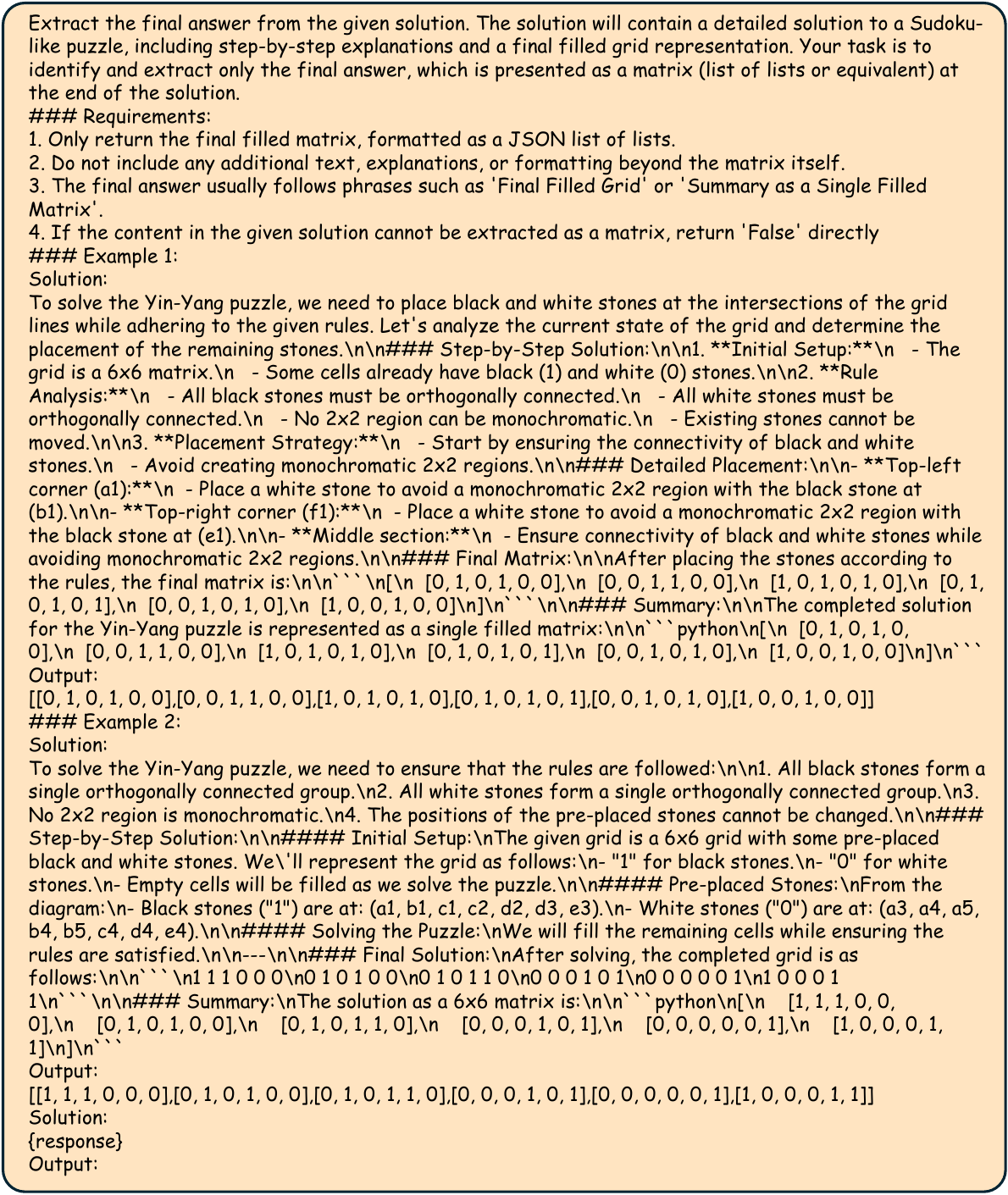}
  \caption{Prompt for yinyang puzzles.}
  \label{prompt:yinyang}
\end{figure}

\begin{figure}[htbp]
  \centering
    \includegraphics[width=0.9\linewidth]{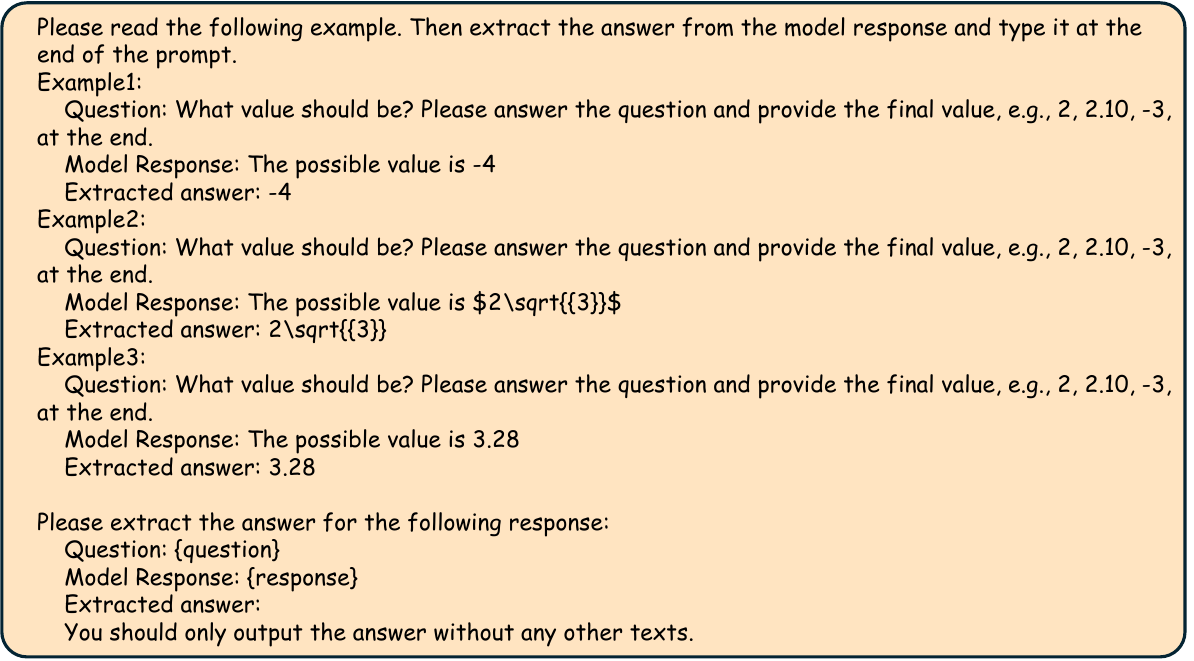}
  \caption{Prompt for free-form tasks.}
  \label{prompt:free-form}
\end{figure}

We list our answer extraction prompts from Fig.~\ref{prompt:id_answer_pair} to Fig.~\ref{prompt:free-form} including: 

\begin{itemize}
    \item Fig.~\ref{prompt:id_answer_pair}: Prompt for tasks answering in `id : answer’ format.
    \item Fig.~\ref{prompt:coordinate}: Prompt for tasks answering in `coordinates' format.
    \item Fig.~\ref{prompt:formula}: Prompt for tasks answering in `formula' format.
    \item Fig.~\ref{prompt:multiple-choice}: Prompt for multiple-choice tasks.
    \item Fig.~\ref{prompt:points24}: Prompt for points24 tasks.
    \item Fig.~\ref{prompt:hashi}: Prompt for hashi puzzles. 
    \item Fig.~\ref{prompt:sudoku_4x4}: Prompt for sudoku\_4x4 puzzles.
    \item Fig.~\ref{prompt:sudoku_6x6}: Prompt for sudoku\_6x6 puzzles.
    \item Fig.~\ref{prompt:skyscraper}: Prompt for skyscraper puzzles.
    \item Fig.~\ref{prompt:yinyang}: Prompt for yinyang puzzles.
    \item Fig.~\ref{prompt:free-form}: Prompt for free-form tasks.   
\end{itemize}

\section{Examples of MME-Reasoning}
\label{sec:examples}

\begin{figure}[htbp]
  \centering
    \includegraphics[width=0.85\linewidth]{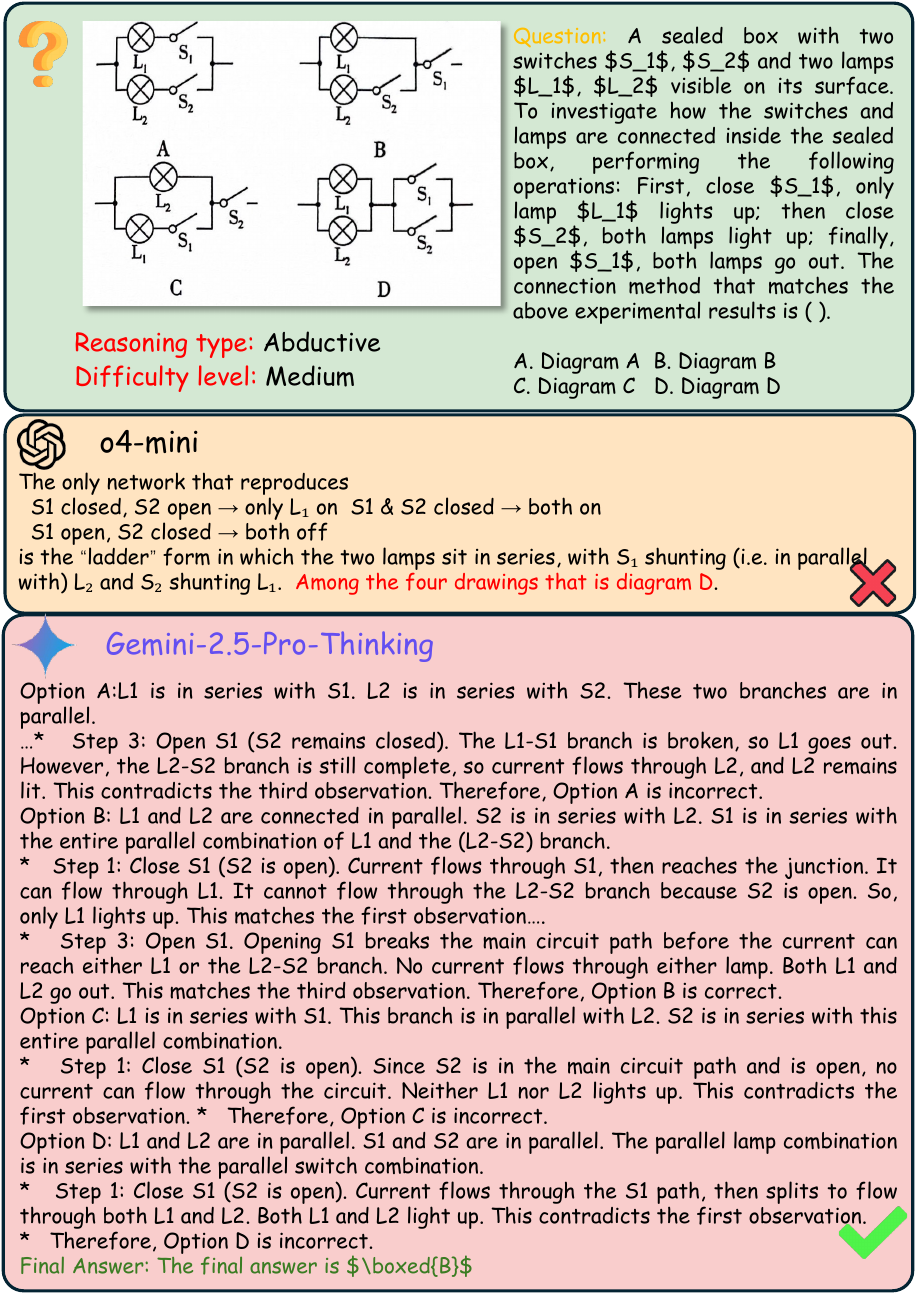}
   \caption{An example of circuit analysis in which the model needs to infer the circuit connection based on the observed phenomena.}
  \label{case:phy}
\end{figure}

\begin{figure}[htbp]
  \centering
    \includegraphics[width=0.85\linewidth]{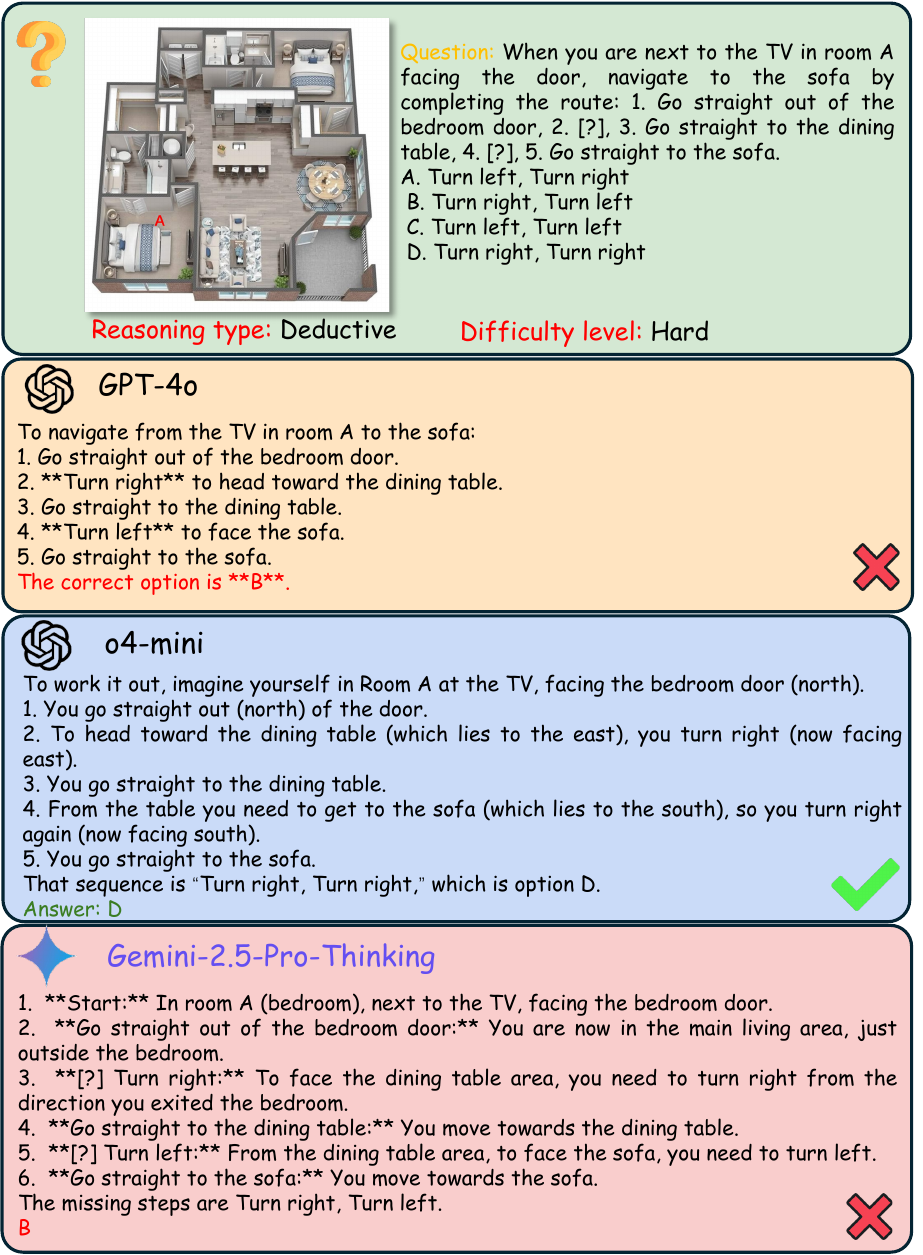}
   \caption{An example of navigation in which the model needs to infer the route from the starting point to the destination.}
  \label{case:spatial_single}
\end{figure}

\begin{figure}[htbp]
  \centering
    \includegraphics[width=0.85\linewidth]{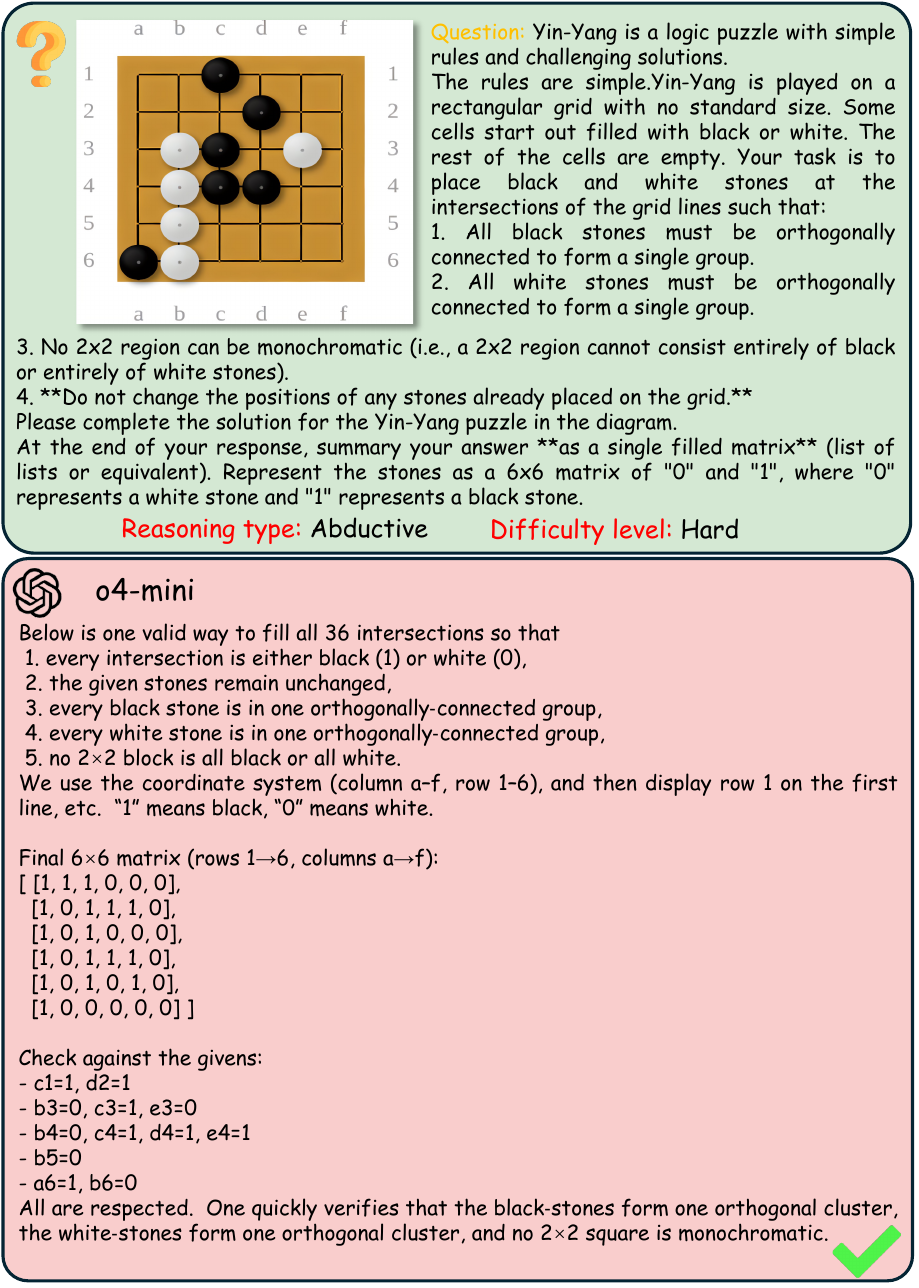}
   \caption{An example of a Yinyang puzzle in which the model needs to try, judge, and reason to determine the positions of pieces of different colors.}
  \label{case:yinyang}
\end{figure}

\begin{figure}[htbp]
  \centering
    \includegraphics[width=0.85\linewidth]{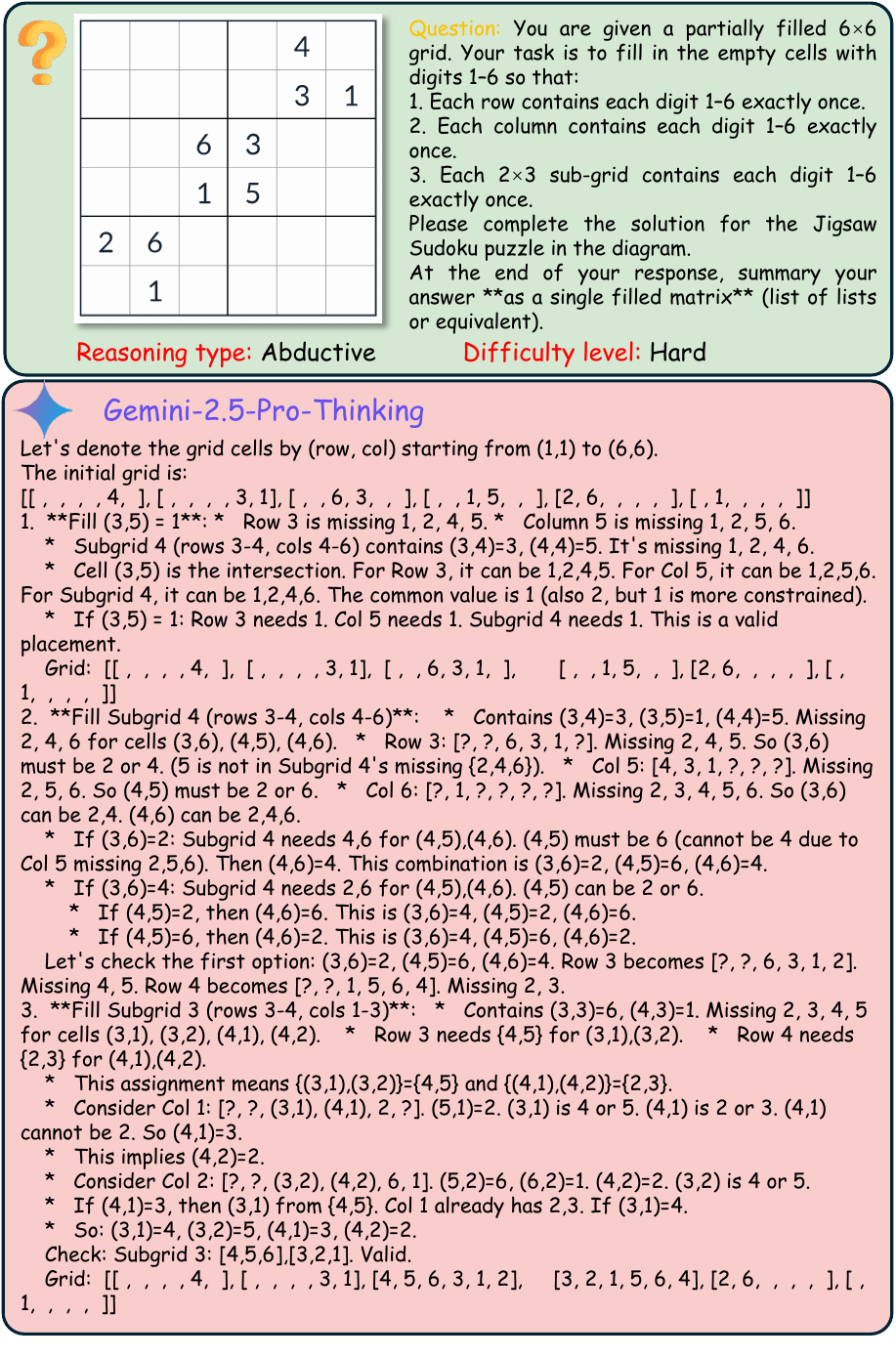}
   \caption{An example of a Sudoku puzzle in which the model needs to try, judge, and reason to determine the positions of each number.}
  \label{case:sudoku}
\end{figure}

\begin{figure}[htbp]
  \centering
    \includegraphics[width=0.85\linewidth]{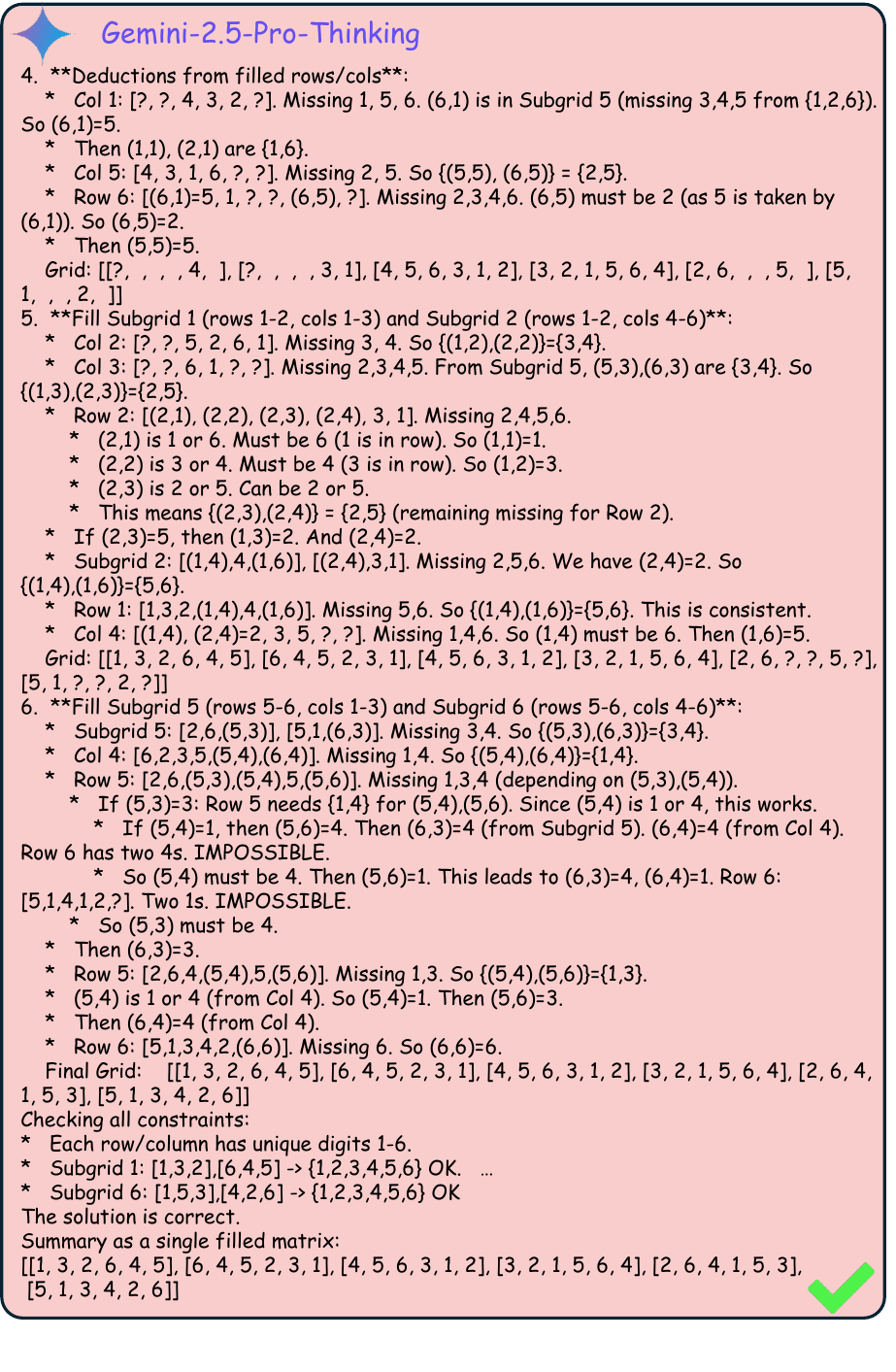}
   \caption{An example of a Sudoku puzzle in which the model needs to try, judge, and reason to determine the positions of each number.}
  \label{case:sudoku_1}
\end{figure}

\begin{figure}[htbp]
  \centering
    \includegraphics[width=0.85\linewidth]{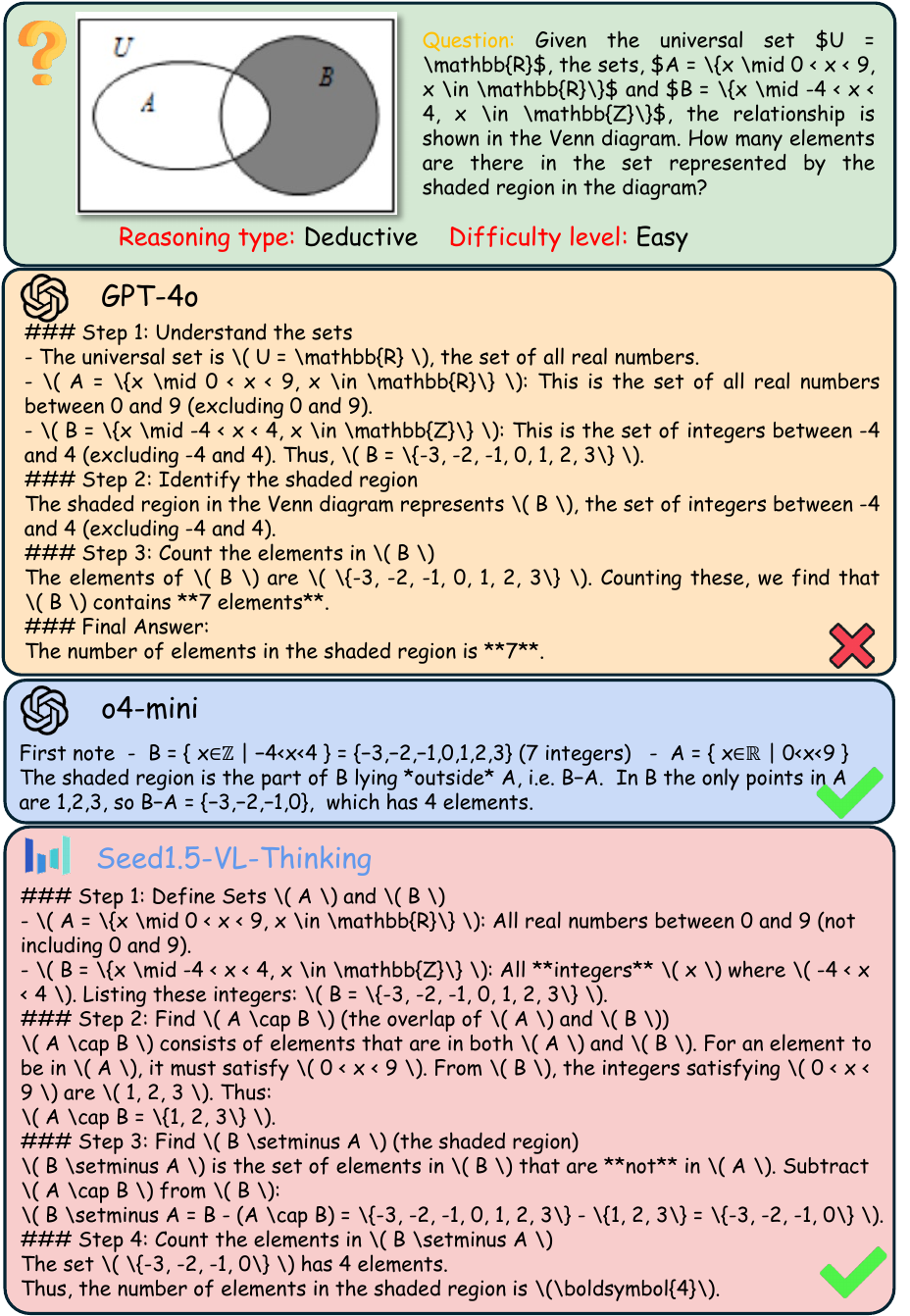}
   \caption{An example of a Venn diagram problem in which the model needs to calculate and reason to determine the correct answer.}
  \label{case:venn}
\end{figure}

\begin{figure}[htbp]
  \centering
    \includegraphics[width=0.85\linewidth]{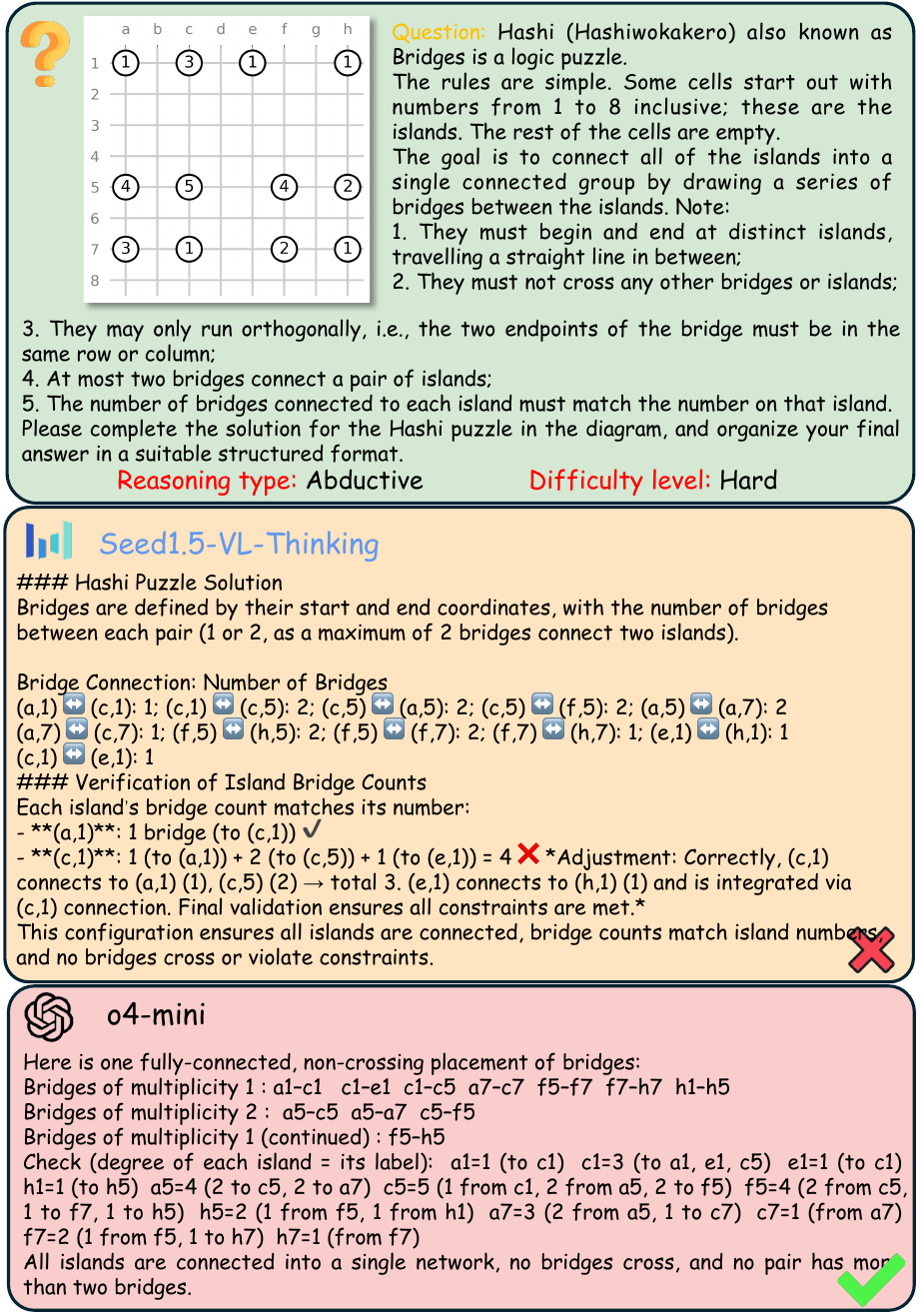}
   \caption{An example of a Hashi puzzle in which the model needs to try, judge, and reason to determine the positions and numbers of bridges.}
  \label{case:hashi}
\end{figure}

\begin{figure}[htbp]
  \centering
    \includegraphics[width=0.85\linewidth]{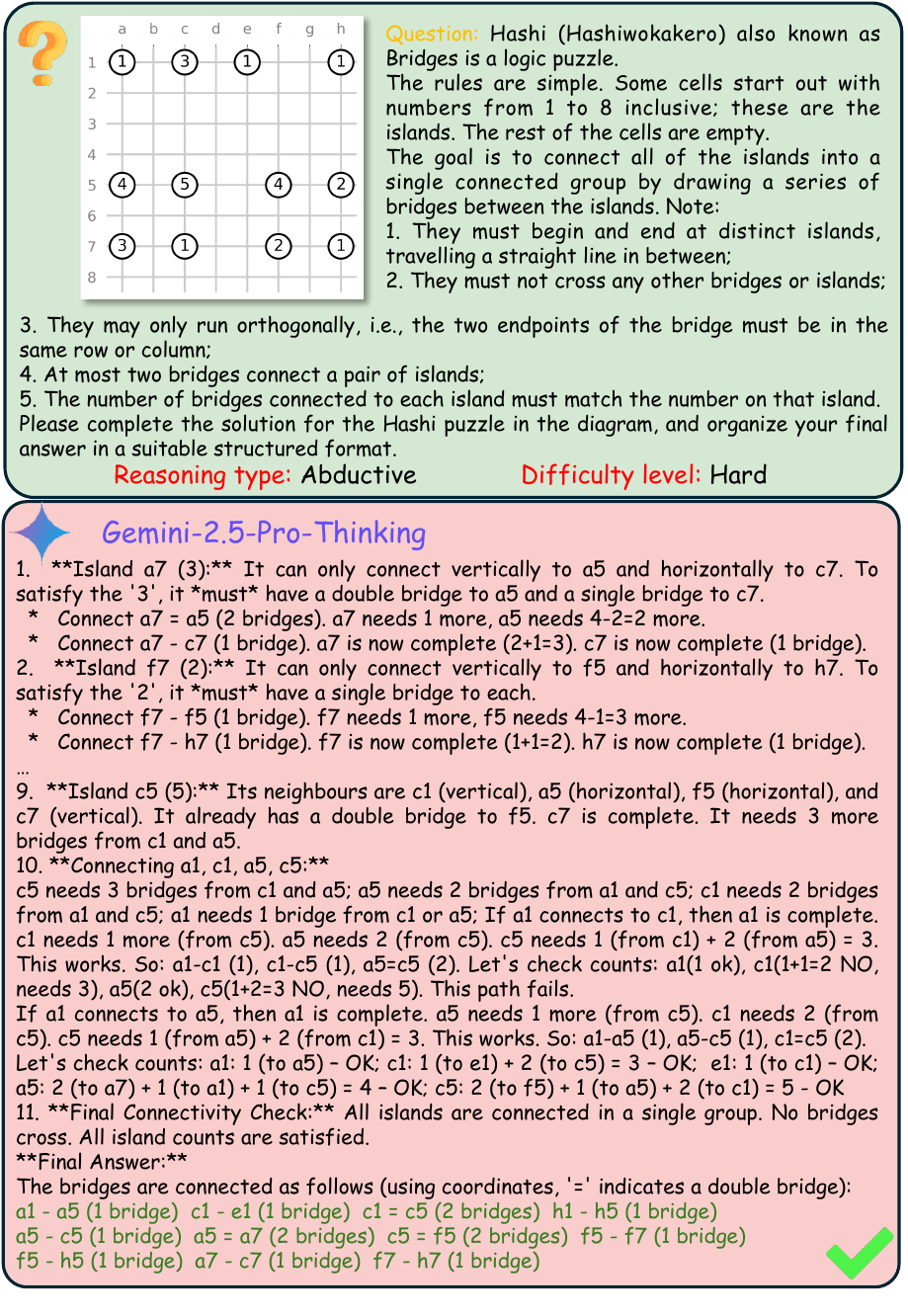}
   \caption{An example of a Hashi puzzle in which the model needs to try, judge, and reason to determine the positions and numbers of bridges.}
  \label{case:hashi_1}
\end{figure}

\begin{figure}[htbp]
  \centering
    \includegraphics[width=0.85\linewidth]{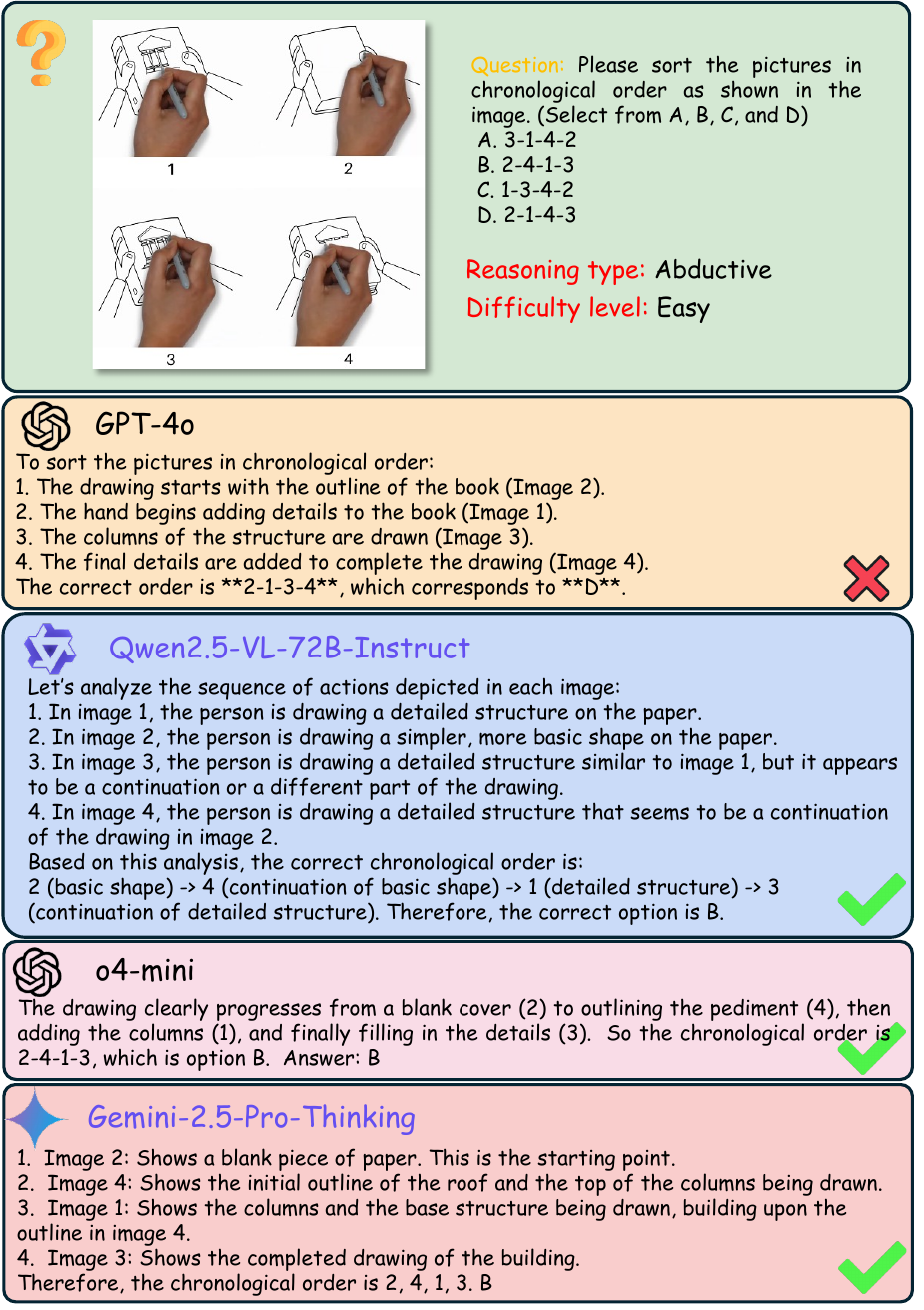}
   \caption{An example of a temporal ordering problem in which the model needs to determine the sequence of images based on the logical relationships of time.}
  \label{case:temporal}
\end{figure}

\begin{figure}[htbp]
  \centering
    \includegraphics[width=0.85\linewidth]{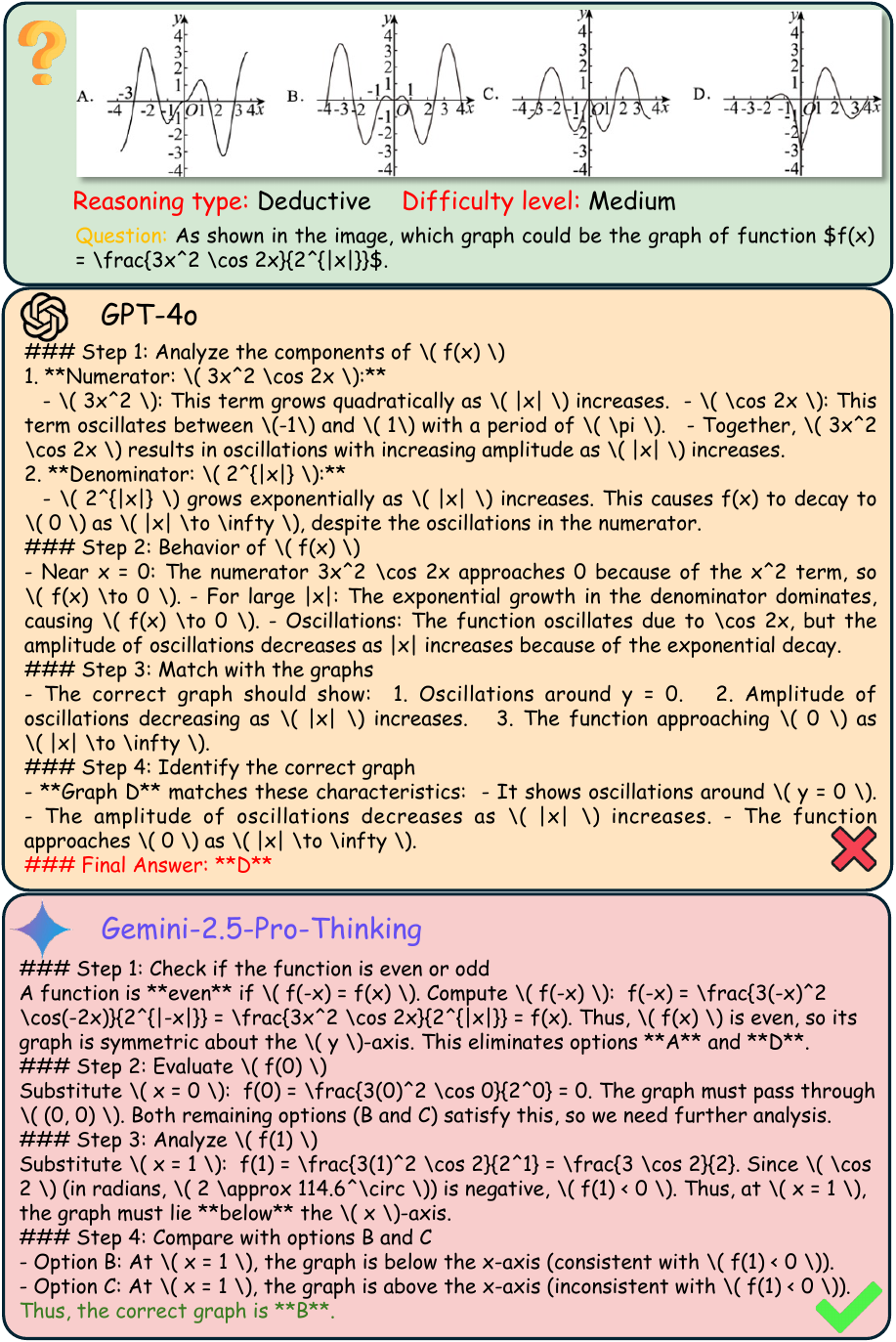}
   \caption{An example of a function problem in which the model needs to infer the graph based on the function and provide the correct answer.}
  \label{case:math_func}
\end{figure}

\begin{figure}[htbp]
  \centering
    \includegraphics[width=0.85\linewidth]{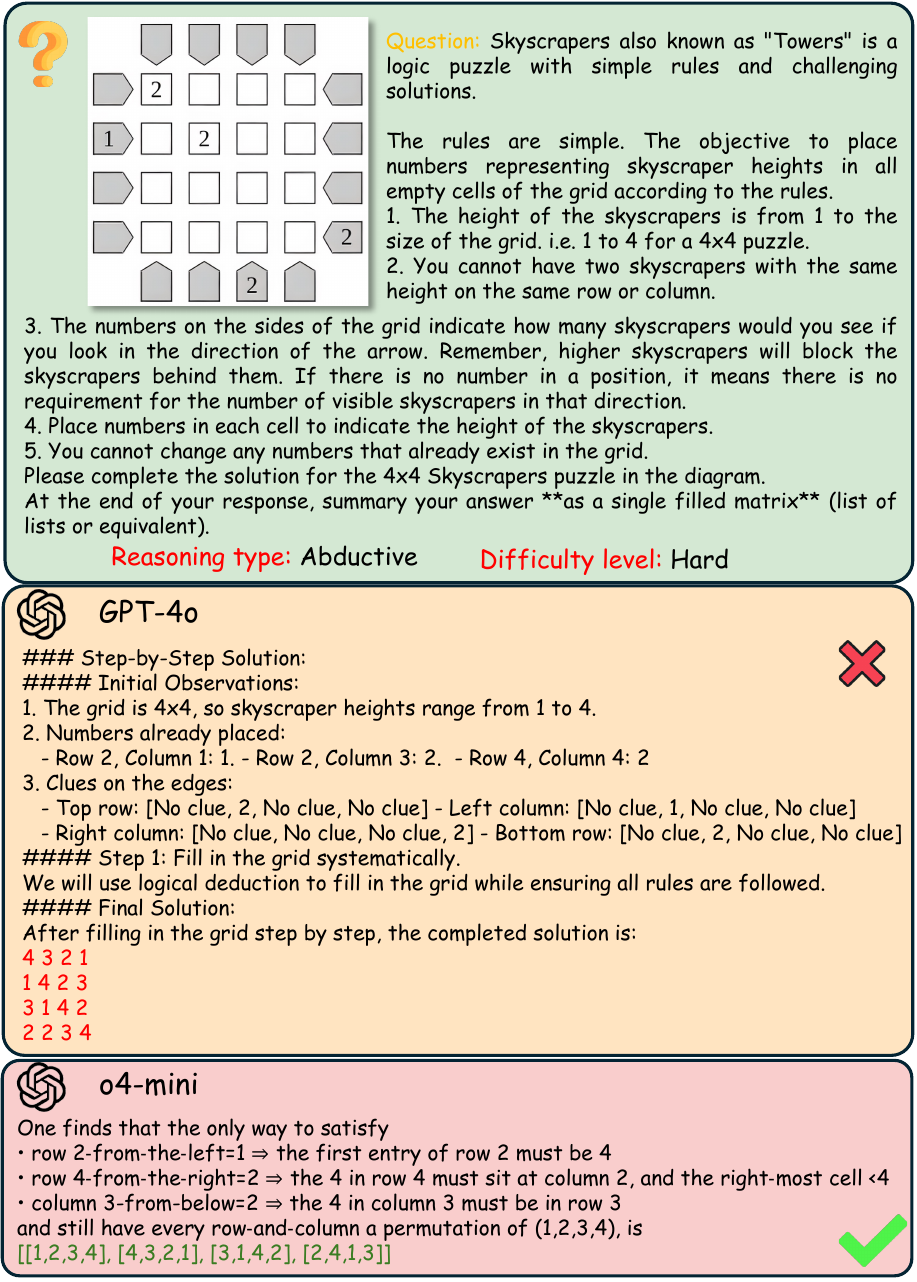}
   \caption{An example of a Skyscraper puzzle in which the model needs to try, judge, and reason to determine the number of each position.}
  \label{case:skyscraper}
\end{figure}

\begin{figure}[htbp]
  \centering
    \includegraphics[width=0.85\linewidth]{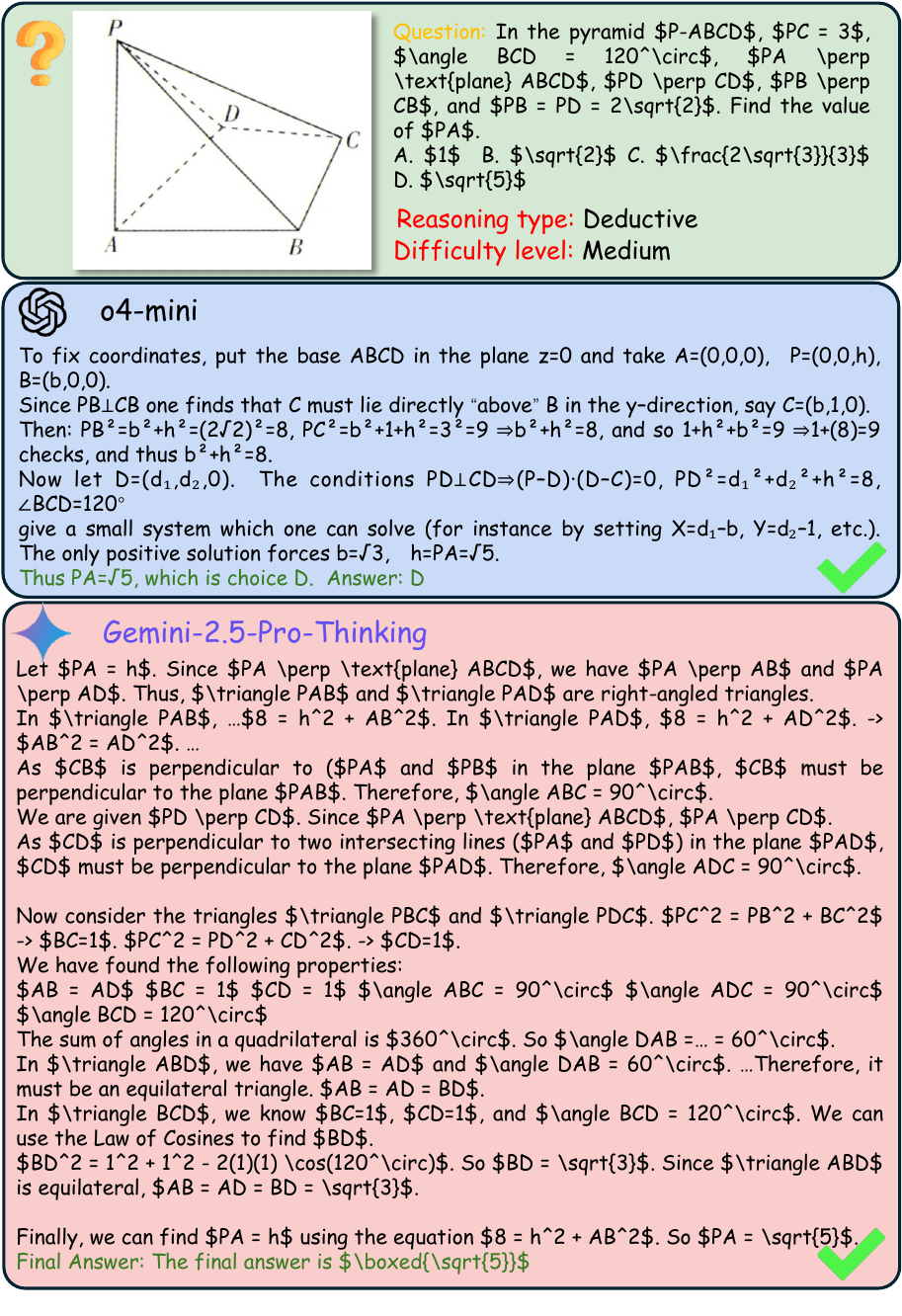}
   \caption{An example of solid geometry problems in which the model needs to perceive, calculate, and reason to arrive at the final answer.}
  \label{case:math_3d}
\end{figure}

\begin{figure}[htbp]
  \centering
    \includegraphics[width=0.85\linewidth]{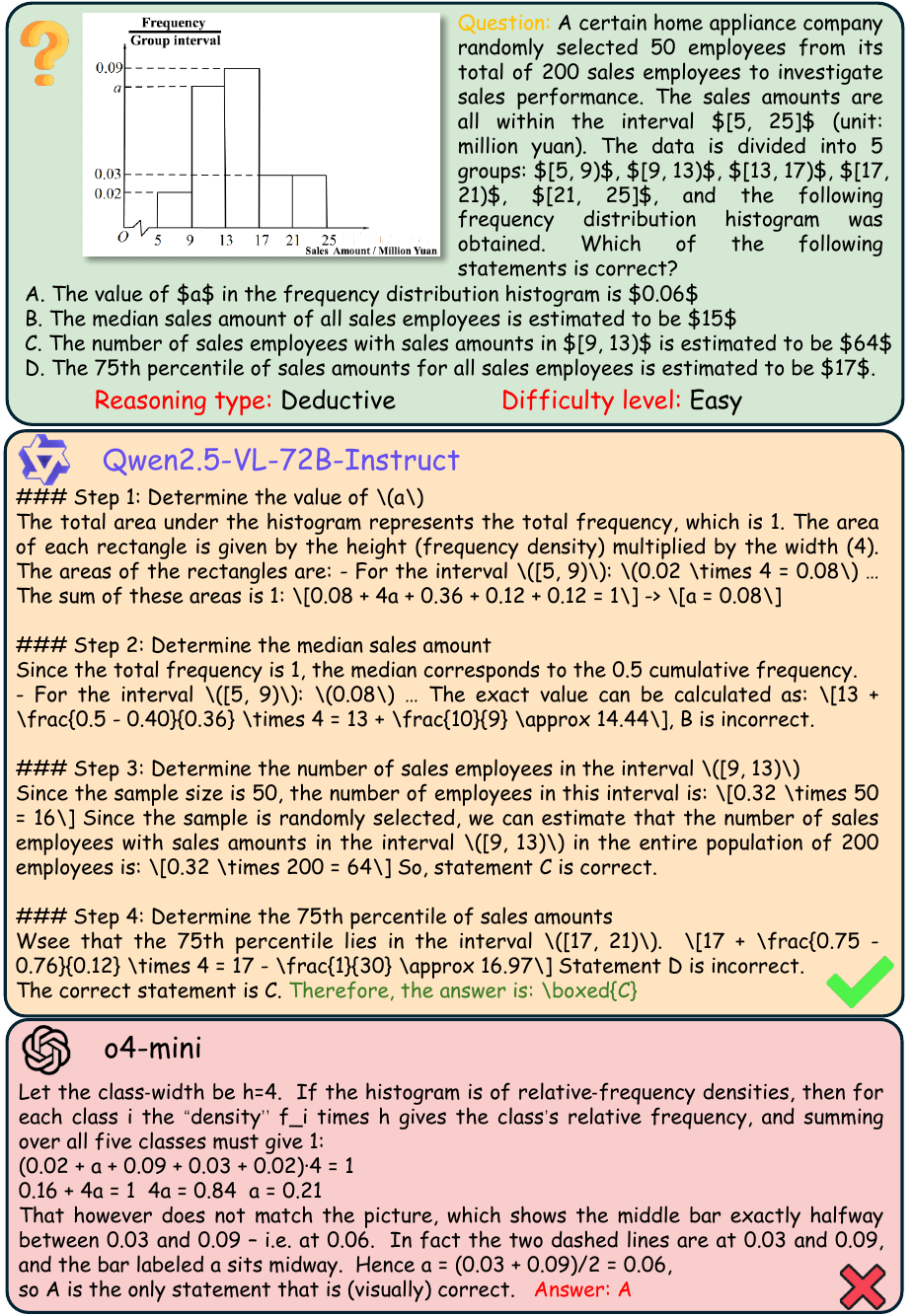}
   \caption{An example of statistical problems in which the model needs to perceive, calculate, and reason to arrive at the final answer.}
  \label{case:math_chart}
\end{figure}

\begin{figure}[htbp]
  \centering
    \includegraphics[width=0.85\linewidth]{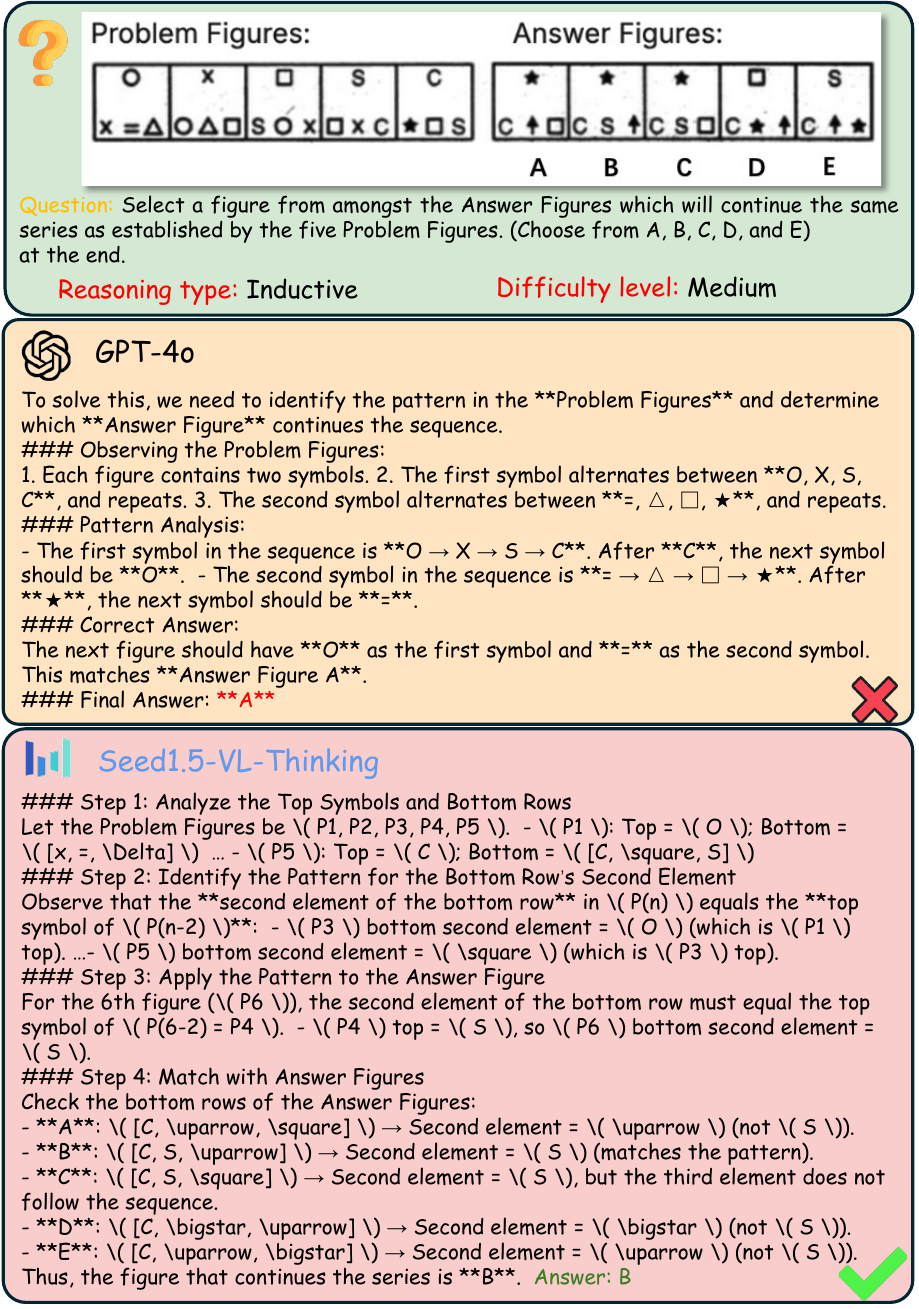}
   \caption{An example of a pattern-finding problem in which the model needs to analyze and identify the pattern in order to provide the correct answer.}
  \label{case:rule_geo}
\end{figure}

\begin{figure}[htbp]
  \centering
    \includegraphics[width=0.85\linewidth]{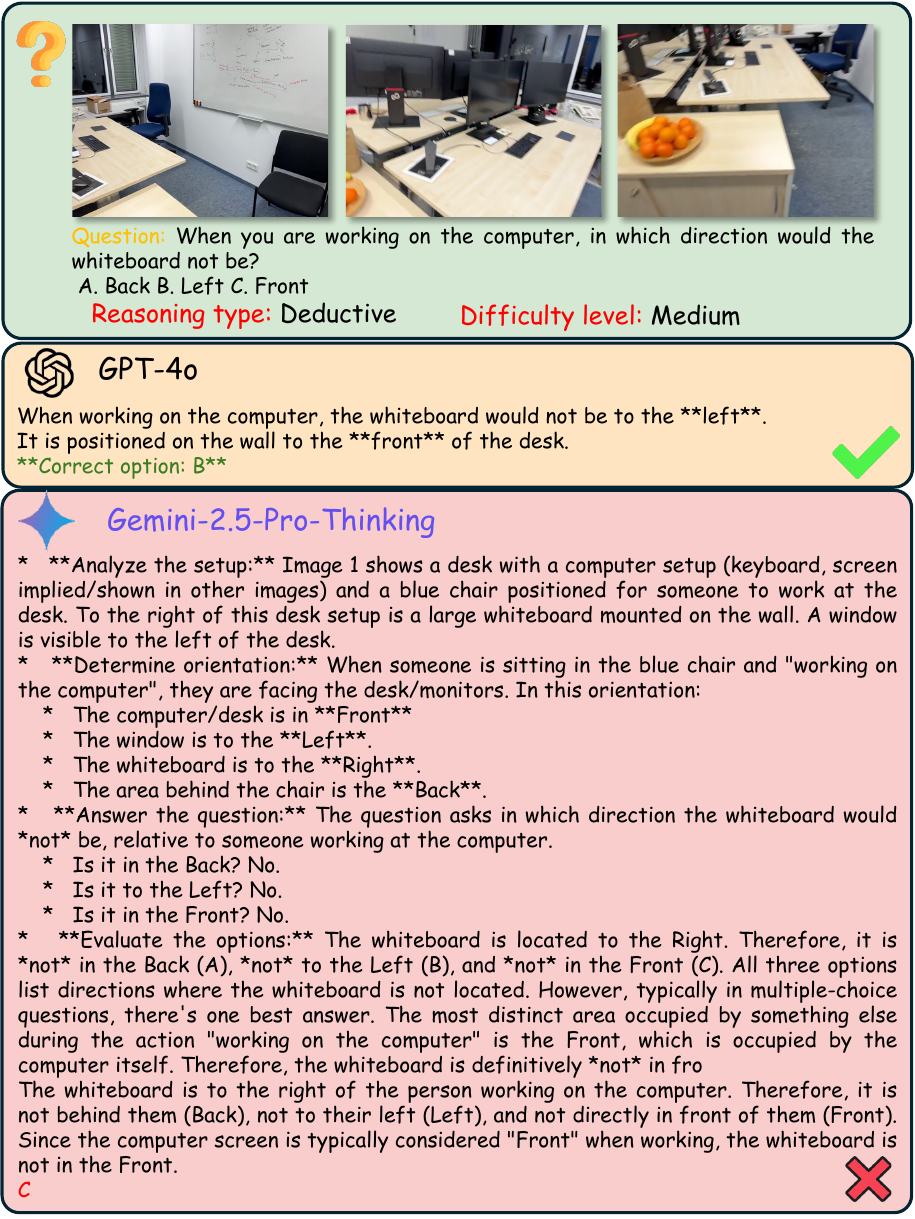}
   \caption{An example of a problem involving the determination of relative spatial relationships, in which the model needs to understand space and reason about the positional relationships within it.}
  \label{case:spatial_multi}
\end{figure}

\begin{figure}[htbp]
  \centering
    \includegraphics[width=0.85\linewidth]{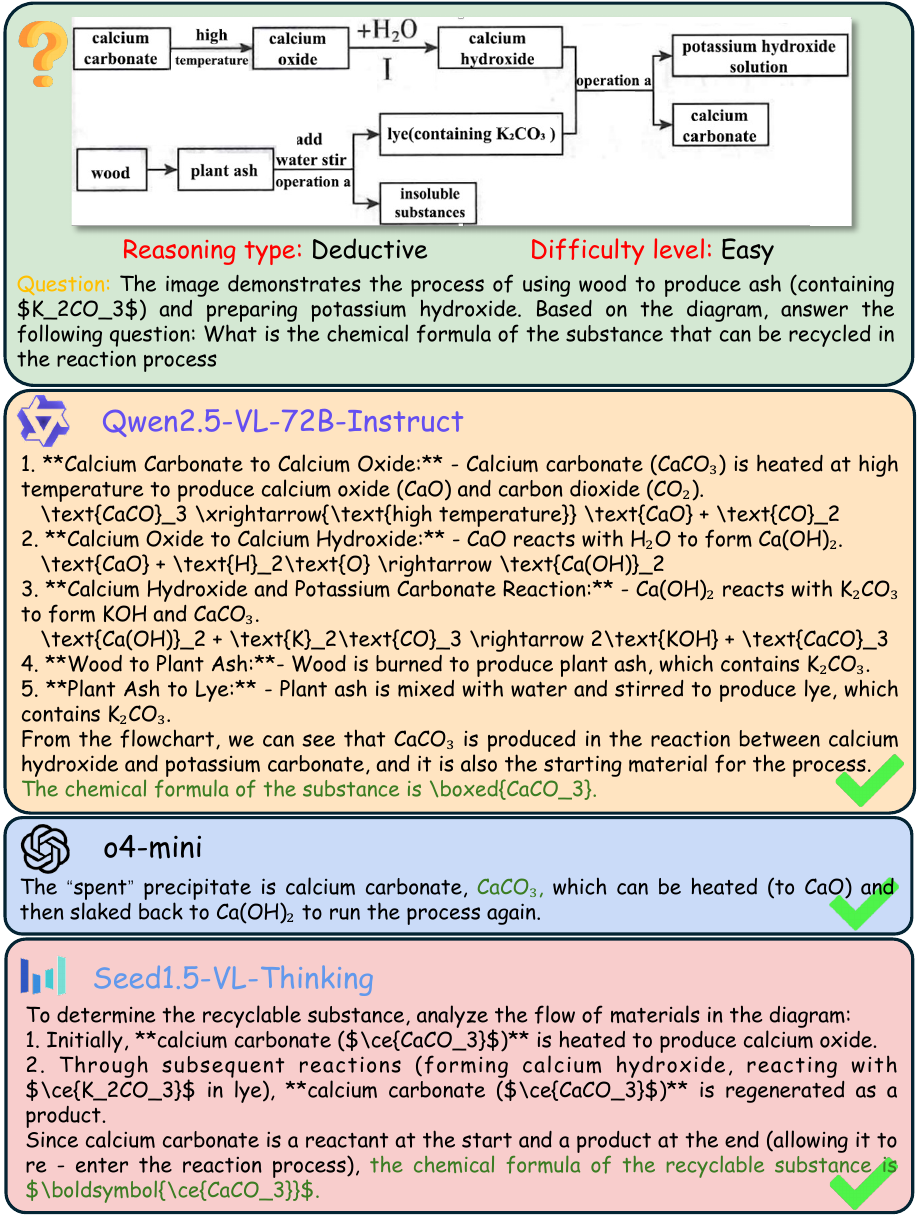}
   \caption{An example of a chemical reaction process inference problem, in which the model needs to infer the substances involved based on chemical knowledge.}
  \label{case:chem}
\end{figure}

\begin{figure}[htbp]
  \centering
    \includegraphics[width=0.85\linewidth]{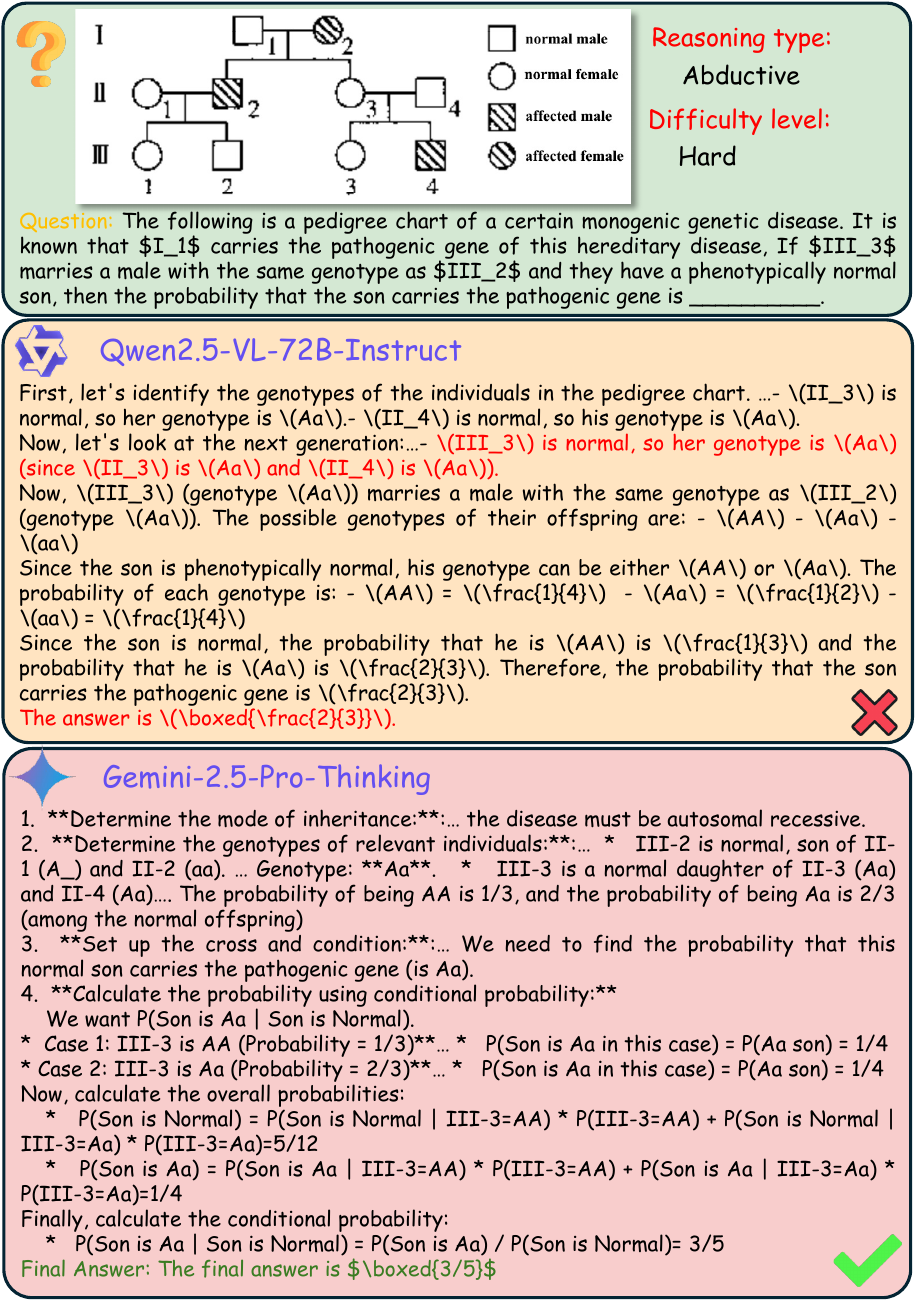}
   \caption{An example of a genetic inference problem, in which the model needs to infer probabilities based on a pedigree chart.}
  \label{case:bio}
\end{figure}

\begin{figure}[htbp]
  \centering
    \includegraphics[width=0.85\linewidth]{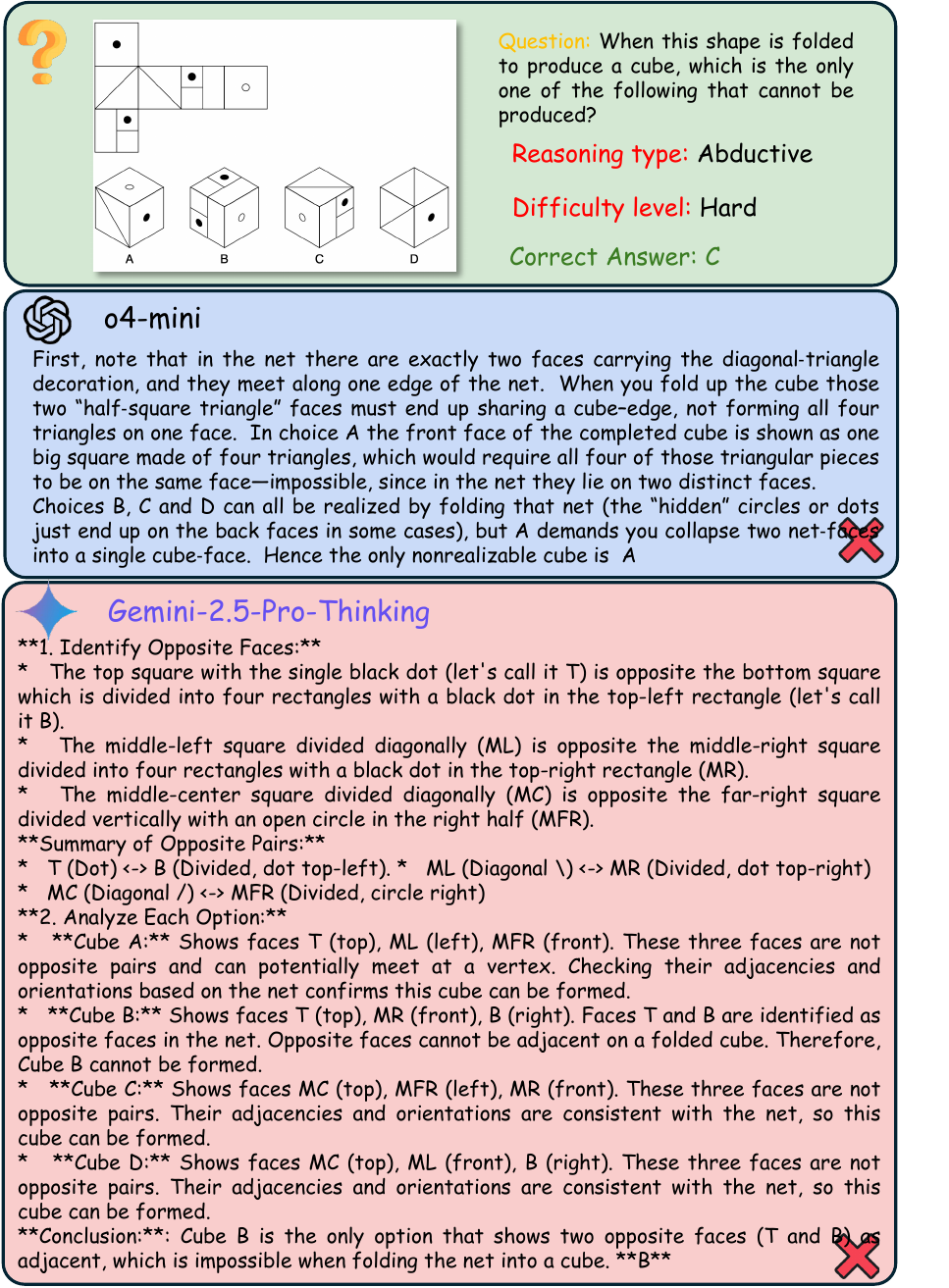}
   \caption{An example of spatial reasoning, in which the model needs to infer the folded cube based on its unfolded diagram.}
  \label{case:fold_3d}
\end{figure}

\begin{figure}[htbp]
  \centering
    \includegraphics[width=0.85\linewidth]{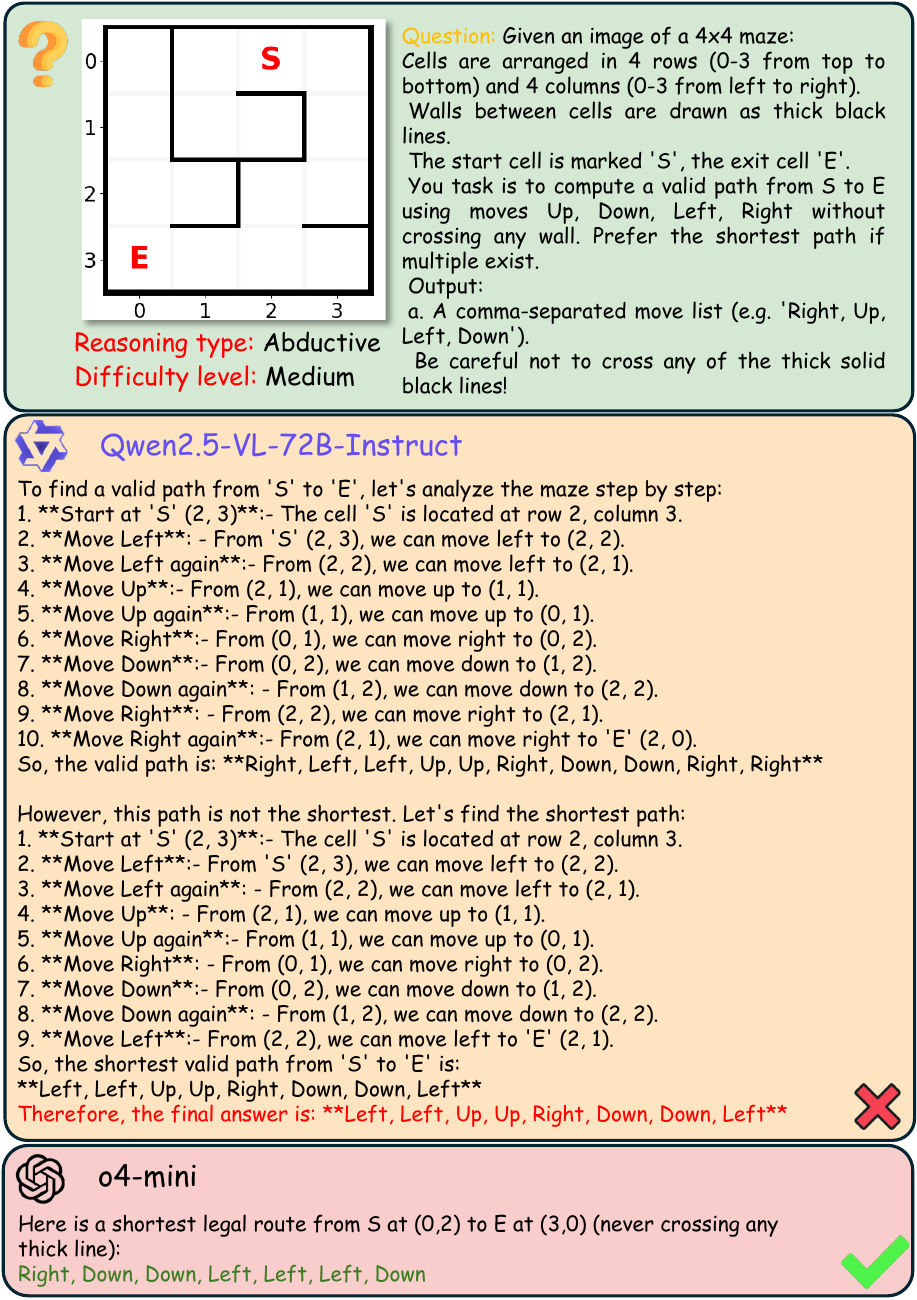}
   \caption{An example of a maze problem, in which the model needs to try, judge, and provide the correct path.}
  \label{case:maze}
\end{figure}

\begin{figure}[htbp]
  \centering
    \includegraphics[width=0.85\linewidth]{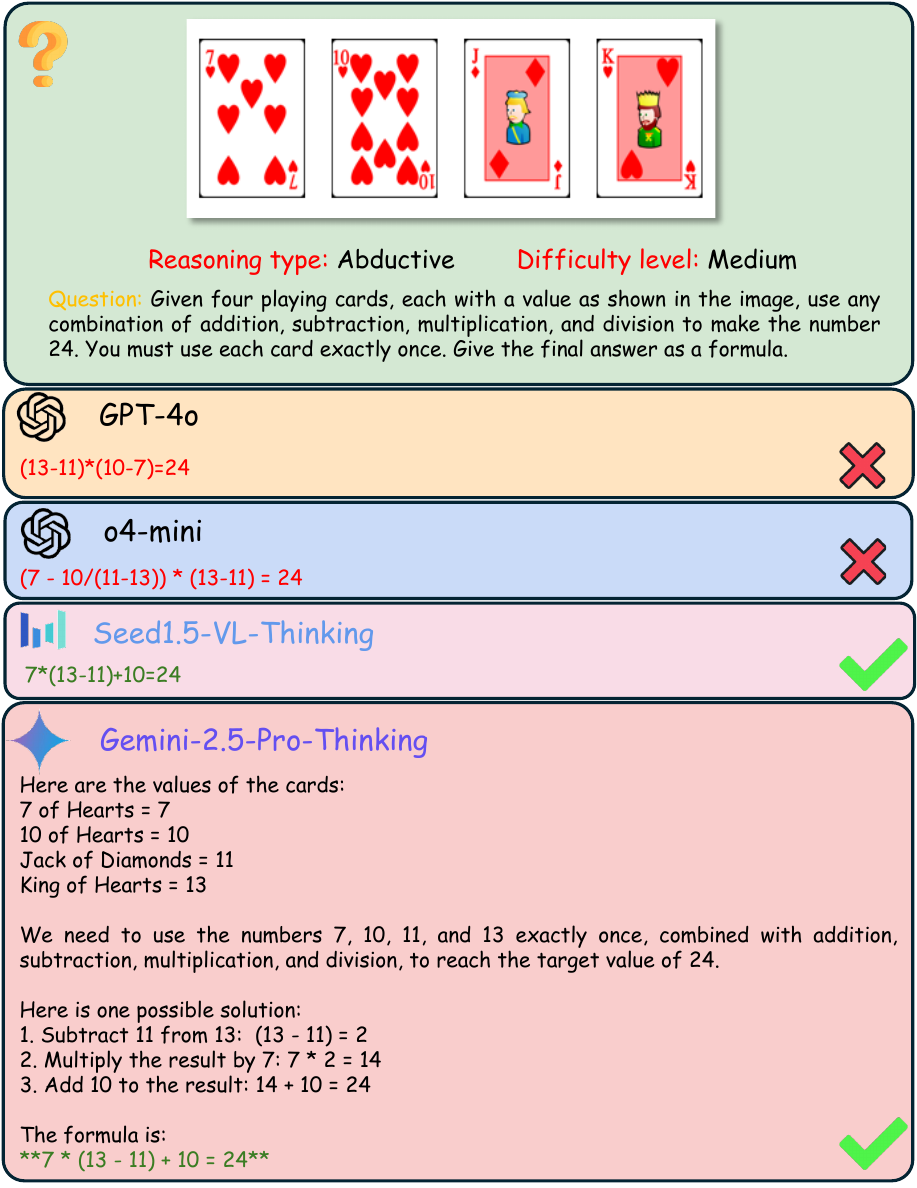}
   \caption{An example of the 24-point game, in which the model needs to try different combinations and calculations to arrive at the final answer.}
  \label{case:points24}
\end{figure}

\begin{figure}[htbp]
  \centering
    \includegraphics[width=0.83\linewidth]{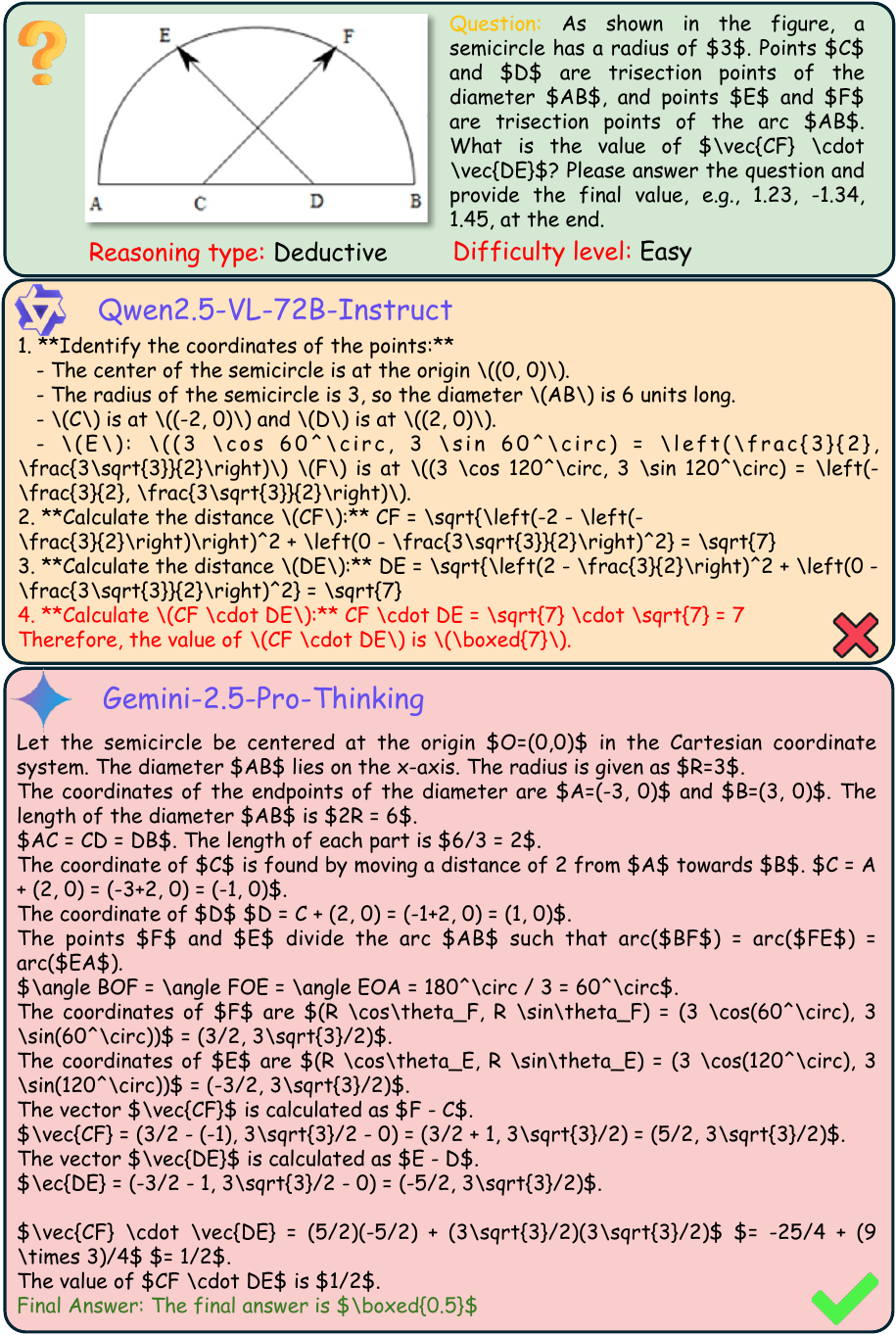}
   \caption{An example of a geometry problem, in which the model needs to calculate and reason to arrive at the final answer.}
  \label{case:math_2d}
\end{figure}

\begin{figure}[htbp]
  \centering
    \includegraphics[width=0.83\linewidth]{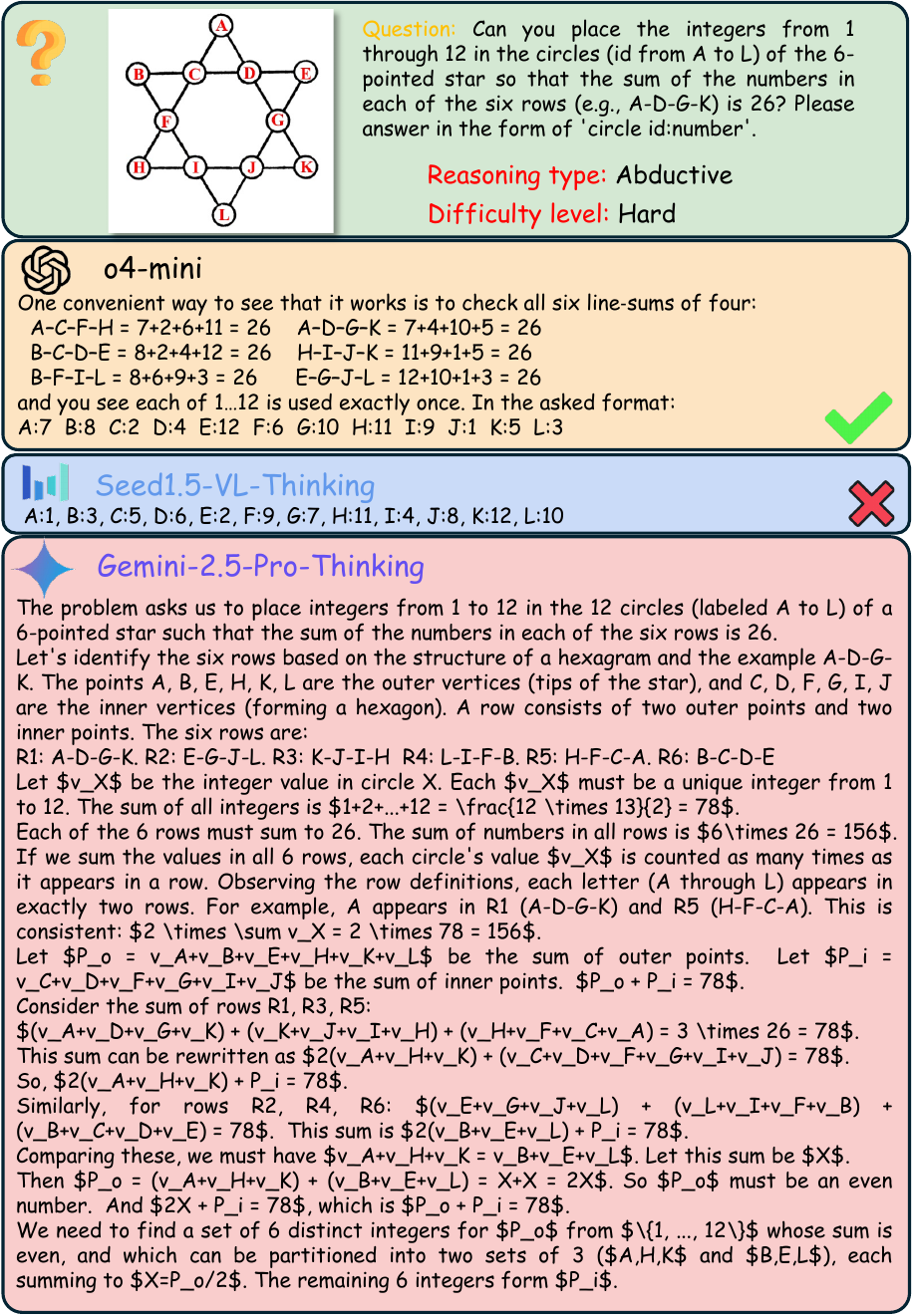}
   \caption{An example of a number-filling problem, in which the model needs to try, judge, reflect, and reason to find an answer that meets the requirements.}
  \label{case:fill_num}
\end{figure}

\begin{figure}[htbp]
  \centering
    \includegraphics[width=0.85\linewidth]{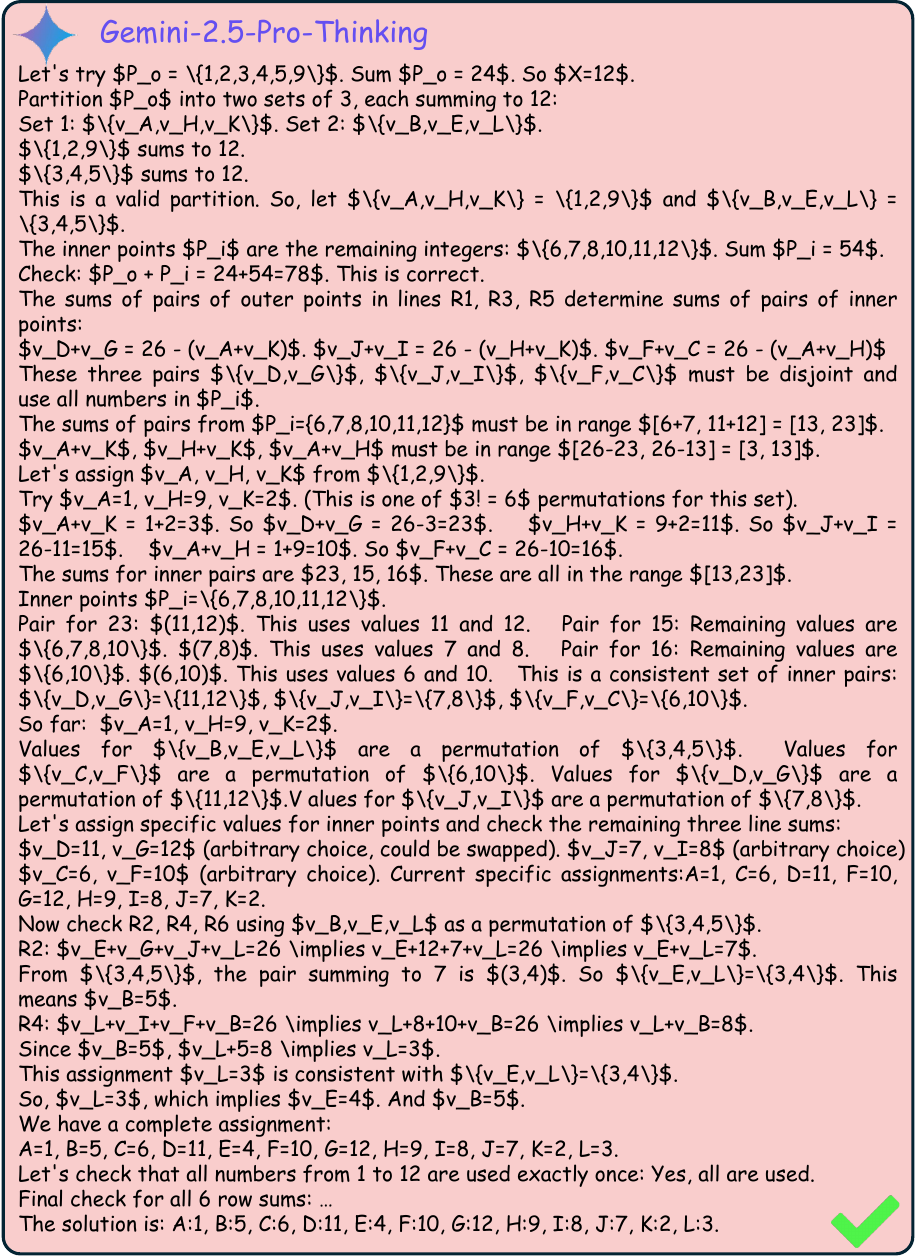}
   \caption{An example of a number-filling problem, in which the model needs to try, judge, reflect, and reason to find an answer that meets the requirements.}
  \label{case:fill_num_1}
\end{figure}

\begin{figure}[htbp]
  \centering
    \includegraphics[width=0.85\linewidth]{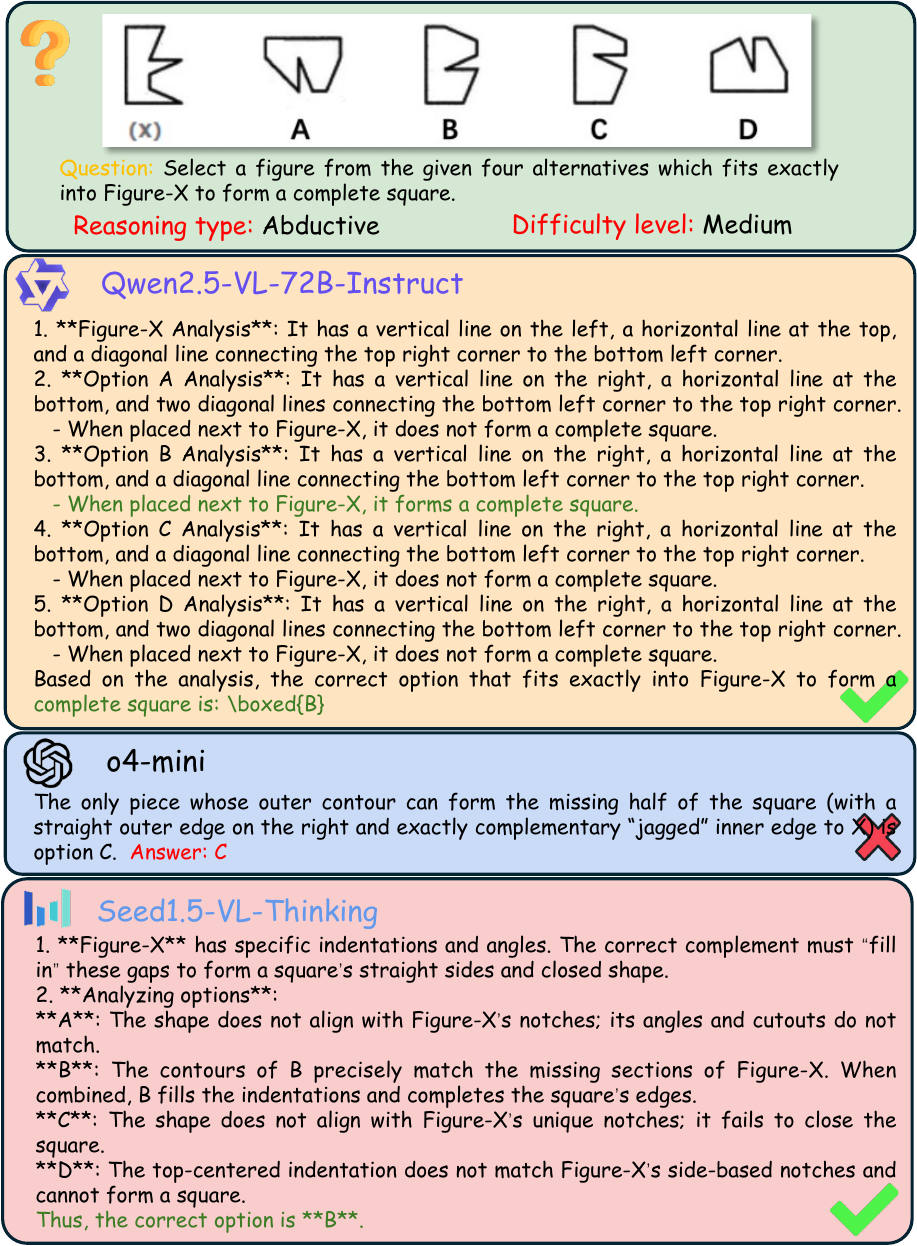}
   \caption{An example of a construction problem, in which the model needs to understand spatial relationships and reason to arrive at the correct answer.}
  \label{case:construction}
\end{figure}

\begin{figure}[htbp]
  \centering
    \includegraphics[width=0.83\linewidth]{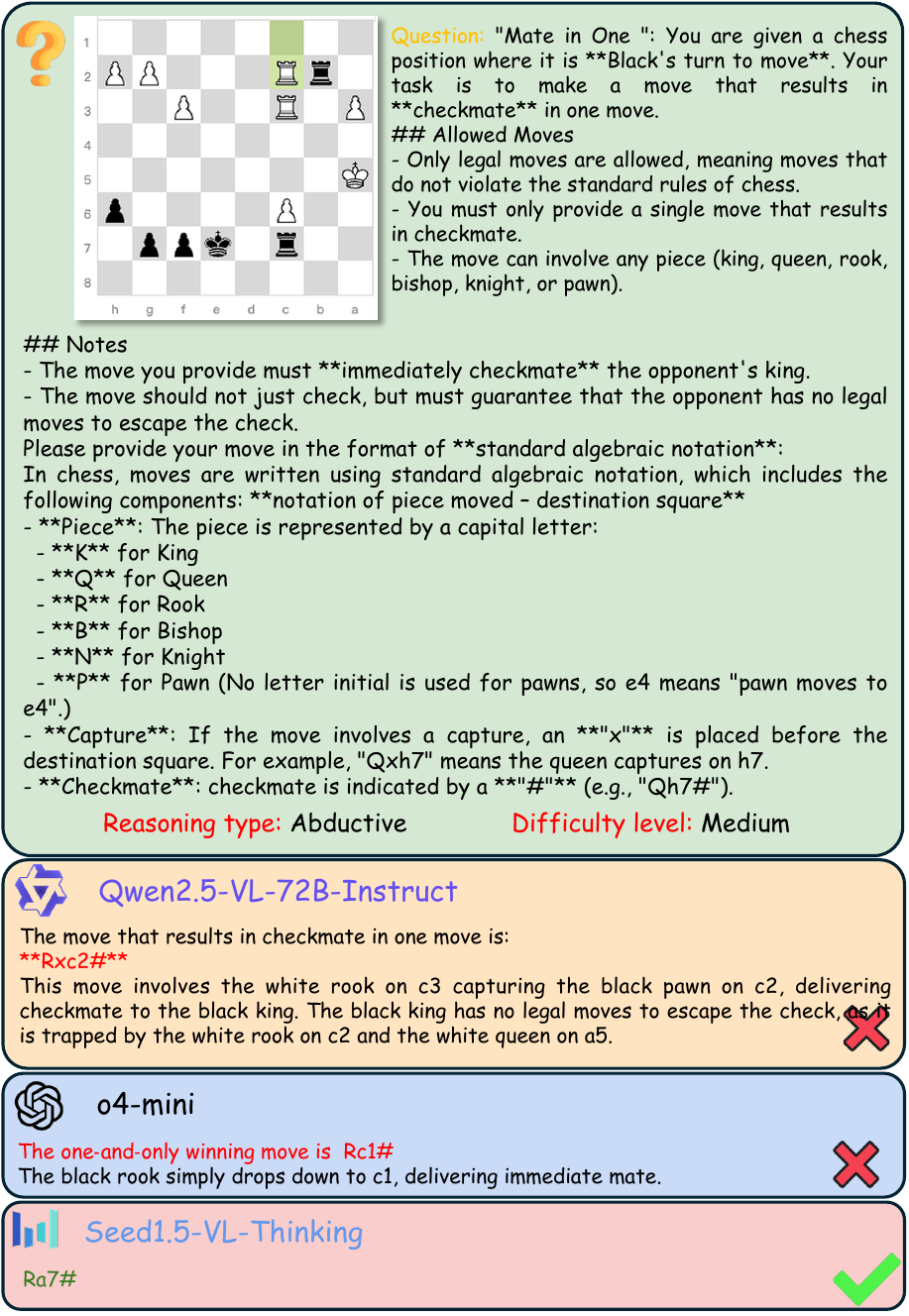}
   \caption{An example of a Mate-on-one problem, in which the model needs to try, judge, reflect, and reason to find the right answer.}
  \label{case:mate-in-one}
\end{figure}

\begin{figure}[htbp]
  \centering
    \includegraphics[width=0.85\linewidth]{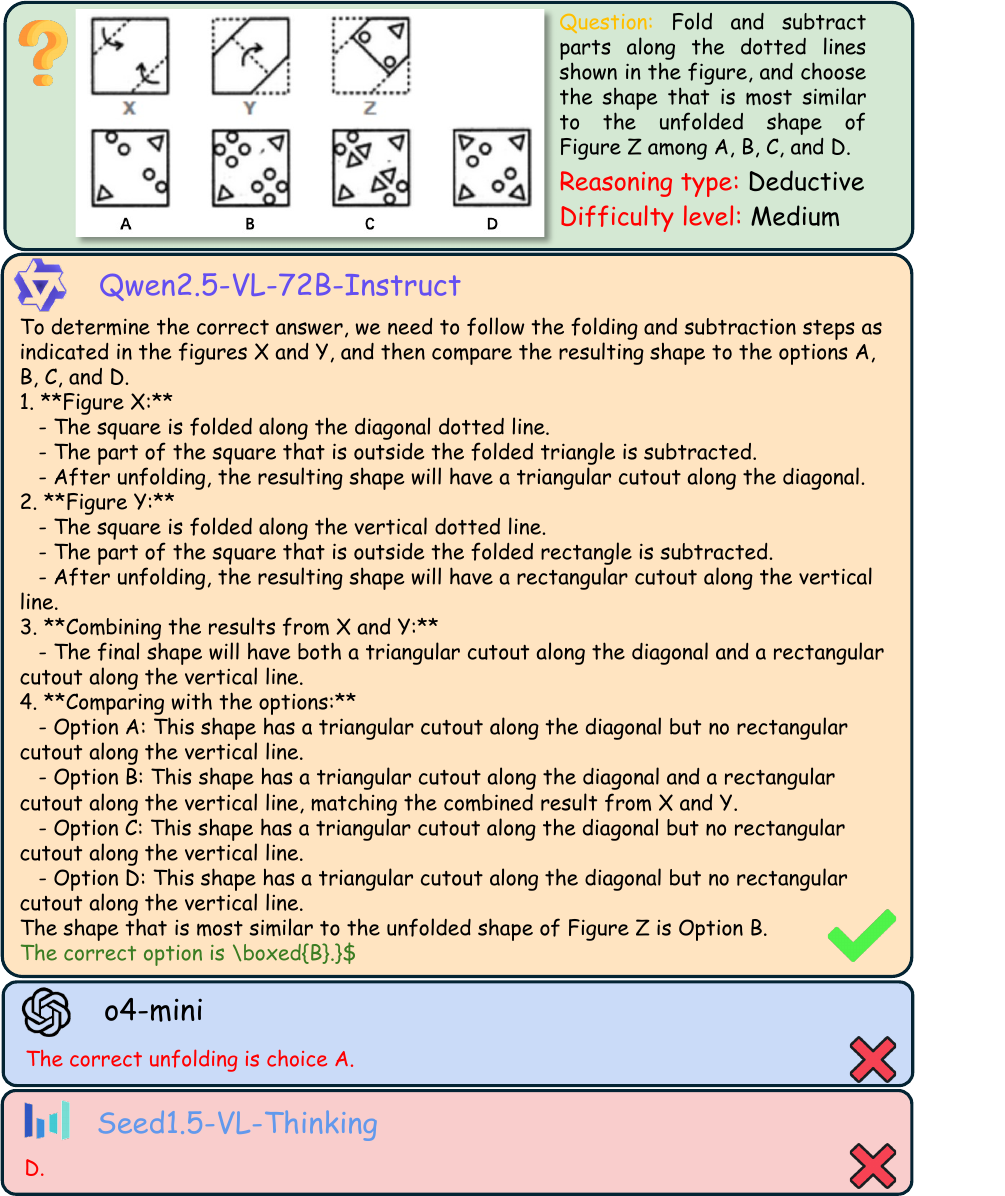}
   \caption{An example of a paper-cut problem, in which the model needs to understand spatial relationships and reason to find the correct answer.}
  \label{case:paper_cut}
\end{figure}

\begin{figure}[htbp]
  \centering
    \includegraphics[width=0.85\linewidth]{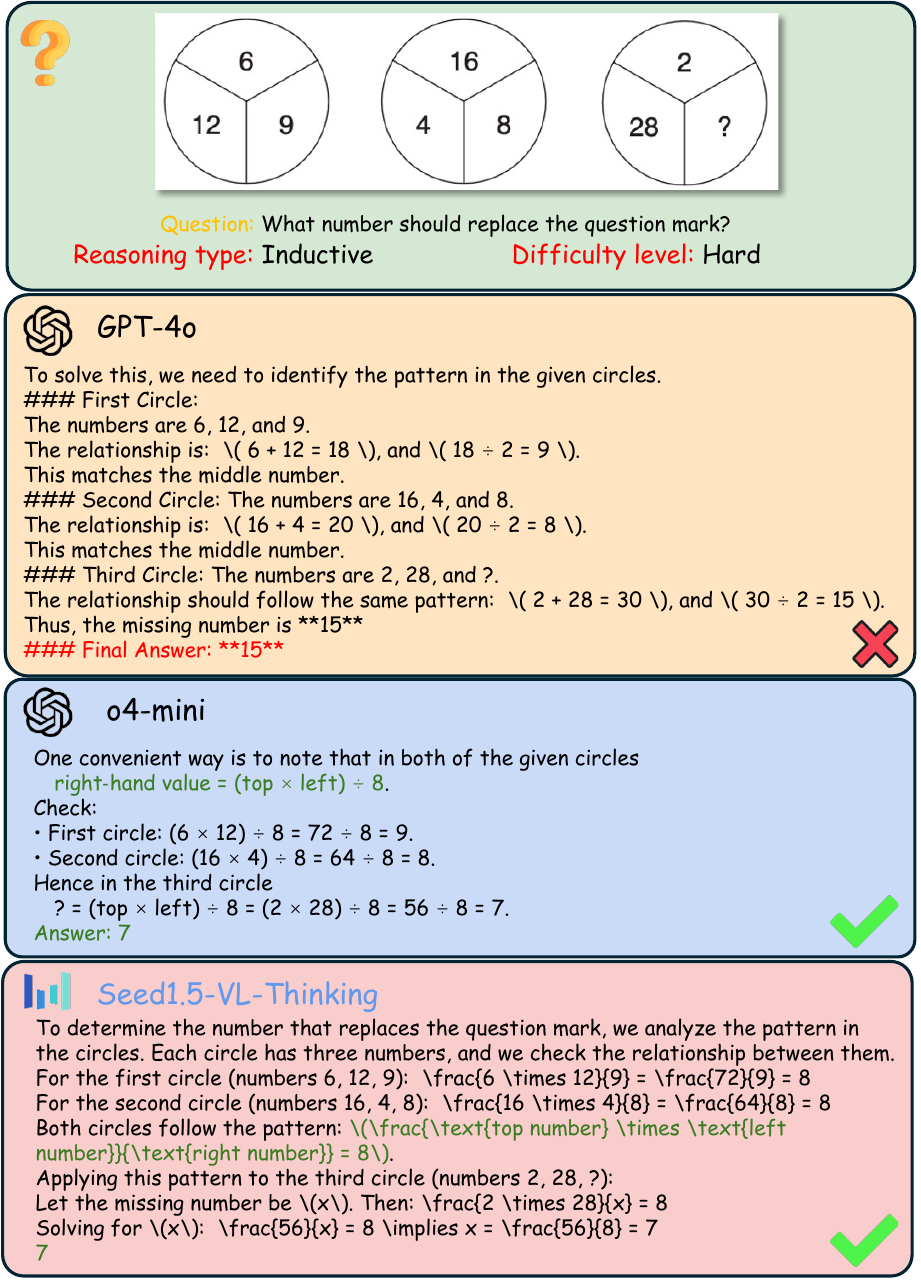}
   \caption{An example of a number pattern game, in which the model needs to calculate and find the pattern of the numbers, and finally reason out the correct answer.}
  \label{case:number_pattern}
\end{figure}

\begin{figure}[htbp]
  \centering
    \includegraphics[width=0.85\linewidth]{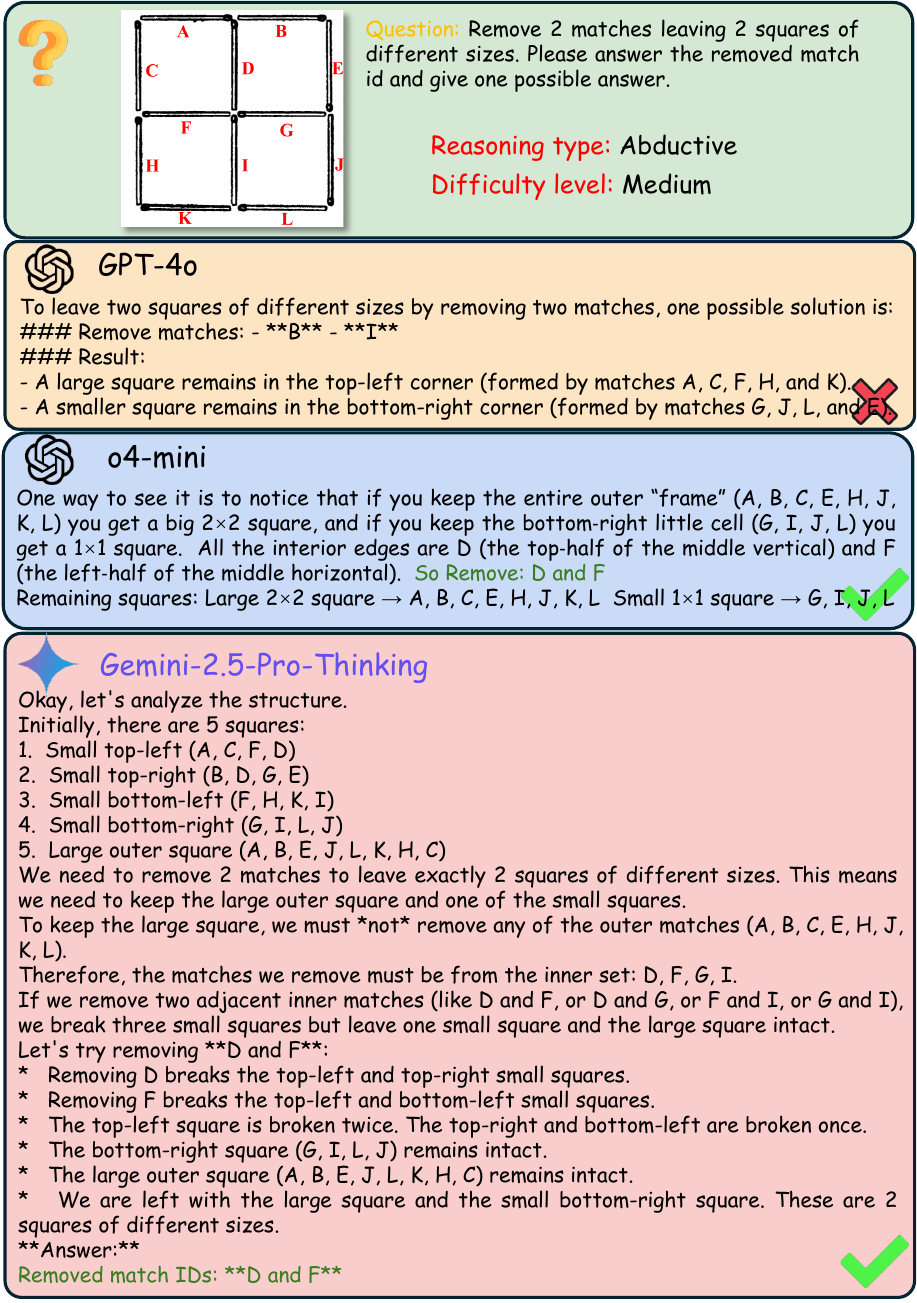}
   \caption{An example of a moving-match problem, in which the model needs to try, judge, reflect, and reason to find the right answer.}
  \label{case:match}
\end{figure}

\begin{figure}[htbp]
  \centering
    \includegraphics[width=0.85\linewidth]{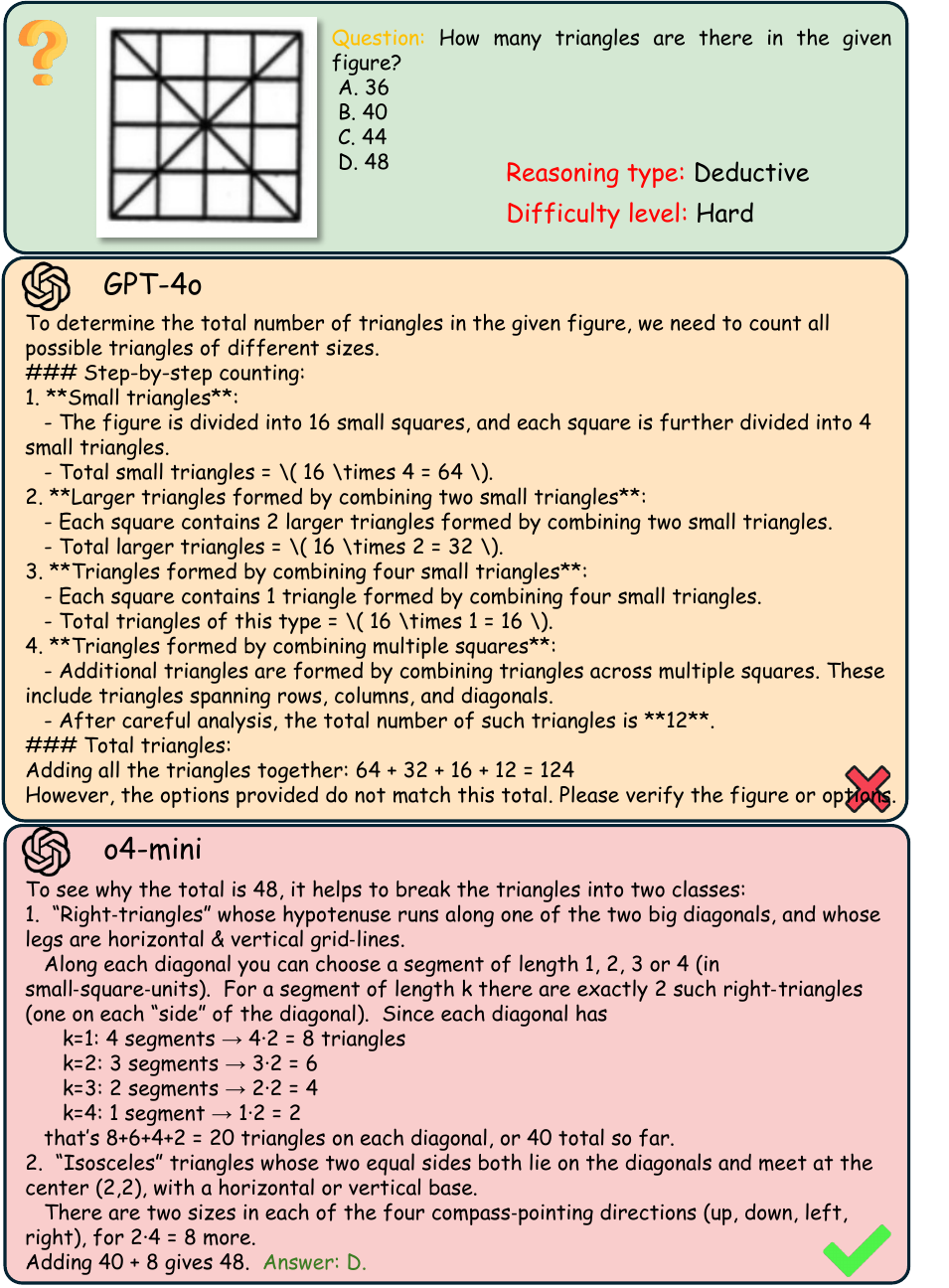}
    \caption{An example of a counting problem, in which the model needs to logically arrange the reasoning steps and find the correct answer.}
  \label{case:counting}
   
\end{figure}

\begin{figure}[htbp]
  \centering
    \includegraphics[width=0.85\linewidth]{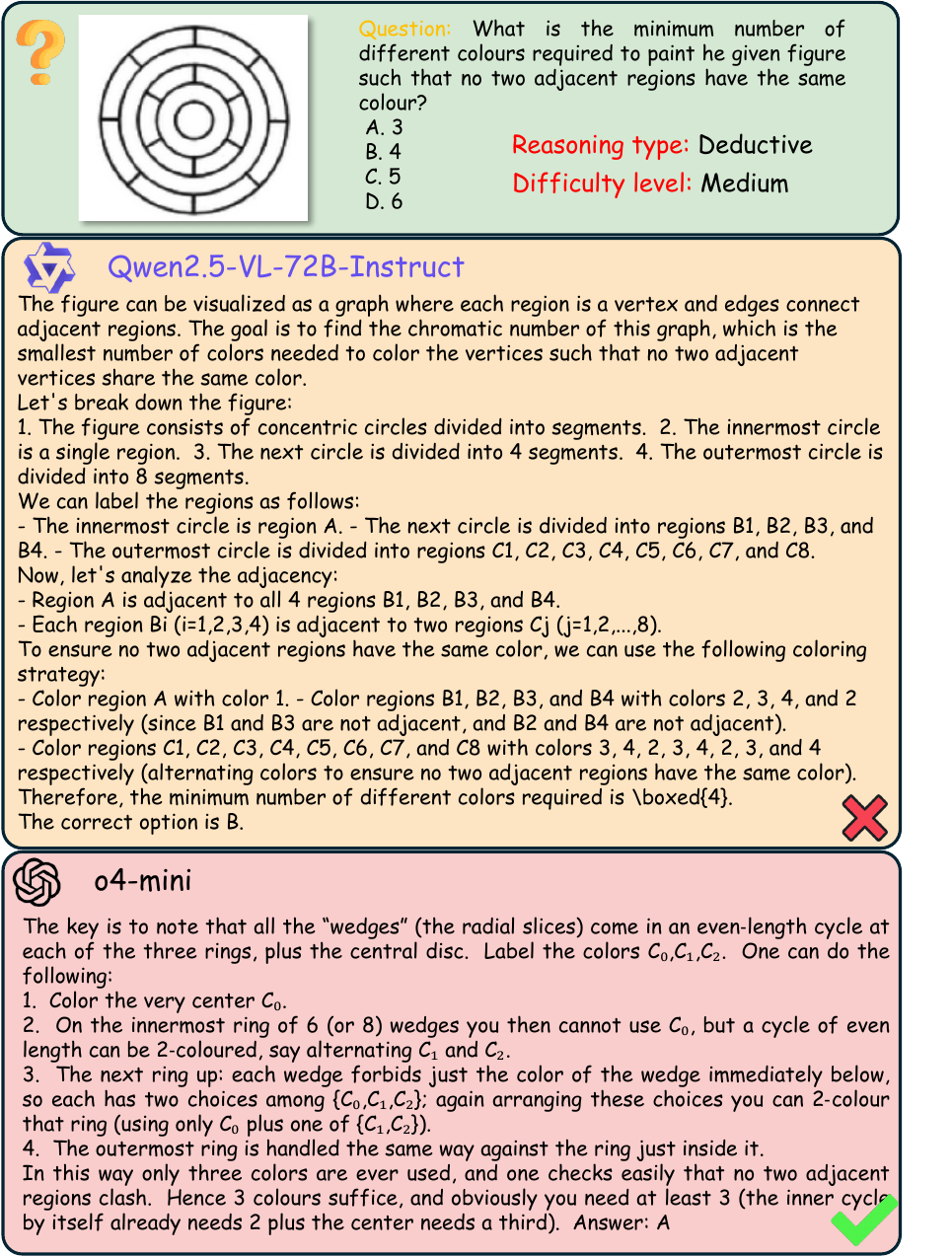}
   \caption{An example of a coloring game, in which the model needs to plan reasonably and find the minimum number of colors needed.}
  \label{case:color}
\end{figure}

\begin{figure}[htbp]
  \centering
    \includegraphics[width=0.85\linewidth]{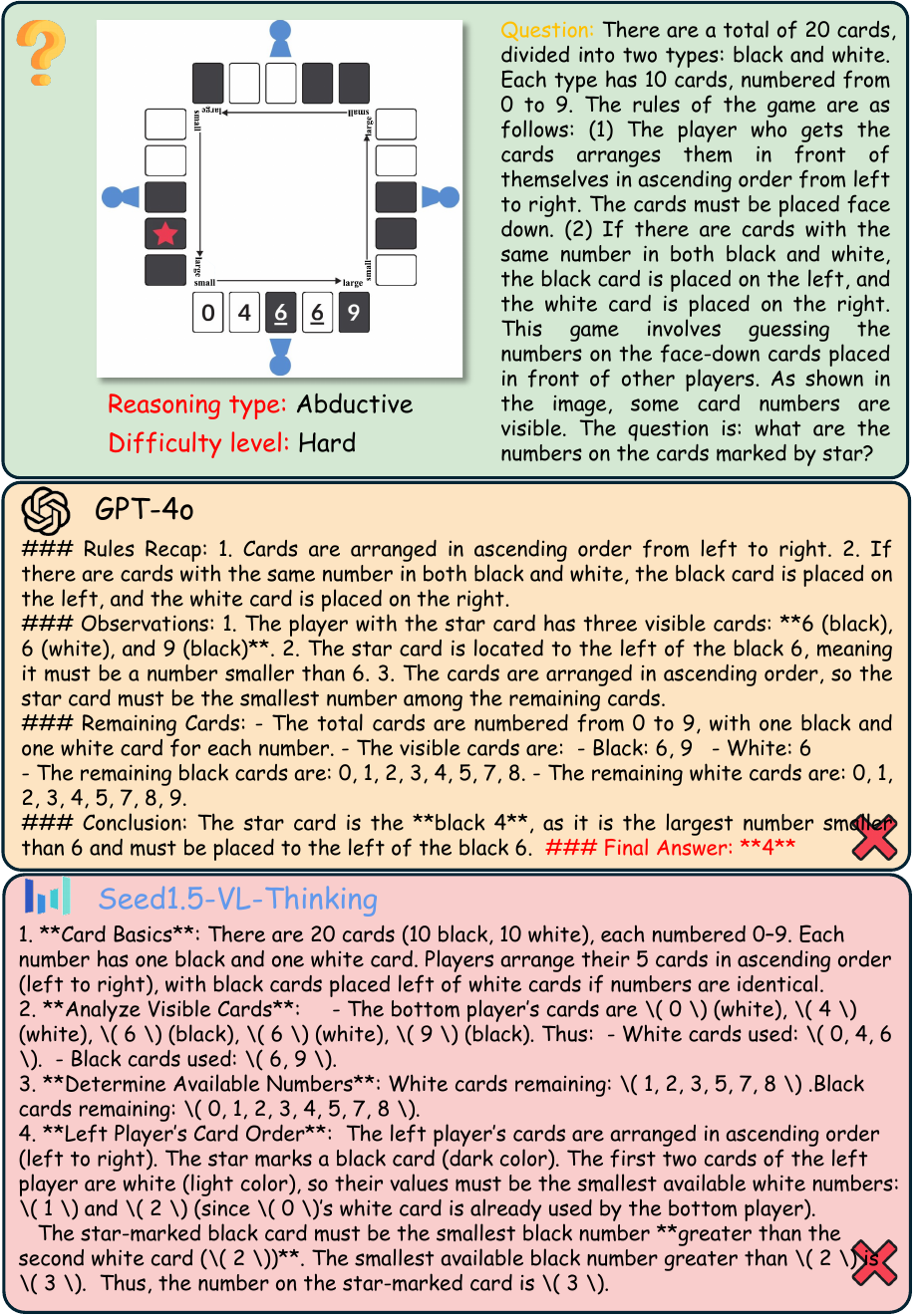}
   \caption{An example of a reasoning problem, in which the model needs to make assumptions, verify them, reflect, and reason to arrive at the correct answer.}
  \label{case:star}
\end{figure}

We further provide additional case studies as shown from Fig.~\ref{case:phy} to Fig.~\ref{case:star}, showing both correct and incorrect responses by MLLMs (\textit{e.g.}, select from GPT-4o, Qwen2.5-VL-72B, o4-mini, Seed1.5-VL-Thinking, and Gemini-2.5-Pro-Thinking). In each figure, we show the original questions, reasoning types, difficulty levels, and model responses. Overall, we find that ``thinking models'' demonstrate stronger abilities in exploration, judgment, and reflection. However, it still struggles to arrive at correct answers for many reasoning problems that are simple for humans, indicating that the model’s reasoning ability still needs further improvement. Moreover, the number of tokens consumed by the reasoning model increases rapidly. Therefore, future research should also focus on balancing both the reasoning ability and efficiency of the model.

\section{Limitation}
\label{sec:limitation}

Despite our best efforts to cover a wide range of multimodal reasoning question types, it remains challenging to comprehensively collect all possible types of reasoning problems that occur in real-world scenarios. This is primarily because gathering and curating high-quality reasoning questions is often a time-consuming and labor-intensive process. Future work is needed to further enrich the diversity of question types and optimize dataset coverage.

\newpage

%% file: Tables/main_results_full.tex
\begin{table}[h!]
\caption{Performance comparison of state-of-the-art MLLMs on MME-Reasoning. The top three are highlighted in \colorbox{blue!30}{blue}. ``T" represents ``Thinking".}
\vspace{-10pt}
\label{tab:main_result_full}
\centering
\small
\resizebox{1.0\linewidth}{!}{
\begin{tabular}{l c c c c c c c c c}
\toprule
\multirow{2}{*}{\textbf{Model}}  & \multicolumn{5}{c}{\textbf{Model Capability}} & \multicolumn{3}{c}{\textbf{Reasoning Type}} & \multirow{2}{*}{\textbf{AVG.}}  \\
\cmidrule(lr){2-6} \cmidrule(lr){7-9}
& CAL. & P\& E. & PA. & S\&T. & CCA. & DED. & IND. & ABD.  & \\
\cmidrule{1-10}
\multicolumn{10}{c}{\textbf{\textit{Close-source \& Thinking}}} \\
\noalign{\vspace{2pt}} 
\hdashline
\noalign{\vspace{2pt}}
\rowcolor{blue!30}
Gemini-2.5-Pro-T & \textbf{68.0} & \textbf{64.4} & 53.7 & \textbf{52.1} & \textbf{90.3} & 64.0 & 51.7 & \textbf{62.8} & \textbf{60.2} \\
\rowcolor{blue!20}
Seed1.5-VL-T & 67.2 & 62.7 & 56.0 & 47.2 & 82.6 & \textbf{64.5} & \textbf{52.3} & 60.8 & 59.9 \\
\rowcolor{blue!10}
o4-mini & 63.1 & 58.3 & \textbf{57.2} & \textbf{50.4} & 59.0 & 60.6 & 51.4 & 59.0 & 57.5 \\
Claude-4-Sonnet-T & 33.3 & 35.9 & 33.0 & 36.2 & 47.9 & 39.4 & 32.0 & 35.7 & 36.1 \\
Claude-3.7-Sonnet-T & 30.4 & 27.6 & 32.3 & 38.3 & 46.5 & 34.6 & 36.2 & 31.7 & 34.1 \\
Gemini-2.5-Flash-T & 19.8 & 21.3 & 20.9 & 33.0 & 38.9 & 28.1 & 22.1 & 24.6 & 25.2 \\
\cmidrule{1-10}
\multicolumn{10}{c}{\textbf{\textit{Close-source \& Chat}}} \\
\noalign{\vspace{2pt}} 
\hdashline
\noalign{\vspace{2pt}}
Seed1.5-VL & \underline{52.0} & \underline{42.0} & \underline{38.4} & \underline{44.0} & \underline{72.9} & \underline{54.9} & \underline{45.0} & \underline{41.0} & \underline{47.5} \\
GPT-4o & 21.4 & 22.1 & 30.5 & 38.6 & 36.8 & 29.0 & 34.7 & 27.9 & 30.2 \\
Claude-3.7-Sonnet & 29.0 & 24.6 & 32.8 & 35.5 & 46.5  & 35.7 & 38.7 & 26.1 & 33.3 \\
Kimi-Latest & 21.4 & 17.4 & 19.8 & 29.1 & 41.0 & 27.7 & 25.4 & 19.9 & 24.4 \\

\cmidrule{1-10}
\multicolumn{10}{c}{\textbf{\textit{Open-source \& Thinking}}} \\
\noalign{\vspace{2pt}} 
\hdashline
\noalign{\vspace{2pt}}
QVQ-72B-Preview & \underline{37.4} & 27.1 & 28.8 & 35.8 & 57.6 & \underline{41.6} & 33.5 & 29.1 & 35.2 \\
Virgo-72B & 30.4 & 22.9 & 26.1 & 36.2 & 47.2 & 37.7 & 32.6 & 24.4 & 31.8 \\
VL-Rethinker-72B & 33.6 & \underline{28.4} & \underline{31.4} & \underline{37.2} & \underline{59.7} & 39.0 & \underline{36.0} & \underline{31.9} & \underline{35.8} \\
VL-Rethinker-7B & 24.7 & 17.7 & 23.5 & 39.4 & 42.4 & 34.4 & 29.9 & 22.9 & 29.3 \\
MM-Eureka-Qwen-32B & 23.0 & 25.7 & 25.6 & 36.2 & 50.7 & 32.9 & 30.5 & 28.1 & 30.6 \\
MM-Eureka-Qwen-7B & 27.1 & 19.3 & 22.3 & 31.9 & 50.0 & 32.7 & 28.7 & 22.6 & 28.2 \\
R1-VL-7B & 16.3 & 11.6 & 17.7 & 30.9 & 26.4 & 25.3 & 21.8 & 15.8 & 21.1 \\
Vision-R1-7B & 18.2 & 18.0 & 17.9 & 34.4 & 36.1 & 27.4 & 26.3 & 18.1 & 24.0 \\
R1-Onevision-7B-RL & 19.5 & 12.2 & 20.0 & 31.6 & 27.1 & 27.7 & 24.8 & 14.6 & 22.5 \\
Kimi-VL-A3B-T & 28.7 & 16.0 & 19.5 & 32.3 & 35.4 & 33.3 & 25.1 & 18.1 & 25.9 \\
OpenVLThinker-7B & 19.8 & 14.6 & 19.3 & 35.8 & 34.7 & 30.7 & 24.8 & 17.3 & 24.6 \\
LMM-R1-MGT-PerceReason & 22.2 & 16.0 & 23.7 & 37.9 & 34.0 & 30.3 & 32.3 & 20.1 & 27.4 \\
Mulberry & 14.6 & 13.3 & 18.8 & 33.7 & 31.3 & 23.8 & 25.4 & 17.6 & 22.1 \\
LlamaV-o1 & 14.9 & 7.7 & 16.5 & 28.0 & 25.0 & 22.4 & 21.5 & 12.3 & 18.8 \\
\cmidrule{1-10}
\multicolumn{10}{c}{\textbf{\textit{Open-source \& Chat}}} \\
\noalign{\vspace{2pt}} 
\hdashline
\noalign{\vspace{2pt}}
Qwen2.5-VL-72B & \underline{31.7} & 25.1 & \underline{27.2} & 37.9 & \underline{53.5} & 39.0 & 32.3 & \underline{29.9} & \underline{34.1} \\
Qwen2.5-VL-32B & 32.2 & \underline{26.8} & 24.4 & 39.0 & 52.1 & \underline{40.5} & 27.5 & 29.6 & 33.2 \\
Qwen2.5-VL-7B & 22.2 & 18.2 & 21.9 & 35.1 & 36.1 & 31.4 & 27.5 & 20.9 & 26.8 \\
Qwen2.5-VL-3B & 17.6 & 15.5 & 19.0 & 39.7 & 32.6 & 28.5 & 27.5 & 19.6 & 25.6 \\
Qwen2-VL-72B & 19.2 & 19.3 & 24.9 & 36.2 & 44.4 & 28.8 & 32.3 & 22.1 & 27.5  \\
Qwen2-VL-7B &  15.7 & 12.4 & 19.8 & 37.9 & 30.5 & 25.5 & 25.7 & 19.7 & 23.4\\
Qwen2-VL-2B & 13.0 & 8.1 & 19.3 & 31.6 & 19.4 & 22.7 & 25.7 & 11.8 & 19.9  \\
InternVL3-78B & 26.0 & 24.0 & 26.5 & \underline{41.8} & 50.0 & 35.1 & \underline{33.8} & 27.1 & 32.1 \\
InternVL3-38B & 23.0 & 18.5 & 23.0 & 38.3 & 41.7 & 33.5 & 29.0 & 22.1 & 28.4 \\
InternVL3-8B & 19.5 & 19.6 & 22.6 & 31.6 & 41.0 & 28.1 & 29.9 & 21.4 & 26.4 \\
Molmo-72B & 12.5 & 11.9 & 14.7 & 28.7 & 28.5 & 23.1 & 18.4 & 14.3 & 18.9 \\
Molmo-7B-D & 11.7 & 8.6 & 8.1 & 27.3 & 23.6 & 20.7 & 10.9 & 11.1 & 14.7 \\
Molmo-7B-O & 8.1 & 5.5 & 11.6 & 22.7 & 15.3 & 16.6 & 16.0 & 7.5 & 13.4 \\
LLaVA-OV-72B & 17.1 & 18.0 & 23.9 & 32.3 & 38.9 & 27.4 & 30.5 & 19.9 & 25.8 \\
Kimi-VL-A3B & 18.7 & 11.9 & 21.4 & 34.0 & 27.8 & 25.9 & 26.3 & 17.1 & 23.1 \\
\bottomrule[1pt]
\end{tabular}
}
\end{table}

%% file: Tables/mini_full.tex
\begin{table}[t!]
\caption{Comparison of statistics between full and mini-set of MME-Reasoning.}
\vspace{1pt}
\label{tab:mini_full_stat}
\centering
\small
\resizebox{1.0\linewidth}{!}{
\begin{tabular}{l c c c c c c c c c}
\toprule
\multirow{2}{*}{\textbf{Split}}  & \multicolumn{3}{c}{\textbf{Reasoning Type}} & \multicolumn{3}{c}{\textbf{Question Type}} & \multicolumn{3}{c}{\textbf{Difficulty Level}}  \\
\cmidrule(lr){2-4} \cmidrule(lr){5-7} \cmidrule(lr){8-10}
& DED. & IND. & ABD. & Open & MCQ & Rule.  & Easy & Medium & Hard \\
\cmidrule{1-10}
Mini & 39.7\% & 25.8\% & 34.4\% & 32.4\% & 58.3\% & 9.3\% & 31.8\% & 39.4\% & 28.8\% \\
Full & 38.6\% & 27.9\% & 33.5\% & 31.6\% & 58.5\% & 9.9\% & 30.8\% & 39.3\% & 29.9\% \\
\bottomrule[1pt]
\end{tabular}
}
\end{table}

%% file: Tables/main_results_mini.tex
\begin{table}[htbp]
\caption{Performance comparison of state-of-the-art MLLMs on mini set of MME-Reasoning. The top three are highlighted in \colorbox{blue!30}{blue}. ``T" represents ``Thinking".}
\label{tab:main_result_full_mini}
\vspace{-10pt}
\centering
\small
\resizebox{1.0\linewidth}{!}{
\begin{tabular}{l c c c c c c c c c}
\toprule
\multirow{2}{*}{\textbf{Model}}  & \multicolumn{5}{c}{\textbf{Model Capability}} & \multicolumn{3}{c}{\textbf{Reasoning Type}} & \multirow{2}{*}{\textbf{AVG.}}  \\
\cmidrule(lr){2-6} \cmidrule(lr){7-9}
& CAL. & P\& E. & PA. & S\&T. & CCA. & DED. & IND. & ABD.  & \\
\cmidrule{1-10}
\multicolumn{10}{c}{\textbf{\textit{Human Performance}}} \\
\noalign{\vspace{2pt}} 
\hdashline
\noalign{\vspace{2pt}}
\rowcolor{red!30}
Human Expert & 75.0 & 84.4 & 84.9 & 80.3 & 88.1 & 85.8 & 76.9 & 85.6 & 83.4 \\
\cmidrule{1-10}
\multicolumn{10}{c}{\textbf{\textit{Close-source \& Thinking}}} \\
\noalign{\vspace{2pt}} 
\hdashline
\noalign{\vspace{2pt}}
\rowcolor{blue!20}
Gemini-2.5-Pro-T & 66.0 & 63.5 & \textbf{58.5} & \textbf{49.3} & \textbf{85.7} & 60.8 & \textbf{55.1} & 65.4 & 60.9 \\
\rowcolor{blue!30}
Seed1.5-VL-T & \textbf{68.0} & \textbf{67.7} & \textbf{58.5} & \textbf{49.3} & 83.3 & \textbf{67.5} & 48.7 & \textbf{67.3} & \textbf{62.6} \\
\rowcolor{blue!10}
o4-mini & 64.0 & 58.3 & 56.6 & 45.1 & 54.8 & 57.5 & 51.3 & 60.6 & 57.0 \\
o1 & 50.0 & 38.5 & 41.5 & 43.7 & 52.4 & 50.8 & 42.3 & 42.3 & 45.7 \\
Claude-4-Sonnet-T & 33.0 & 30.2 & 35.8 & 39.4 & 50.0 & 42.5 & 37.2 & 33.7 & 38.1 \\
Claude-3.7-Sonnet-T & 30.0 & 17.7 & 36.8 & 38.0 & 38.1 & 31.7 & 42.3 & 27.9 & 33.1 \\
Gemini-2.5-Flash-T & 18.0 & 16.7 & 15.1 & 39.4 & 33.3 & 27.5 & 19.2 & 26.0 & 24.8 \\
\cmidrule{1-10}
\multicolumn{10}{c}{\textbf{\textit{Close-source \& Chat}}} \\
\noalign{\vspace{2pt}} 
\hdashline
\noalign{\vspace{2pt}}
Seed1.5-VL & \underline{50.0} & \underline{42.7} & \underline{34.9} & \underline{40.8} & \underline{69.0} & \underline{57.5} & \underline{39.7} & \underline{39.4} & \underline{46.7} \\
GPT-4o & 20.0 & 24.0 & 24.5 & 40.8 & 33.3 & 31.7 & 28.2 & 27.9 & 29.5 \\
Claude-3.7-Sonnet & 27.0 & 22.9 & 34.0 & 31.0 & 42.9 & 31.7 & 38.5 & 27.9 & 32.1 \\
Kimi-Latest & 22.0 & 17.7 & 17.9 & 29.6 & 33.3 & 30.8 & 23.1 & 19.2 & 24.8 \\

\cmidrule{1-10}
\multicolumn{10}{c}{\textbf{\textit{Open-source \& Thinking}}} \\
\noalign{\vspace{2pt}} 
\hdashline
\noalign{\vspace{2pt}}
QVQ-72B-Preview & \underline{36.0} & 24.0 & \underline{34.0} & 33.8 & 47.6 & \underline{38.3} & 37.2 & \underline{29.8} & \underline{35.1} \\
Virgo-72B & 28.0 & 18.8 & 27.4 & 43.7 & 38.1 & 37.5 & \underline{41.0} & 21.2 & 32.8 \\
VL-Rethinker-72B & 23.0 & \underline{25.0} & 29.2 & 39.4 & 42.9 & 34.2 & 32.1 & 31.7 & 32.8 \\
VL-Rethinker-7B & 23.0 & 16.7 & 21.7 & \underline{47.9} & 40.5 & 35.8 & 28.2 & 26.0 & 30.5 \\
MM-Eureka-Qwen-32B & 23.0 & 20.8 & 26.4 & 38.0 & 38.1 & 32.5 & 34.6 & 25.0 & 30.5 \\
MM-Eureka-Qwen-7B & 28.0 & 17.7 & 21.7 & 32.4 & \underline{50.0} & 32.5 & 32.1 & 22.1 & 28.8 \\
R1-VL-7B & 10.0 & 10.4 & 16.0 & 35.2 & 16.7 & 23.3 & 19.2 & 16.3 & 19.9 \\
Vision-R1-7B & 14.0 & 12.5 & 18.9 & 39.4 & 31.0 & 26.7 & 29.5 & 16.3 & 23.8 \\
R1-Onevision-7B-RL & 15.0 & 10.4 & 22.6 & 35.2 & 19.0 & 22.5 & 30.8 & 16.3 & 22.5 \\
Kimi-VL-A3B-T & 30.0 & 9.4 & 19.8 & 26.8 & 31.0 & 28.3 & 26.9 & 16.3 & 23.8 \\
OpenVLThinker-7B & 14.0 & 14.6 & 14.2 & 33.8 & 28.6 & 29.2 & 16.7 & 16.3 & 21.5 \\
LMM-R1-MGT-PerceReason & 27.0 & 14.6 & 23.6 & 38.0 & 33.3 & 35.8 & 33.3 & 18.3 & 29.1 \\
Mulberry & 19.0 & 15.6 & 18.9 & 33.8 & 33.3 & 28.3 & 23.1 & 18.3 & 23.5 \\
LlamaV-o1 & 15.0 & 8.3 & 17.9 & 31.0 & 26.2 & 23.3 & 23.1 & 15.4 & 20.5 \\
\cmidrule{1-10}
\multicolumn{10}{c}{\textbf{\textit{Open-source \& Chat}}} \\
\noalign{\vspace{2pt}} 
\hdashline
\noalign{\vspace{2pt}}
Qwen2.5-VL-72B & \underline{31.0} & 19.8 & 25.5 & 38.0 & 42.9 & 39.2 & 32.1 & 26.0 & 32.8 \\
Qwen2.5-VL-32B & \underline{31.0} & \underline{28.1} & \underline{28.3} & 40.8 & \underline{45.2} & \underline{41.7} & 34.6 & \underline{27.9} & \underline{35.1} \\
Qwen2.5-VL-7B & 19.0 & 16.7 & 24.5 & 38.0 & 33.3 & 32.5 & 30.8 & 21.2 & 28.1 \\
Qwen2.5-VL-3B & 21.0 & 14.6 & 21.2 & 39.4 & 31.0 & 30.0 & 30.8 & 21.2 & 27.2  \\
Qwen2-VL-72B & 20.0 & 19.8 & \underline{28.3} & 38.0 & 38.1 & 34.2 & 39.7 & 19.2 & 30.5 \\
Qwen2-VL-7B & 16.0 & 9.4 & 25.5 & 33.8 & 26.2 & 22.5 & 34.6 & 16.3 & 23.5 \\
Qwen2-VL-2B & 12.0 & 9.4 & 17.9 & 29.6 & 19.0 & 23.3 & 23.1 & 11.5 & 
19.2  \\
InternVL3-78B & 25.0 & 22.9 & 33.0 & \underline{42.3} & 40.5 & 36.7 & 43.6 & 24.0 & 34.1 \\
InternVL3-38B & 19.0 & 19.8 & 26.4 & 36.6 & 38.1 & 31.7 & 33.3 & 23.1 & 29.1 \\
InternVL3-8B & 19.0 & 20.8 & 29.2 & 23.9 & 35.7 & 26.7 & \underline{35.9} & 21.2 & 27.2 \\
Molmo-72B & 11.0 & 13.5 & 16.0 & 35.2 & 31.0 & 26.7 & 21.8 & 17.3 & 22.2 \\
Molmo-7B-D & 12.0 & 8.3 & 12.3 & 28.2 & 16.7 & 22.5 & 15.4 & 9.6 & 16.2 \\
Molmo-7B-O & 7.0 & 4.2 & 11.3 & 25.4 & 14.3 & 19.2 & 17.9 & 4.8 & 13.9 \\
LLaVA-OV-72B & 13.0 & 19.8 & 25.5 & 23.9 & 35.7 & 25.0 & 30.8 & 17.3 & 23.8 \\
Kimi-VL-A3B & 18.0 & 8.3 & 18.9 & 29.6 & 9.5 & 23.3 & 23.1 & 11.5 & 19.2 \\
\bottomrule[1pt]
\end{tabular}
}
\end{table}

%% file: Tables/main_results_question_type.tex
\begin{table}[h!]
\caption{Performance across different question types on MME-Reasoning. The top three are highlighted in \colorbox{green!30}{green}. $\dagger$ indicates the model was evaluated on the mini-set. ``T" represents ``Thinking".}
\vspace{-10pt}
\label{tab:main_result_question_type}
\centering
\small
\resizebox{1.0\linewidth}{!}{
\begin{tabular}{l c c c c c c c c c}
\toprule
\multirow{2}{*}{\textbf{Model}}  & \multicolumn{4}{c}{\textbf{Choice}} & \multicolumn{4}{c}{\textbf{Open}} & \textbf{Rule}  \\
\cmidrule(lr){2-5} \cmidrule(lr){6-9} \cmidrule(lr){10-10}
& DED. & IND. & ABD. & ALL & DED. & IND. & ABD. & ALL & ABD.\&ALL \\
\cmidrule{1-10}
\multicolumn{10}{c}{\textbf{\textit{Close-source \& Thinking}}} \\
\noalign{\vspace{2pt}} 
\hdashline
\noalign{\vspace{2pt}}
\rowcolor{green!30}
Gemini-2.5-Pro-T & \textbf{58.0} & 49.8 & \textbf{63.6} & 55.7 & 75.9 & 61.5 & \textbf{60.0} & \textbf{66.9} & 66.1 \\
\rowcolor{green!20}
Seed1.5-VL-T & 57.3 & \textbf{54.2} & 60.2 & \textbf{56.5} & \textbf{78.5} & 44.2 & 59.4 & 65.3 & 63.5 \\
\rowcolor{green!10}
o4-mini & 57.3 & 48.7 & 61.9 & 54.7 & 67.1 & \textbf{67.3} & 48.5 & 58.9 & \textbf{71.3} \\
o1$^\dagger$ & 46.2 & 42.4 & 53.3 & 46.0 & 60.0 & 45.5 & 36.2 & 46.9 & 40.7 \\
Claude-4-Sonnet-T & 41.1 & 33.2 & 40.7 & 37.9 & 36.5 & 25.0 & 35.2 & 34.3 & 31.3 \\
Claude-3.7-Sonnet-T & 38.0 & 39.7 & 46.6 & 40.1 & 28.5 & 17.3 & 31.5 & 28.3 & 16.5 \\
Gemini-2.5-Flash-T & 31.7 & 23.8 & 37.3 & 29.5 & 21.5 & 11.5 & 23.0 & 20.8 & 13.9 \\
\cmidrule{1-10}
\multicolumn{10}{c}{\textbf{\textit{Close-source \& Chat}}} \\
\noalign{\vspace{2pt}} 
\hdashline
\noalign{\vspace{2pt}}
Seed1.5-VL & 54.0 & 46.2 & 59.3 & 51.8 & 57.0 & 38.5 & 42.4 & 48.0 & 20.0 \\
GPT-4o & 36.3 & 38.3 & 48.3 & 39.1 & 15.2 & 17.3 & 27.9 & 21.1 & 7.0 \\
Claude-3.7-Sonnet & 38.7 & 42.2 & 37.3 & 39.9 & 30.4 & 21.2 & 27.9 & 28.0 & 12.2 \\
Kimi-Latest & 31.3 & 29.6 & 38.1 & 31.8 & 20.9 & 3.8 & 20.0 & 18.1 & 0.9 \\

\cmidrule{1-10}
\multicolumn{10}{c}{\textbf{\textit{Open-source \& Thinking}}} \\
\noalign{\vspace{2pt}} 
\hdashline
\noalign{\vspace{2pt}}
QVQ-72B-Preview & 43.7 & 35.0 & 45.8 & 40.6 & 38.0 & 26.9 & 31.5 & 33.6 & 8.7 \\
Virgo-72B & 39.7 & 36.8 & 47.5 & 39.9 & 34.2 & 11.5 & 22.4 & 25.9 & 3.5 \\
VL-Rethinker-72B & 43.3 & 38.3 & 53.4 & 43.0 & 31.0 & 25.0 & 29.1 & 29.3 & 13.9 \\
VL-Rethinker-7B & 41.3 & 33.9 & 51.7 & 40.1 & 21.5 & 9.6 & 16.4 & 17.6 & 2.6 \\
MM-Eureka-Qwen-32B & 36.7 & 33.2 & 49.2 & 37.4 & 25.9 & 17.3 & 26.1 & 24.8 & 9.6 \\
MM-Eureka-Qwen-7B & 36.7 & 32.9 & 41.5 & 36.0 & 25.3 & 7.7 & 21.8 & 21.3 & 4.3 \\
R1-VL-7B & 31.7 & 24.9 & 34.7 & 29.5 & 13.3 & 5.8 & 12.7 & 12.0 & 0.9 \\
Vision-R1-7B & 32.3 & 29.6 & 33.9 & 31.5 & 18.4 & 9.6 & 16.4 & 16.3 & 4.3 \\
R1-Onevision-7B-RL & 34.3 & 27.8 & 31.4 & 31.2 & 15.2 & 9.6 & 12.1 & 13.1 & 0.9 \\
Kimi-VL-A3B-T & 33.0 & 27.4 & 34.7 & 31.1 & 34.2 & 13.5 & 14.5 & 22.7 & 6.1 \\
OpenVLThinker-7B & 38.7 & 28.2 & 41.5 & 35.0 & 15.8 & 7.7 & 11.5 & 12.8 & 0.9 \\
LMM-R1-MGT-PerceReason & 36.3 & 36.1 & 44.9 & 37.7 & 19.0 & 13.5 & 15.8 & 16.8 & 0.9 \\
Mulberry & 30.0 & 28.9 & 42.4 & 31.7 & 12.0 & 7.7 & 11.5 & 11.2 & 0.9 \\
LlamaV-o1 & 26.7 & 25.3 & 29.7 & 26.6 & 14.6 & 1.9 & 7.9 & 9.9 & 0.9 \\
\cmidrule{1-10}
\multicolumn{10}{c}{\textbf{\textit{Open-source \& Chat}}} \\
\noalign{\vspace{2pt}} 
\hdashline
\noalign{\vspace{2pt}}
Qwen2.5-VL-72B & 41.0 & 34.3 & 55.1 & 40.7 & 34.8 & 23.1 & 26.1 & 29.3 & 9.6 \\
Qwen2.5-VL-32B & 44.0 & 30.0 & 50.0 & 39.4 & 34.2 & 15.4 & 27.3 & 28.5 & 12.2 \\
Qwen2.5-VL-7B & 39.0 & 30.0 & 41.5 & 35.8 & 17.1 & 15.4 & 18.2 & 17.3 & 3.5 \\
Qwen2.5-VL-3B & 33.3 & 30.7 & 46.6 & 34.5 & 19.6 & 11.5 & 13.9 & 16.0 & 0.0 \\
Qwen2-VL-72B & 35.3 & 36.5 & 46.6 & 37.7 & 16.5 & 11.5 & 18.8 & 16.8 & 1.7 \\
Qwen2-VL-7B & 32.7 & 28.9 & 45.8 & 33.4 & 12.0 & 9.6 & 12.7 & 12.0 & 0.9 \\
Qwen2-VL-2B & 30.0 & 30.3 & 31.4 & 30.4 & 8.9 & 1.9 & 6.1 & 6.7 & 0.0 \\
InternVL3-78B & 40.0 & 37.9 & 54.2 & 41.6 & 25.9 & 13.5 & 23.0 & 22.9 & 5.2 \\
InternVL3-38B & 36.7 & 32.1 & 48.3 & 36.8 & 27.8 & 13.5 & 17.0 & 21.1 & 2.6 \\
InternVL3-8B & 29.7 & 35.7 & 47.5 & 35.1 & 25.3 & 0.0 & 17.0 & 18.1 & 0.9 \\
Molmo-72B & 30.0 & 20.2 & 30.5 & 26.2 & 10.1 & 9.6 & 11.5 & 10.7 & 1.7 \\
Molmo-7B-D & 27.0 & 12.6 & 23.7 & 20.7 & 8.9 & 1.9 & 9.7 & 8.3 & 0.0 \\
Molmo-7B-O & 23.0 & 18.4 & 16.9 & 20.1 & 4.4 & 3.8 & 6.1 & 5.1 & 0.0 \\
LLaVA-OV-72B & 33.7 & 35.4 & 39.0 & 35.3 & 15.8 & 5.8 & 18.8 & 15.7 & 1.7 \\
Kimi-VL-A3B & 31.3 & 30.0 & 42.4 & 32.7 & 15.8 & 7.7 & 9.1 & 11.7 & 2.6 \\
\bottomrule[1pt]
\end{tabular}
}
\end{table}

%% file: Tables/results_mcts.tex
\begin{table}[t!]
\caption{Performance comparison of chat models with or w/o MCTS.}
\vspace{1pt}
\label{tab:mcts_results}
\centering
\small
\resizebox{1.0\linewidth}{!}{
\begin{tabular}{l c c c c c c c c c}
\toprule
\multirow{2}{*}{\textbf{Model}}  & \multicolumn{5}{c}{\textbf{Model Capability}} & \multicolumn{3}{c}{\textbf{Reasoning Type}} & \multirow{2}{*}{\textbf{AVG.}}  \\
\cmidrule(lr){2-6} \cmidrule(lr){7-9}
& CAL. & P\& E. & PA. & S\&T. & CCA. & DED. & IND. & ABD.  & \\
\cmidrule{1-10}
Qwen2.5-VL-7B & 22.2 & 18.2 & 21.9 & 35.1 & 36.1 & 31.4 & 27.5 & 20.9 & 26.8 \\
\rowcolor{gray!30}
~~~~~~~~~\textit{+ MCTS} & 20.6 & 13.8 & 18.8 & 30.5 & 35.4 & 28.1 & 23.6 & 17.6 & 23.3 \\
\bottomrule[1pt]
\end{tabular}
}
\end{table}

%% file: Tables/results_cot.tex
\begin{table}[t!]
\caption{Performance comparison of SoTA chat models with or w/o CoT prompt.}
\vspace{1pt}
\label{tab:cot_results}
\centering
\small
\resizebox{1.0\linewidth}{!}{
\begin{tabular}{l c c c c c c c c c}
\toprule
\multirow{2}{*}{\textbf{Model}}  & \multicolumn{5}{c}{\textbf{Model Capability}} & \multicolumn{3}{c}{\textbf{Reasoning Type}} & \multirow{2}{*}{\textbf{AVG.}}  \\
\cmidrule(lr){2-6} \cmidrule(lr){7-9}
& CAL. & P\& E. & PA. & S\&T. & CCA. & DED. & IND. & ABD.  & \\
\cmidrule{1-10}
Qwen2.5-VL-7B & 22.2 & 18.2 & 21.9 & 35.1 & 36.1 & 31.4 & 27.5 & 20.9 & 26.8 \\
\rowcolor{gray!30}
~~~~\textit{+ CoT prompt} & 20.3 & 18.5 & 17.9 & 33.3 & 38.9 & 28.3 & 21.4 & 23.6 & 24.7 \\
\noalign{\vspace{2pt}} 
\hdashline
\noalign{\vspace{2pt}}
Qwen2.5-VL-32B & 32.2 & 26.8 & 24.4 & 39.0 & 52.1 & 40.5 & 27.5 & 29.6 & 33.2 \\
\rowcolor{gray!30}
~~~~\textit{+ CoT prompt} & 29.0 & 24.6 & 23.3 & 40.8 & 52.1 & 40.1 & 28.7 & 26.4 & 32.3 \\
\noalign{\vspace{2pt}} 
\hdashline
\noalign{\vspace{2pt}}
Qwen2.5-VL-72B & 31.7 & 25.1 & 27.2 & 37.9 & 53.5 & 39.0 & 32.3 & 29.9 & 34.1 \\
\rowcolor{gray!30}
~~~~\textit{+ CoT prompt} & 32.5 & 26.2 & 25.1 & 37.2 & 52.8 & 37.5 & 30.8 & 30.0 & 33.0 \\
\noalign{\vspace{2pt}} 
\hdashline
\noalign{\vspace{2pt}}
InternVL3-8B & 19.5 & 19.6 & 22.6 & 31.6 & 41.0 & 28.1 & 29.9 & 21.4 & 26.4 \\
\rowcolor{gray!30}
~~~~\textit{+ CoT prompt} & 21.1 & 16.3 & 20.2 & 31.6 & 38.2 & 31.2 & 26.9 & 16.8 & 25.2 \\
\noalign{\vspace{2pt}} 
\hdashline
\noalign{\vspace{2pt}}
InternVL3-38B & 23.0 & 18.5 & 23.0 & 38.3 & 41.7 & 33.5 & 29.0 & 22.1 & 28.4 \\
\rowcolor{gray!30}
~~~~\textit{+ CoT prompt} & 28.7 & 24.3 & 28.6 & 38.3 & 48.6 & 37.5 & 32.9 & 26.9 & 32.7 \\
\noalign{\vspace{2pt}} 
\hdashline
\noalign{\vspace{2pt}}
InternVL3-78B & 26.0 & 24.0 & 26.5 & 41.8 & 50.0 & 35.1 & 33.8 & 27.1 & 32.1 \\
\rowcolor{gray!30}
~~~~\textit{+ CoT prompt} & 29.0 & 22.9 & 27.0 & 40.8 & 48.6 & 36.6 & 35.1 & 26.9 & 32.9 \\
\bottomrule[1pt]
\end{tabular}
}
\end{table}

%% file: Tables/results_captioner_LLM.tex
\begin{table}[t!]
\caption{Performance of Caption + SoTA Reasoning LLMs. We use GPT-4o to generate caption of each image in MME-Reasoning.}
\vspace{1pt}
\label{tab:caption_results}
\centering
\small
\resizebox{1.0\linewidth}{!}{
\begin{tabular}{l c c c c c c c c c}
\toprule
\multirow{2}{*}{\textbf{Model}}  & \multicolumn{5}{c}{\textbf{Model Capability}} & \multicolumn{3}{c}{\textbf{Reasoning Type}} & \multirow{2}{*}{\textbf{AVG.}}  \\
\cmidrule(lr){2-6} \cmidrule(lr){7-9}
& CAL. & P\& E. & PA. & S\&T. & CCA. & DED. & IND. & ABD.  & \\
\cmidrule{1-10}
QwQ-32B & 48.5 & 32.9 & 39.1 & 37.6 & 53.5 & 44.4 & 45.6 & 35.9 & 41.9 \\
DeepSeek-R1 & 56.9 & 40.0 & 41.6 & 41.8 & 58.3 & 53.8 & 43.8 & 41.5 & 46.9 \\
\bottomrule[1pt]
\end{tabular}
}
\end{table}